\def\allfiles{}
\documentclass[lettersize,journal]{IEEEtran}
\usepackage{amsmath, amsfonts}
\usepackage{algorithmic}
\usepackage{algorithm}
\usepackage{array}
\usepackage{xr-hyper}
\makeatletter
\newcommand*{\addFileDependency}[1]{
  \typeout{(#1)}
  \@addtofilelist{#1}
  \IfFileExists{#1}{}{\typeout{No file #1.}}
}
\makeatother

\newcommand*{\myexternaldocument}[1]{
    \externaldocument{#1}
    \addFileDependency{#1.tex}
    \addFileDependency{#1.aux}
}
\usepackage[colorlinks, linkcolor=red, citecolor=blue]{hyperref}
\usepackage[caption=false]{subfig}
\usepackage{textcomp}
\usepackage{stfloats}
\usepackage{url}
\usepackage{verbatim}
\usepackage{graphicx}
\usepackage{caption}
\usepackage{mathrsfs}
\usepackage{array}
\usepackage{multirow}
\usepackage{multicol}
\usepackage{cite}
\usepackage{amsthm}
\usepackage{booktabs}
\usepackage{threeparttable}

\usepackage{amssymb}

\makeatletter
\DeclareRobustCommand\onedot{\futurelet\@let@token\@onedot}
\def\@onedot{\ifx\@let@token.\else.\null\fi\xspace}

\makeatother

\renewcommand{\eqref}[1]{Eq.~\ref{#1}}

\newcommand{\boldparagraph}[1]{\vspace{0.2cm}\noindent{\bf #1:} }

\newif\ifcomment
\commenttrue
\ifcomment

	\newcommand{\yl}[1]{ \noindent {\color{red} {\bf yl:} {#1}} }
\else
	\newcommand{\yl}[1]{}
\fi

\usepackage[dvipsnames]{xcolor}
\newtheorem{lemma}{Lemma}
\newtheorem{theorem}{Theorem} 
\newtheorem{definition}{Definition}
\begin{document}

% \title{RING++: Rotation-Invariant and Density-Tolerant Place Recognition in Resource-Limited Environments}
\title{RING++: Roto-translation Invariant Gram for Global Localization on a Sparse Scan Map}

\author{Xuecheng Xu$^{*}$, Sha Lu$^{*}$, Jun Wu, Haojian Lu, Qiuguo Zhu, Yiyi Liao, Rong Xiong and Yue Wang
        % <-this % stops a space
\thanks{$^{*}$These two authors contribute equally to this work.}
\thanks{Code will be available at https://github.com/MaverickPeter/MR\_SLAM}
\thanks{The authors are with Zhejiang University, Zhejiang, China. \textit{Corresponding author: Yue Wang.} (e-mail: xuechengxu@zju.edu.cn; lusha@zju.edu.cn; \mbox{wujun\_csecyber@zju.edu.cn;} luhaojian@zju.edu.cn; qgzhu@iipc.zju.edu.cn; lyyecho1119@gmail.com; rxiong@zju.edu.cn; wangyue@iipc.zju.edu.cn)}% <-this % stops a space

}% \thanks{Manuscript received April 19, 2021; revised August 16, 2021.}}

% \markboth{Journal of \LaTeX\ Class Files,~Vol.~14, No.~8, August~2022}%


\maketitle

\ifx\allfiles\undefined

\begin{document}
\fi

\begin{abstract}
Global localization plays a critical role in many robot applications. LiDAR-based global localization draws the community's focus with its robustness against illumination and seasonal changes. To further improve the localization under large viewpoint differences, we propose RING++ which has roto-translation invariant representation for place recognition, and global convergence for both rotation and translation estimation. With the theoretical guarantee, RING++ is able to address the large viewpoint difference using a lightweight map with sparse scans. In addition, we derive sufficient conditions of feature extractors for the representation preserving the roto-translation invariance, making RING++ a framework applicable to generic multi-channel features. To the best of our knowledge, this is the first learning-free framework to address all subtasks of global localization in the sparse scan map. Validations on real-world datasets show that our approach demonstrates better performance than state-of-the-art learning-free methods, and competitive performance with learning-based methods. Finally, we integrate RING++ into a multi-robot/session SLAM system, performing its effectiveness in collaborative applications. 
\end{abstract}


\ifx\allfiles\undefined
\end{document}
\fi
\ifx\allfiles\undefined

\begin{document}
\fi

\section{Introduction}
\IEEEPARstart{G}{lobal} localization aims to estimate the pose of a robot in a map using onboard sensor measurements without priors. This task is essential for many robotics applications, including loop closures and map alignments in SLAM systems, and relocalization in navigation systems. Vision-based methods have advanced rapidly in the last decade by exploiting image cues. However, these approaches are sensitive to illumination and seasonal changes between current and mapping session \cite{valgren2007sift,lowry2015visual}. Recent studies have shown that LiDAR can be employed to overcome the difficulty \cite{kim2018scan,wang2020lidar,kavisha2021locus,komorowski2021minkloc3d}.
%However, when current and mapping session have trajectories with large viewpoint difference, reliable global localization still remains a challenge even for LiDAR based methods.
However, even with LiDAR-based methods, reliable localization is still challenging when the current scan and mapping sessions have trajectories with large viewpoint differences.

% While in situation (2), considering the larger field of view of LiDAR, we raise a question: \textit{Is a robot equipped with a LiDAR able to localize itself when a large viewpoint difference presents?}

\begin{figure}[ht]
	\centering
	\includegraphics[width=\linewidth]{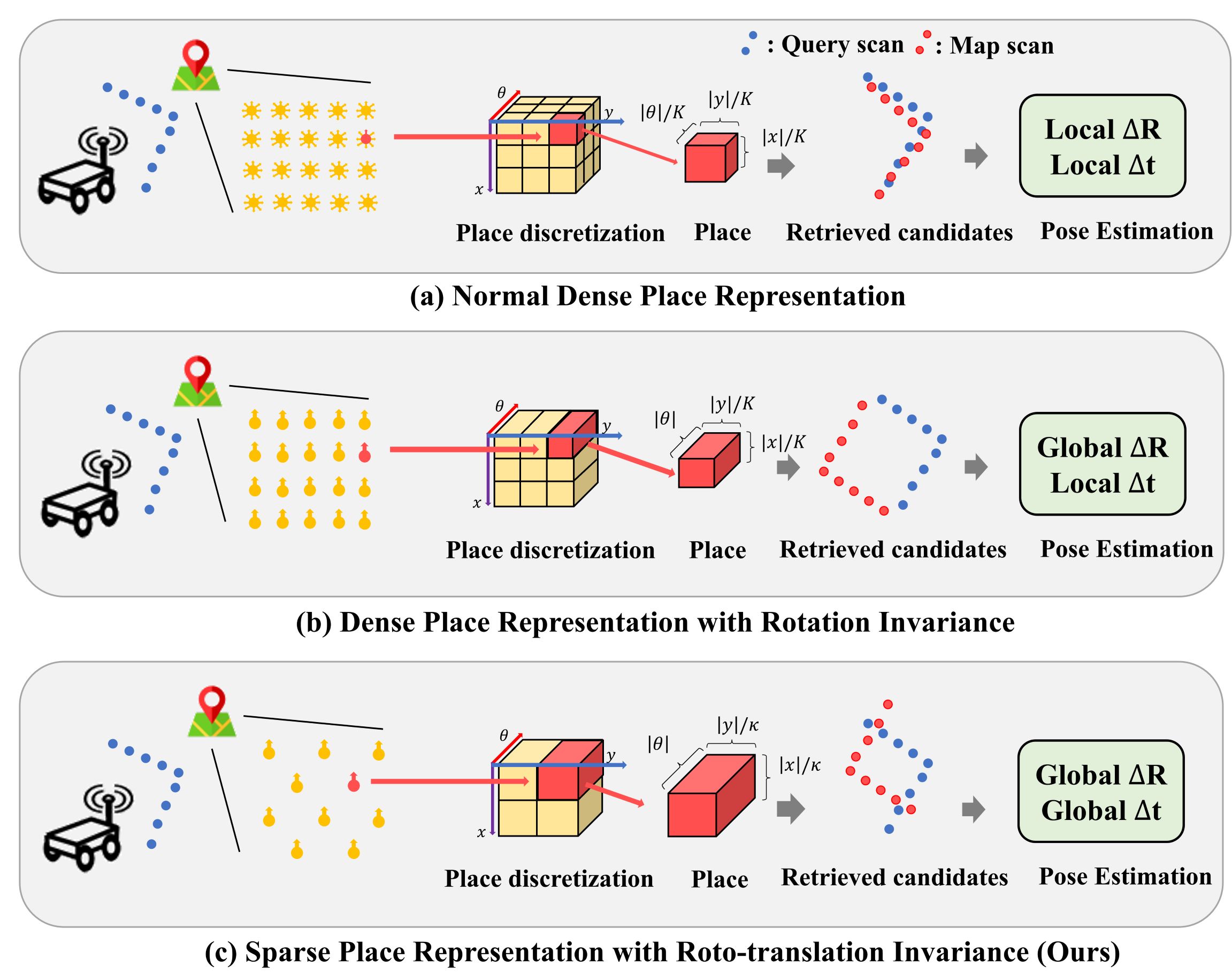}
	\caption{Demonstration of global localization pipelines consisting of place recognition and pose estimation. $|\cdot|$ means the range of the axis. $K$ and $\kappa$ are the number of discretized bins and $K \gg \kappa$, which determines the density of place i.e. reference scans in the map. Our method aims at leveraging a sparse scan map for localization under large viewpoint differences using roto-translation invariant representation, as well as global convergent solvers.}
	\label{fig:Teaser}
    \vspace{-0.5cm}
\end{figure}

To clarify the challenge, we define the global localization as a search problem: the query is the current LiDAR scan, and the search space is the pose space covering the entire map. A typical global localization pipeline consists of two steps: place recognition and pose estimation, solving the problem in a coarse-to-fine process. As shown in Fig.~\ref{fig:Teaser}, these two steps are divided by the \textit{place}, which encodes a representative scan and its pose. At the coarse level, the search space is discretized into many places. Accordingly, place recognition aims at finding the place to which the robot is closest. At the fine level, the search space is the continuous pose space centered at the pose of the place. Taking the pose of the place as initialization, the pose estimation process aims to obtain the accurate pose of the query. When the viewpoint difference is large, a query and a map scan taken in the same place can be different, which brings ambiguity to place recognition and calls for a large convergence basin in pose estimation. 

To address the viewpoint difference, early works utilize 3-DoF (1-DoF rotation and 2-DoF translation) discretization~\cite{rohling2015fast,he2016m2dp,uy2018pointnetvlad,liu2019lpd}, regarding scans with slight viewpoint differences as different places. This is illustrated in the first row of Fig.~\ref{fig:Teaser}. Such formulation relaxes the place recognition to only considering small viewpoint difference, and simplifies the pose estimation to be local convergent, demonstrating good performance. But it calls for large memory to save dense places. The second row of Fig.~\ref{fig:Teaser} illustrates a pioneering work, Scan Context~\cite{kim2018scan,kim2021scan}, which regards scans with arbitrary rotation as the same place by explicitly modeling the rotation invariance for place representation. Besides, it proposes a global convergent rotation solver in the pose estimation stage, as no rotation initialization is available from place recognition with rotation invariance. Thus the discretization is reduced to 2-DoF (2D translation). Unfortunately, the rotation invariance is sensitive to the translation difference, hence requiring dense discretization in translation space. Considering the advantage of explicitly modeling rotation invariance, we raise a question: \textit{Is it possible to build a representation that is invariant to both rotation and translation?}

In this paper, we address the large viewpoint difference in global localization using a sparse scan map. As illustrated in the third row in Fig.~\ref{fig:Teaser}, we propose a framework RING++ that achieves \textit{roto-translation invariance} for place recognition, and \textit{global convergence} for both rotation and translation estimation, theoretically guaranteeing the performance under large viewpoint differences. Therefore, only a sparse discretization of map pose space in 2-DoF translation is sufficient. RING++ takes bird-eye view (BEV) of the scan, and processes in two passes: the representation pass for building roto-translation invariant representation and the solving pass for place recognition and pose estimation. In the representation pass, we exploit the properties of Radon transform and Fourier transform on BEV representation to generate roto-translation invariant place representation, and derive sufficient conditions of the feature extractor for the representation to preserve the roto-translation invariance. In the solving pass, the roto-translation invariant representation leads the successful place recognition even when large viewpoint differences exist between the query and map scans. By employing the cross correlation, the rotation and translation are further estimated with global convergence. To the best of our knowledge, RING++ is the first learning-free framework to address all subtasks of global localization in the sparse scan map. The experimental results validate our method in both place recognition and pose estimation, and show a superior performance compared to existing methods. To summarize, this work presents the following contributions:
\begin{itemize}
\item We propose a novel learning-free framework named RING++ for global localization on the sparse scan map, which can simultaneously solve place recognition and pose estimation tasks.
\item We theoretically prove the roto-translation invariance property of our place representation. Meanwhile, the pose estimation solvers are presented with global convergence.
\item We further derive sufficient conditions of feature extractors for the representation preserving the roto-translation invariance, enabling the framework to aggregate multi-channel features.
\item We validate RING++ on three real-world datasets and multi-robot SLAM applications in the wild. We make the code, SLAM system, and evaluation tools all  publicly available.
\end{itemize}

In the preliminary conference paper \cite{lu2022one}, we proposed a method that utilizes a single-channel occupancy map to generate roto-translation invariant place representation. In this paper, we formally state and prove the roto-translation invariance of our representation to provide a theoretical guarantee. Then we introduce the aggregation method to take multi-channel features into the framework, yielding better discrimination to enhance both place recognition and pose estimation. We also derive the sufficient conditions of the feature extractor that can preserve the roto-translation invariance of our representation. Besides, we develop a global and effective correlation-based translation solver to address the outlier sensitivity in the conference version, which is non-trivial as the scans with large viewpoint differences must have less overlap. With all efforts above, the overall success rate of RING++ increases by almost $20\%$. We also validate the performance of RING++ with more ablation studies, substantial experiments with both learning-free and learning methods in diverse scenarios, as well as multi-robot collaborative SLAM.
\ifx\allfiles\undefined
\end{document}
\fi

\ifx\allfiles\undefined

\begin{document}
\fi

\section{Related Works}
In this section, we provide a literature review on LiDAR-based recognition and estimation tasks with two main lines embedded in previous works. To begin with, local point features are evolving to represent the point cloud in a compact and efficient manner. To overcome noise and variances, different aggregation strategies are proposed to generate robust global representations with invariance properties from local features.

\subsection{Feature Extraction}
In the early stages of scan-based recognition, researchers focus on how to generate compact representations of scans. There are many ways to achieve this goal, but the early consensus is to find compact and effective local features. Geometric relations are first explored in many approaches. Stein et al. \cite{stein1992structural} propose Structural Indexing (SI) which constructs a representation from 3D curves and splashes. Rusu et al. \cite{rusu2008aligning} build PFH (Point Feature Histograms) by aggregating four handcrafted features which stem from normals and the distance between k-nearest point pairs. In the large-scale scene interpretation task, Weinmann et al. \cite{weinmann2014semantic} provide eight semantic point features calculated by normalized eigenvalues of each point with its neighbors. With the wide use of local features, researchers found that fine resolution features are easily influenced by noises. To overcome this limitation, some coarser features are presented. Spin Image \cite{johnson1999using} counts point numbers in each volume of cylinder support around a keypoint. ESF \cite{wohlkinger2011ensemble} method uses voxel grids to approximate the surface and encodes shape properties. SHOT \cite{tombari2011combined, SALTI2014251} combines signature and histogram in a local reference frame and encodes the cosine value of the angle between the normal and the local z-axis at the feature point.

Although these features are widely used for scan-based recognition tasks, they are unstable when applied to the LiDAR-based place recognition scenario. The poor generalization performance is caused by the sensor characteristics of the newly equipped LiDAR. Points collected by LiDAR are unstructured and the sparsity of points varies along with the sensor range. In order to tackle this problem, learning-based methods are proposed. Due to the strong descriptive ability of the 2D convolution neural network (CNN), some researchers extend it to 3D cases by representing point clouds in 3D volumes \cite{maturana2015voxnet,wu20153d,zhou2018voxelnet}. However, these CNN-based methods usually introduce quantization errors and high computational costs. To alleviate the drawbacks brought by CNN, PointNet \cite{qi2017pointnet} first learns features directly from the raw 3D point cloud data. PointNet++ \cite{qi2017pointnet++} further introduces the hierarchical feature learning to learn local features with increasing scales. To better acquire relationships between local points, graph-based \cite{wang2019dynamic,liu2019lpd} and kernel-based \cite{shen2018mining} networks are proposed. Other than geometry information, OverlapNet \cite{chen2021overlapnet} and SSC \cite{li2021ssc} also take semantic clues into account. Some recent works propose fusion frameworks for incorporating image features as well \cite{komorowski2021minkloc++, pan2021coral, lai2021adafusion}. Despite the fact that different methods provide different local features, they are all presented in a multi-channel manner. With this characteristic in mind, we propose a multi-channel framework that can be viewed as a feature aggregation module, allowing us to use different local features and improve discrimination with roto-translation invariance.

\subsection{Aggregation and Invariance}
With local features extracted, global descriptors are often generated by aggregating the local features. Early aggregation methods can be divided into two classes: signature and histogram.

Signature describes the 3D support by defining a local reference frame and encoding local features into the subset of the support. Histogram describes the support by encoding counts of local features. Structural Indexing \cite{stein1992structural} is one of the first methods to use signatures to capture 3D curves. 3D SURF \cite{knopp2010hough} extends the mature 2D SURF \cite{bay2006surf} by voxelizing the 3D data and utilizing Haar wavelet response to determine the saliency of each voxel.

More previous works prefer the histogram since it provides a coarser representation of the point cloud that is robust to slight variance. PFH \cite{rusu2008aligning, rusu2009fast} and VFH \cite{rusu2010fast} are the early methods that explicit introduce translation and rotation invariance. VFH finds the viewpoint directions and counts the angles between the normals in a histogram. SHOT \cite{tombari2011combined} collects the histogram of normal angles in a spherical bin around a keypoint to build the descriptor. Rohling \cite{rohling2015fast} first utilize histogram-based similarity measures in robotics systems to find loop closures.

With the appearance of learning-based local features, new aggregation methods are proposed in the place recognition task. NetVLAD \cite{arandjelovic2016netvlad}, first introduced in visual place recognition, modifies VLAD \cite{jegou2010aggregating} with learnable weights and integrates it into a CNN. Following the idea of NetVLAD, several methods \cite{uy2018pointnetvlad,liu2019lpd} apply NetVLAD to supervise the feature learning. DiSCO \cite{xu2021disco} utilizes Fourier transform to generate a global descriptor in the frequency domain. Apart from these specially designed aggregation methods, MinkLoc3D and its extensions \cite{komorowski2021minkloc3d,zywanowski2021minkloc3d,komorowski2021egonn} adopt the GeM \cite{radenovic2018fine} pooling layer to generate discriminative global descriptors. 

Inspired by early works on rotation invariance, many approaches also focus on achieving invariance against viewpoint difference. M2DP \cite{he2016m2dp} presents a multi-view projection on the point cloud and uses PCA to compensate viewpoint difference. LocNet \cite{yin2018locnet} aggregates the distance of consecutive points in the same ring to a rotation invariant histogram and adopts a siamese network for feature learning. Iris \cite{wang2020lidar} explicitly models rotation invariance using Fourier transform on polar images. Previous works only present rotation invariant representation for place recognition, Scan context and its extension \cite{kim2018scan, kim2021scan} also realize lateral invariance and simultaneously estimate 1-DoF rotation and 1-DoF lateral translation, which is effective in autonomous driving. However, in Scan context, translation invariance is achieved through augmentation based on urban road assumptions. When there is a large difference in viewpoint, translation invariance is invalid. DiSCO \cite{xu2021disco} adopts the invariance property of the Fourier transform to simultaneously estimate rotation and achieve invariance. Since the polar transform used in DiSCO is not translation invariant, it suffers from the distortion caused by large translation variance. To overcome the influence brought by the large translation, Ding et al. \cite{ding2022translation} utilized Radon transform to estimate 1-DoF rotation with robust translation invariance. In this work, we further derive a multi-channel framework based on the Radon transform which can estimate the 3-DoF pose and preserve roto-translation invariance.

\ifx\allfiles\undefined
\end{document}
\fi
\ifx\allfiles\undefined

\myexternaldocument{10-appendix.tex}
\begin{document}
\fi

\section{Overview}
\label{problem statement}
%We claim that an excellent place recognition algorithm needs to be invariant to both orientation and translation with compact and sparse place representation. In this section, we will define the global localization problem, give mathematical definitions of equivariance and invariance, introduce place density concept, and then formally demonstrate the requirements of an excellent place recognition algorithm.

LiDAR-based global localization aims to estimate the pose $T_Q$ of the query scan $P_Q$ in a scan map $\mathfrak{M} \triangleq \{P_i,T_i\}$, which is a database populated by map scans $P_i$ with registered poses $T_i$. In general, global localization is solved in two stages: place recognition and pose estimation. In the place recognition stage, a map scan with a large overlap with the query scan is retrieved from the map, denoted as $P_n$, which indicates that $P_Q$ is collected at the place near $P_n$ in the map. Then in pose estimation stage, the metric pose $T_Q$ is achieved by estimating the relative pose $T_{nQ}$ between $P_Q$ and $P_n$, and applying $T_{nQ}$ to $T_n$.

When the robot trajectory in the current session is different from the map session, $P_Q$ and $P_n$ can overlap but with $T_Q$ and $T_n$ being different, especially in rotation e.g. opposite direction. At the same time, $|\mathfrak{M}|$, the place density of a map, is expected to be sparse for storage and efficiency. As a result, the main challenge for place recognition is to build representation that is similar under large viewpoint differences, i.e. variance of the relative pose $T_{nQ}$, which further causes the challenge in pose estimation: the estimator should be globally convergent, as no reliable initial value for $T_{nQ}$ is available.

\subsection{Equivariance and Invariance}

To deal with the pose variance in place recognition, the invariant representation of the scan is built. However, when the representation is invariant to relative pose, the pose estimation becomes impossible to solve, calling for additional representations bridging the raw scan and the invariant representation. Following this idea, we formally introduce equivariance and invariance in the context of global localization. The representation is denoted as an operation $g$ taking the scan $P$ as input. $P_{d}$ and $P_{\alpha}$ indicate that the scan is translated by $d$, or rotated by $\alpha$. $g_d$ and $g_{\alpha}$ indicate that the representation is translated by $d$ or rotated by $\alpha$. Then we have the definitions:
\begin{definition}[Translation Equivariance]
\label{translation equivariance}
If the operation $g$ is translation equivariant, it satisfies
\begin{equation}
    g_{d}(P) = g(P_{d})
\end{equation}
\end{definition}

\begin{definition}[Rotation Equivariance]
\label{rotation equivariance}
If the operation $g$ is rotation equivariant, it satisfies the equation
\begin{equation}
    g_{\alpha}(P) = g(P_{\alpha})
\end{equation}
\end{definition}

\begin{definition}[Translation Invariance]
\label{translation invariance}
If the operation $g$ is translation invariant, it satisfies the equation
\begin{equation}
    g(P) = g(P_{d})
\end{equation}
\end{definition}

\begin{definition}[Rotation Invariance]
\label{rotation invariance}
If the operation $g$ is rotation invariant, it satisfies the equation
% Let $g$ be a representation of a place and $g_{o}$ be the representation of the same place with an arbitrary rotation. If the algorithm is rotation invariant, then it satisfies the equation
\begin{equation}
    g(P) = g(P_{\alpha})
\end{equation}
\end{definition}

\subsection{Framework}

\begin{figure*}[htbp]
	\centering
	\includegraphics[width=18cm]{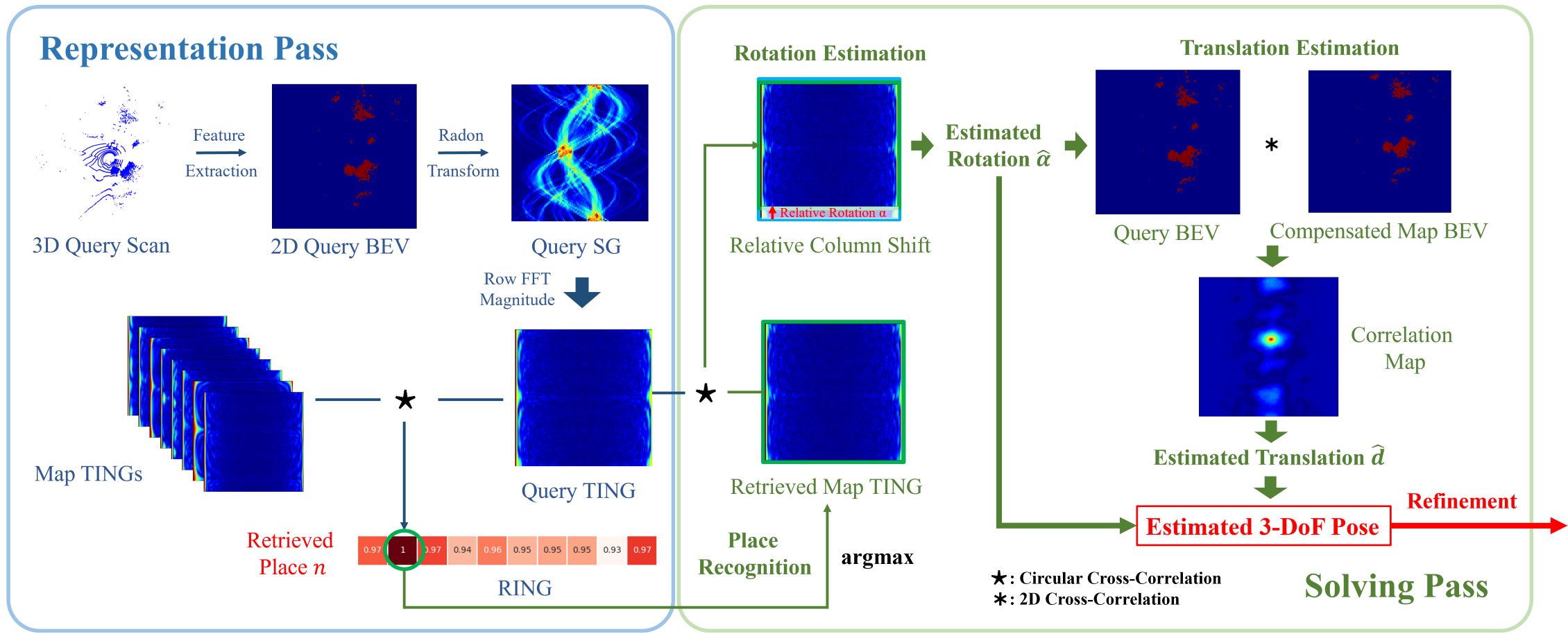}
	\caption{Overall framework of the proposed method. RING representation is used for place recognition, TING representation is utilized for rotation estimation and BEV representation is leveraged for translation estimation.}
	\label{fig:framework}
    \vspace{-0.2cm}
\end{figure*}

As shown in Fig.~\ref{fig:framework}, RING++ extracts features from scans, and builds representations for place recognition, rotation estimation and translation estimation respectively. Totally, there are two passes in RING++, a forward representation pass and a backward solving pass.

In the representation pass, the rotation equivariant scan feature of $P_Q$ is represented as sinogram (SG) by Radon transform (RT), then represented as translation invariant rotation equivariant gram (TING) by discrete Fourier transform (DFT), and finally represented as roto-translation invariant gram (RING) by batch circular cross-correlation.
% complex-valued translation invariant gram (CTING) by DFT for roto-translation similarity metric construction in the solving pass.
% roto-translation invariant gram (RING) by DFT. 

In the solving pass, RING is utilized for place recognition with different query/mapping trajectories and low place density, by which the caused pose difference becomes invariant in RING. Then TING is utilized for relative rotation estimation given the retrieved map scan $P_n$, since the translation in TING is invariant. Finally, BEV is utilized for relative translation estimation after the rotation in BEV is compensated by the estimation. Thanks to the decoupling leveraged by invariance, rotation and translation in $T_{nQ}$ can be solved independently via globally convergent solvers.

\section{Representation and Solving}
\label{framework}

\begin{figure}[tp]
	\centering
	\includegraphics[width=7cm]{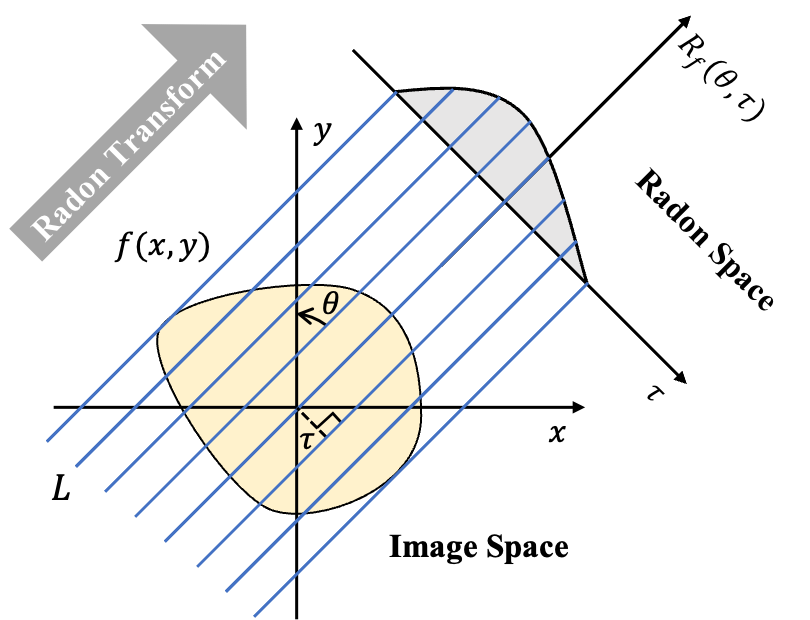}
	\caption{Graph illustration of Radon transform, which demonstrates a single row of the sinogram with a constant $\theta$ after Radon transform.}
	\label{fig:RT}
	\vspace{-0.5cm}
\end{figure}

\subsection{Representation Pass}
\subsubsection{BEV}
The first step of RING++ is to extract the features from the scan. Following common pipelines, we eliminate the ground from the 3D point cloud. Then we voxelize the scan into a 3D volume, in which each voxel encodes the occupancy, indicating whether there is a point inside. With the extracted features, we accumulate the height dimension of 3D volume to generate a 2D BEV representation $f(x,y) \in \mathbb{R}$.

\subsubsection{Sinogram} 
Given $f(x,y)$, we apply Radon transform $\mathcal{R}$ to yield a sinogram. Radon transform is a linear integral transform which maps $f(x, y)$ from the original image space $(x, y)$ to the Radon parameter space $(\theta,\tau)$, which is demonstrated in Fig.~\ref{fig:RT}. Denoting the line for integral in Radon transform as $L$, we have
\begin{equation}
L:~x\cos\theta + y\sin\theta = \tau
\end{equation}
where $\theta \in [0, 2\pi)$ represents the angle between $L$ and the $y$ axis, and $\tau \in (-\infty, \infty)$ represents the perpendicular distance from the origin to $L$. By formalizing the integral results into a 2D function with the Radon parameter as axes, we have $\mathcal{R}_{f}(L)=\mathcal{R}_{f}(\theta, \tau)$, namely sinogram (SG). Specifically, the Radon transform is calculated as:
\begin{equation}
\begin{split}
& \mathcal{R}_{f}(\theta, \tau)=\int_{x\cos\theta + y\sin\theta = \tau} f(x,y) \mathrm{d}x \mathrm{d}y \\ &=\int_{-\infty}^{\infty}\int_{-\infty}^{\infty} f(x,y) \delta(\tau-x\cos\theta-y\sin\theta) \mathrm{d}x \mathrm{d}y
\end{split}
\end{equation}
where $\delta(\cdot)$ is the Dirac delta function. 

When the robot revisits the same place with different rotation $\alpha$ and translation $d \triangleq (\Delta x,\Delta y)^T$, SG can reflect such relative pose as follows:

\boldparagraph{Rotation}A rotation of $f(x, y)$ by a rotation angle $\alpha$ results in a circular shift along $\theta$ axis of $SG$:
\begin{equation}
\mathcal{R}_{f}(\theta, \tau) \stackrel{\alpha}{\longrightarrow} \mathcal{R}_{f}(\theta + \alpha, \tau)	
\label{rtor}
\end{equation}
A rotation leads to a uniform shift along $\theta$ axis of $SG$, which satisfies Definition~\ref{rotation equivariance} of rotation equivariance, arriving at the following result: 

\begin{lemma}
\label{lemma 1}
% Radon Transform $\mathcal{R}$ is a rotation-equivariant operation.
SG is rotation equivariant.
\end{lemma}

\boldparagraph{Translation}A translation of $f(x, y)$ by $d$ results in an angle-dependent shift of $\tau$ parameter on $SG$:
\begin{equation}
\mathcal{R}_{f}(\theta,\tau) \stackrel{d}{\longrightarrow} \mathcal{R}_{f}(\theta, \tau - (\cos\theta,\sin\theta)d)	
\label{rttr}
\end{equation}
Different from the rotation, translation leads to a non-uniform shift along $\tau$ axis of $SG$.

\boldparagraph{Pose}With both rotation $\alpha$ and translation $d$, $SG$ can be described by combining Eq.~\ref{rtor} and Eq.~\ref{rttr}: 
\begin{equation}
\label{sync}
\mathcal{R}_{f}(\theta, \tau) \stackrel{\alpha, d}{\longrightarrow} \mathcal{R}_{f}(\theta + \alpha, \tau - (\cos(\theta + \alpha), \sin(\theta + \alpha))d)
\end{equation}

\subsubsection{TING}
To further eliminate the coupled effect of translation on the $\tau$ axis of SG in Eq.~\ref{rttr}, we apply row-wise 1D DFT to SG along $\tau$ axis, and then calculate the magnitude of the resultant frequency spectrum. We name this representation as TING, denoted as $M_{f}(\theta, \omega)$, where $\omega$ is the sampled frequency in discrete the frequency spectrum.

\begin{lemma}
\label{lemma 2}
TING is rotation equivariant and translation invariant.    
\end{lemma}

\begin{proof}[Proof of Lemma~\ref{lemma 2}]
Suppose $M^{'}_{f}(\theta, \omega)$ is the TING representation constructed from transformed BEV $f{'}(x, y)$ by random translation $d$. Referring to the shift property of Fourier transform, we have:
% \begin{equation}
% \label{TING}
% \begin{aligned}
% {M}_{f}(\theta_{j}, \omega) &= |\mathcal{F}({R}_{f}(\theta_{j}, \tau))| \\
% &= |\mathcal{F}({R}_{f}(\theta_{j}, \tau - (\cos\theta_{j},\sin\theta_{j})d))| \\ 
% &= {M}_{f}(\theta_{j}, \omega)
% \end{aligned}
% \end{equation}
\begin{equation}
\label{TING}
\begin{aligned}
{M}^{'}_{f}(\theta_{j}, \omega) &= |\mathcal{F}({R}_{f}(\theta_{j}, \tau - (\cos\theta_{j}, \sin\theta_{j})d))| \\
&= |\mathcal{F}({R}_{f}(\theta_{j}, \tau)) e^{-i 2\pi w(\cos\theta_{j}, \sin\theta_{j})d}| \\
&= |\mathcal{F}({R}_{f}(\theta_{j}, \tau))| |e^{-i 2\pi w(\cos\theta_{j}, \sin\theta_{j})d}| \\ 
&= |\mathcal{F}({R}_{f}(\theta_{j}, \tau))| \\
&= {M}_{f}(\theta_{j}, \omega)
\end{aligned}
\end{equation}
where $|\cdot|$ is the operation of taking magnitude, $\mathcal{F}(\cdot)$ is the DFT operator, and $\theta_{j}$ means a row in SG and the same row in TING. Therefore, TING satisfies Definition~\ref{translation invariance} of translation invariance. 

As we only apply DFT to $\tau$ axis of SG, $\theta$ axis of TING keeps the same as that of SG, thus the rotation equivariance is reserved according to Lemma~\ref{lemma 1}. 
\end{proof}

\subsubsection{RING}
The final step of the representation pass is to eliminate the effect of rotation. According to Eq.~\ref{rtor}, note that the rotation only leads to the cyclic shift of $\theta$ axis of TING, we build the rotation-invariant representation in two steps. First, we employ a batch circular cross-correlation between TING $M_{Q}$ of query scan $P_{Q}$ and TING $M_{i}$ of every map scan $P_{i}$ in $\mathfrak{M}$, resulting in a batch of correlation maps as:
\begin{equation}
\begin{aligned}
    \mathfrak{C}_i(k_{\theta}, k_{\omega}) &= M_Q(\theta, \omega) * M_i(\theta, \omega)\\
    &= \sum_{\theta_j} \sum_{\omega_m} M_Q(\theta_j + k_{\theta}, \omega_m + k_{\omega}) M_i(\theta_j, \omega_m)
\end{aligned}
\label{2dcorr}
\end{equation}
where $\mathfrak{C}_i(k_{\theta}, k_{\omega})$ is the resultant 2D correlation map, $k_{\theta}$ and $k_{\omega}$ are the axes of the correlation map, and $*$ is the 2D cross-correlation operation between two images. 

Since TING is translation invariant by Lemma~\ref{lemma 2}, $\omega$ axis does not make any difference in 2D cross correlation calculation. Therefore, 2D cross correlation between two TINGs can be reduced to 1D cross correlation along $\theta$ axis, which is derived by substituting $k_{\omega} = 0$ into Eq.~\ref{2dcorr}:
\begin{equation}
\begin{aligned}
    \mathfrak{C}_i(k_{\theta}) &= M_Q(\theta, \omega) \star M_i(\theta, \omega)\\
    &= \sum_{\theta_j} \sum_{\omega_m} M_Q(\theta_j + k_{\theta}, \omega_m) M_i(\theta_j, \omega_m)\\
    &= \sum_{\theta_j} M_Q(\theta_j + k_{\theta}) M_i(\theta_j)^T
\end{aligned}
\label{corrcirc}
\end{equation}
where $\mathfrak{C}_i(k_{\theta})$ is the resultant 1D correlation map, $k_{\theta}$ is the axis of $\mathfrak{C}_i(k_{\theta})$, and $\star$ is the 1D circular cross-correlation that we derive from 2D one. Then, we take the max pooling on the correlation map $\mathfrak{C}_i$ of each TING pair $(M_{Q}, M_{i})$, comprising the final vector representation named RING $N_{Q} \in \mathbb{R}^{|\mathfrak{M}|}$:
\begin{equation}
\label{corr}
N_{Q,i} = \max_{k_{\theta}} \mathfrak{C}_i (k_{\theta})
\end{equation}
where $N_{Q,i}$ is the $i$th element of $N_{Q}$. Then we arrive at the first theorem:

\begin{theorem}
\label{theorem 1}
RING is roto-translation invariant. 
\end{theorem}

% Circular cross-correlation is a circular sliding inner-product in nature, similar to circular convolution operation, which is shift equivariant \cite{xxxxxx}. By adding a maximum operation to the resulted correlation matrix, shift invariance is achieved. According to Lemma \ref{lemma 1}, SG is rotation equivariant, whose column shift reflects rotation motion in the original image space. Thus, the shift invariance property of circular cross-correlation followed by maximum operation corresponds to rotation invariance of RING. 
\begin{proof}[Proof of Theorem~\ref{theorem 1}]
Applying the rotation $\alpha$ and translation $d$ to $P_{Q}$, we denote the resultant SG as $R^{'}_{Q}$ as Eq.~\ref{sync}:                                             
\begin{equation}
\label{rt-SG}
R^{'}_{Q}(\theta, \tau) = {R}_{Q}(\theta+\alpha, \tau - (\cos(\theta+\alpha),\sin(\theta+\alpha))d)
\end{equation}

Based on Lemma~\ref{lemma 2}, the TING $M^{'}_{Q}$ of transformed SG $R^{'}_{Q}$ is:
\begin{equation}
\label{rt-TING}
\begin{aligned}
M^{'}_{Q}&(\theta_{j}, \omega) = |\mathcal{F}({R}^{'}_{Q}(\theta_{j}, \tau))| \\
&= |\mathcal{F}({R}_{Q}(\theta_{j}+\alpha, \tau - (\cos(\theta_{j}+\alpha),\sin(\theta_{j}+\alpha))d))| \\ 
%&= {M}_{Q}(\theta_{j}+\alpha, \omega - (\cos(\theta_{j}+\alpha),\sin(\theta_{j}+\alpha))d) \\
&= {M}_{Q}(\theta_{j}+\alpha, \omega)
\end{aligned}
\end{equation}

The correlation map between $M^{'}_{Q}$ and $M_{i}$ is formulated as:
\begin{equation}
\label{rt-corr}
\begin{aligned}
    \mathfrak{C}^{'}_{i}(k_{\theta}) &= M^{'}_{Q}(\theta, \omega) \star M_i(\theta, \omega)\\
    &={M}_{Q}(\theta+\alpha, \omega)  \star M_i(\theta, \omega)\\
    &=\mathfrak{C}_i(k_{\theta}+\alpha)
\end{aligned}
\end{equation}

Based on Eq.~\ref{rt-corr}, the invariance of max value with respect to shift leads to the equality between RING $N_Q$ and RING $N^{'}_{Q}$ derived as:
\begin{equation}
N_{Q,i} = \max_{k_{\theta}} \mathfrak{C}_i(k_{\theta}) = \max_{k_{\theta}} \mathfrak{C}_i(k_{\theta}+\alpha) = N^{'}_{Q,i}
\end{equation}

Thus RING is roto-translation invariant.
\end{proof}

Now we summarize the representations in forward representation pass in brief: the rotation equivariant SG, the rotation equivariant and translation invariant TING, and the roto-translation invariant RING. RING of a scan stays the same even if large viewpoint changes are present.
% \begin{equation}

% c^{'}_{i} &= \max_\alpha (M_{i} \star M^{'}_{Q})[\alpha] \\
% &= \max_\alpha \sum_{\theta_{j}} {M}_{i}(\theta_{j}, \omega) \cdot {M}^{'}_{Q}(\theta_{j}+\alpha, \omega) \\
% &= \max_\alpha \sum_{\theta_{j}} {M}_{i}(\theta_{j}, \omega) \cdot {M}_{Q}(\theta_{j}+\beta+\alpha, \omega) \\

% \end{equation} 

% Let $\alpha^{'} = \alpha+\beta$ and substitute it into Eq. \ref{rt-corr}, and then we can find that
% \begin{equation}
% c^{'}_{i} = \max_{\alpha^{'}} \sum_{\theta_{j}} {M}_{i}(\theta_{j}, \omega) \cdot {M}_{Q}(\theta_{j}+\alpha^{'}, \omega) = c_{i} 
% \end{equation}

% Since each element $c^{'}_{i}$ in RING $N^{'}_{Q}$ under arbitrary transformation ($\beta$, $d$) is equal to the corresponding element $c_{i}$ in $N_{Q}$, we can conclude that
% \begin{equation}
% \label{RING}
% N^{'}_{Q} = N_{Q} 
% \end{equation}

\subsection{Solving Pass}

\subsubsection{Place Recognition}
Put Theorem~\ref{theorem 1} in a real scenario. Due to the finite scan range occlusion, the invariance of RING cannot be guaranteed when translation is significant with respect to the scan range occlusion. Therefore, we have RING that is invariant given an arbitrary rotation change, but gradually degenerated with respect to larger translation changes. Following this result, we can set the range of a place by measuring the change of RING. On the other hand, the density of the place for successful global localization is able to reflect the robustness against the difference between the trajectory in the query and mapping session.

Based on the difference between RINGs of query and map scans, we can retrieve the minimum one as the place $n$ where the query scan lies, arriving at place recognition: 
\begin{equation}
    n = \arg\min_i \|N_Q-N_{i}\|
    \label{prt}
\end{equation}

However, the dimension of RING is $|\mathfrak{M}|$, which is computed and stored for all $|\mathfrak{M}|$ map scans, resulting in total storage and nearest neighbor search of $O(|\mathfrak{M}|^2)$, which may not be affordable for onboard processing.

To reduce the computation, we introduce an approximation to Eq.~\ref{prt} by checking the maximum element in $N_Q$:
\begin{equation}
    n = \arg\max_i \tilde{N}_{Q,i}
    \label{pr}
\end{equation}
where $\tilde{N}$ is the normalized RING calculated from normalized TING $\tilde{M}$ with zero mean and unit variance. The aim of normalization is to eliminate the effect from the finite scan range e.g. number of valid scan points. In this way, we avoid the computation and storage of RINGs for all map scans. The equivalence between Eq.~\ref{prt} to Eq.~\ref{pr} is shown in Appendix~\ref{appendix proof 1}. We show that the prerequisite for Eq.~\ref{pr} is practical in real applications and hence is employed in the experiments.

% For fast implementation, we carry out circular cross-correlation efficiently by fast Fourier transform (FFT). Specifically, we multiply the Fourier transform of one TING by the complex conjugate of the Fourier transform of the other, and then take inverse Fourier transform (IFT) to get the correlation $c_{i}$, outputting roto-translation invariant representation RING, $N_{Q} \triangleq \{c_{i}\}$. 

% As the correlation value between query scan $P_{Q}$ and map scan $P_{i}$, $c_{i}$ actually indicates the similarity of pair $(P_{Q}, P_{i})$. After normalizing RING to make $c_{i} \in \{0, 1\}$, the feature space distance between $P_{Q}$ and $P_{i}$, $D(P_{Q}, P_{i})$, can be defined as
% \begin{equation}
% \label{distance}
% D(P_{Q}, P_{i}) = 1 - c_{i}
% \end{equation}

% Place recognition can be easily performed by comparing $D(P_{Q}, P_{i})$ with a pre-defined threshold $th$. If $D(P_{Q}, P_{i})$ is within the matching threshold $th$, then pair $(P_{Q}, P_{i})$ is considered a positive pair, otherwise a negative pair.  

% The maximum correlation value serves as a similarity metric between two scans. If it is higher than the matching threshold, then the two scans are considered to be collected in the same place.

\subsubsection{Pose Estimation}
After place recognition, we estimate the relative pose $T_{nQ}$ between $P_{Q}$ and $P_{n}$. Denote $f_{n}$ and $M_{n}$ as the BEV and TING of the retrieved map scan $P_{n}$ respectively. Given the local planar ground surface \cite{Lu_2019_CVPR} , the relative pitch, roll and height between query and map pose should be small. Therefore, we first estimate a reduced 3-DoF relative pose globally without initial value, comprising 1-DoF rotation $\alpha$ and 2-DoF planar translation $d$, then estimate the 6-DoF relative pose with the resultant 3-DoF relative pose as an initial value. To address the first step, we leverage TING to estimate rotation $\alpha$, which is then utilized to compensate BEV for estimation of translation $d$. For the second step, with the pose above as the initial value, we refine the metric pose using ICP \cite{besl1992method}.
% \ls{keep the form of dof consistent across the whole manuscript} 
% From Eq. \ref{sync}, the relationship of $R_{Q}$ and $R_{n}$ is described as
% \begin{equation}
% \label{transformation}
% \mathcal{R}_{n}(\theta, \tau) = \mathcal{R}_{Q}(\theta + \alpha, \tau - (\cos(\theta + \alpha), \sin(\theta + \alpha))d)
% \end{equation}

% The objective of this module is to acquire the metric pose of query scan $P_{Q}$, denoted by $T_{Q}$. The module can be divided into two parts: rotation estimation and translation estimation. 
% Coarse pose estimation is a by-product of aforementioned correlation-based place recognition, which provides an initial guess for fine pose estimation process. 

\boldparagraph{Rotation Estimation}Rotation estimation is achieved based on the TING. In Eq.~\ref{corr}, the maximum value occurs when $M_{Q}$ is equal to $M_{n}$. Given correct place recognition, as TING is translation invariant according to Lemma~\ref{lemma 2}, $M_Q$ and $M_n$ should be related by a relative rotation, and the equality is achieved when the shift $k_{\theta}$ equals to the real relative rotation.

Therefore, we regard the shift achieving the maximum value along $\theta$ axis as the estimation of relative rotation between $P_{Q}$ and $P_{n}$, denoted as $\hat{\alpha}$
\begin{equation}
\label{rotest}
\hat{\alpha} = \arg\max_{k_{\theta}} \tilde{\mathfrak{C}}_n (k_{\theta})
\end{equation}
where $\tilde{\mathfrak{C}}$ is the normalized correlation map calculated from the normalized TING $\tilde{M}$. The normalization is employed to relieve the effect of finite scan range and discretization in practice. It is important that the estimator be global convergent without dependency on any initial values. It is actually an exhaustive search, thus keeping the optimality. With the aid of fast Fourier transform (FFT) and parallel computing, this exhaustive process is fast. Refer to Appendix~\ref{appendix_proof_2} for derivation.

\boldparagraph{Translation Estimation}Based on the rotation estimation, we can rotate the BEV $f(x,y)$ by $-\hat{\alpha}$ to eliminate the relative rotation between $f_Q{(x,y)}$ and $f_n{(x,y)}$, yielding a compensated BEV ${f}^{'}_{n}{(x,y)}$. With the rotation variance eliminated, 2D cross correlation can be applied to solve the translation.
\begin{equation}
\mathfrak{C}_{n}(k_x,k_y) = f_Q(x,y)*{f}^{'}_{n}{(x,y)}
\label{phasecorrelation}
\end{equation}
where $\mathfrak{C}(k_x,k_y)$ is the correlation map of two BEVs. $k_{x}$ and $k_{y}$ are the axes of the correlation map. As for rotation, the maximum correlation in Eq.~\ref{phasecorrelation} should peak at the real relative translation. Denoting the estimated translation in correlation map as $\hat{d}$, we have
\begin{equation}
\hat{d} = \arg\max_{k_x, k_y} \mathfrak{C}_{n}(k_x,k_y)
\label{cross_correlation}
\end{equation}
Benefiting from the cross correlation, the translation estimation is globally convergent.

\boldparagraph{Refinement}With the rotation $\hat{\alpha}$ and translation $\hat{d}$ estimated above, an initial value is built by setting the relative pitch, roll and height with zeros. Theoretically, the estimation accuracy is up to the resolution of the BEV and Radon transform, thus keeping the convergence of the local refinement algorithm with high probability. In this paper, ICP is employed for refinement. Finally, we arrive at the 6-DoF pose $\hat{T}_{nQ}$, which is further applied to $T_n$ as the query pose estimation $\hat{T}_Q$. 

\subsection{Perspective of Feature Aggregation}

A typical pipeline \cite{jegou2010aggregating} for representation in place recognition consists of feature extraction and feature aggregation. In feature extraction, local features of sparse keypoints \cite{bay2006surf,rusu2009fast,rusu2010fast,sipiran2011harris}, or dense points \cite{liu2019lpd} are extracted, which is fed to the feature aggregation module to build a global feature for the scan, say global max pooling, generalized-mean pooling (GeM) \cite{radenovic2018fine}, or VLAD \cite{jegou2010aggregating, arandjelovic2013all} etc. 

\begin{figure*}[t]
	\centering
	\includegraphics[width=18cm]{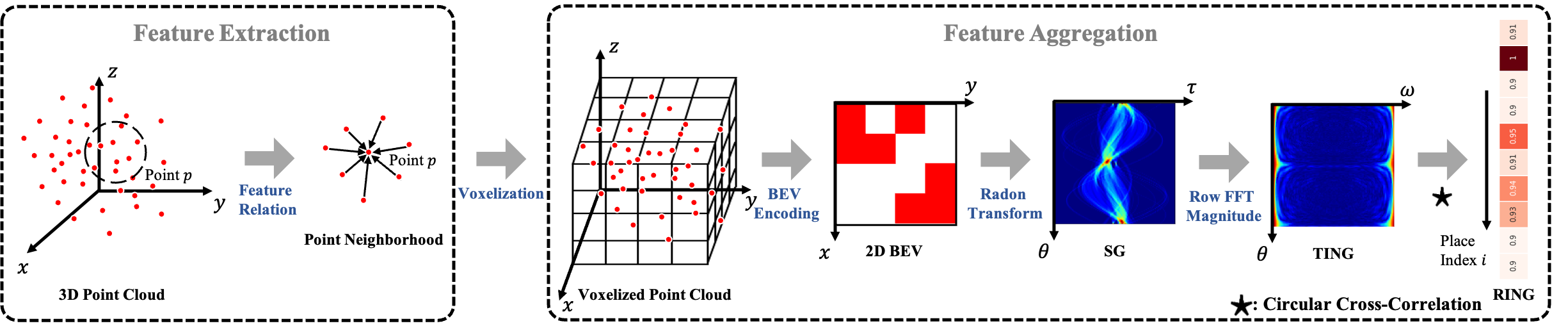}
	\caption{Pipeline of the single-channel framework of our method, including feature extraction and feature aggregation.}
	\label{fig:pipeline}
    \vspace{-0.5cm}
\end{figure*}

Our method also fits the pipeline, as illustrated in Fig.~\ref{fig:pipeline}. The feature extraction is achieved by occupancy BEV representation from raw scan. The feature aggregation is achieved by the representation pass. The advantage of the proposed feature aggregation is the representation property, which is roto-translation invariant for place recognition, and equivariant for pose estimation. As a general feature aggregation $g(\cdot)$, the feature extraction part can be switched to others, which is discussed in the next section.
 
% Furthermore, the features fused by RING aggregator are robust to synthetic transformations (rotation and translation) according to Lemma \ref{lemma 1}, Lemma \ref{lemma 2} and Theorem \ref{theorem 1}. The result of RING aggregator can be leveraged in various tasks such as image retrieval, image classification, etc., which acts as a roto-translation invariant place representation for place recognition task in our context. 

\section{Multi-channel Framework}
\label{multi-framework}
An intuitive improvement is to replace the simple binary BEV feature with one that takes more structured information in the height dimension. Naturally, we investigate the question: what are the requirements for feature extraction that is able to preserve the equivariance/invariance of SG, TING and RING?

% \subsection{Sufficient Condition}
\subsection{Feature Extraction}
\label{feat_extract}
We begin with the second theorem to present sufficient conditions for feature extraction:
\begin{theorem}
\label{theorem 2}
Let $E(\cdot)$ be an feature extractor, $g(\cdot)$ be a roto-translation invariant feature aggregation. Then the representation produced by $g(E(\cdot))$ preserves both translation and rotation invariance if one of the following requirements is satisfied:
\begin{itemize}
\item $E(\cdot)$ is equivariant to both translation and rotation.
\item $E(\cdot)$ is invariant to both translation and rotation.
\end{itemize}
% If the local features are translation/rotation-equivariant or rotation/translation invariant and the feature aggregator is translation/rotation-invariant, then the aggregated global features reserve translation/rotation invariance.
\end{theorem}
 
\begin{proof}[Proof of Theorem~\ref{theorem 2}]
\label{theorem 2 proof}
Let $P$ be a point cloud as the initial input of $E(\cdot)$. Utilizing the feature aggregation method $g(\cdot)$ to the features exploited by the feature extractor $E(\cdot)$, we can obtain the aggregated result $g(E(P))$. Since $g(\cdot)$ is invariant to both translation and rotation, $g(E(P) )  = g(E_{d}(P))$ by Definition~\ref{translation invariance} and $g(E(P))  = g(E_{\alpha}(P))$ by Definition~\ref{rotation invariance}.
Theorem~\ref{theorem 2} includes totally two cases, so we divide the proof into two sub-proofs corresponding to the two cases. 

\boldparagraph{Case 1} If $E(\cdot)$ is equivariant to both translation and rotation, then $E_{d}(P)  = E(P_{d})$ by Definition~\ref{translation equivariance} and $E_{\alpha}(P)  = E(P_{\alpha})$ by Definition~\ref{rotation equivariance}. 

Combining $E_{d}(P)  = E(P_{d})$ and $g(E(P) )  = g(E_{d}(P))$, we have:
\begin{equation}
\label{case1_trans}
g(E(P)) = g(E_{d}(P)) =  g(E(P_{d}))
\end{equation}

Combining $E_{\alpha}(P)  = E(P_{\alpha})$ and $g(E(P))  = g(E_{\alpha}(P))$, we have:
\begin{equation}
\label{case1_rot}
g(E(P)) = g(E_{\alpha}(P)) = g(E(P_{\alpha}))
\end{equation}

From Definition~\ref{translation invariance} and~\ref{rotation invariance}, we can conclude that the the final representation $g(E(P))$ is translation invariant and rotation invariant.

\boldparagraph{Case 2} If $E(\cdot)$ is invariant to both translation and rotation, then $E(P)  = E(P_{d})$ by Definition~\ref{translation invariance} and $E(P)  = E(P_{\alpha})$ by Definition~\ref{rotation invariance}. 

Combining $E(P)  = E(P_{d})$ and $g(E(P) )  = g(E_{d}(P))$, we have:
\begin{equation}
\label{case2_trans}
g(E_{d}(P)) = g(E(P)) = g(E(P_{d}))
\end{equation}

Combining $E(P)  = E(P_{\alpha})$ and $g(E(P))  = g(E_{\alpha}(P))$, we have:
\begin{equation}
\label{case2_rot}
g(E_{\alpha}(P)) = g(E(P)) = g(E(P_{\alpha}))
\end{equation}

From Definition ~\ref{translation invariance} and ~\ref{rotation invariance}, we can conclude that the the final representation $g(E(P))$ is translation invariant and rotation invariant.
\end{proof}

\begin{table}[t]
\renewcommand\arraystretch{1.5}
\centering
\caption{Extracted Features of RING++}
\label{feature}
\begin{threeparttable}
\begin{tabular}{lc}
\toprule
Feature & Formulation \\ \hline
Change of curvature & $\frac{\lambda_3}{\sum^{3}_{j=1}\lambda_j}$ \\ 
Omni-variance & $\frac{\sqrt[3]{{\textstyle \prod_{3}^{j=1}} \lambda_j}}{\sum^{3}_{j=1}\lambda_j}$ \\ 
Eigenvalue-entropy & $-{\sum^{3}_{j=1} (\lambda_j \ln{\lambda_j})}$ \\ 
2D linearity & $\frac{\lambda_{2D, 2}}{\lambda_{2D, 1}}$ \\ 
Maximum height difference & $Z_{max} - Z_{min}$ \\
Height variance & $\sum^{30}_{k=1} \frac{(Z_k - \frac{\sum^{30}_{k=1}{Z_{k}}}{30})^2}{30}$ \\ 
\bottomrule
\end{tabular}
\begin{tablenotes}
    \footnotesize
    \item[*] $\lambda_1 \geq \lambda_2 \geq \lambda_3 \geq 0$ represent the ordered eigenvalues of the symmetric positive-definite covariance matrix of the neighborhood.
\end{tablenotes}
\end{threeparttable}
\vspace{-0.3cm}
\end{table}

As stated in Theorem~\ref{theorem 1}, the proposed aggregation method $g(\cdot)$ possesses translation and rotation invariance, which satisfies the requirement of an aggregation method in Theorem~\ref{theorem 2}. Thus, we build rotation invariant and translation equivariant features to preserve the property of the aggregation. We select six rotation invariant translation equivariant features from previous literature\cite{weinmann2014semantic} shown in Tab.~\ref{feature}. Note that popular features like Harris3D \cite{sipiran2011harris}, SIFT3D \cite{scovanner20073}, FPFH \cite{rusu2009fast} and SHOT \cite{tombari2011combined} can also fit in the framework, some of which are leveraged for ablation study in Section~\ref{ablation}. Specifically, the local feature is built in two steps. First, for each point $p$ in the scan, we extract features from its $30$-neighborhood as shown in Tab.~\ref{feature}. Then we accumulate the point features along the dimension of height. For all point features belonging to the same BEV grid, we use the channel-wise max pooling, resulting in a 6-channel BEV, $f(x,y) \in \mathbb{R}^6$. Denote each channel of $f(x,y)$ as $f_c(x,y) \in \mathbb{R}$, where $c \in \{1,2,3,4,5,6\}$ indicates the channel dimension. Compared with the occupancy, this feature better encodes the information in the height dimension such as the building facade.
% \begin{itemize}
% \item Change of curvature: $C_{p} = \frac{\lambda^{p}_{3}}{\sum^{3}_{j=1}\lambda^{p}_{j}}$
% \item Omni-variance: $O_{p} = \frac{\sqrt[3]{{\textstyle \prod_{3}^{j=1}} \lambda^{p}_{j}}}{\sum^{3}_{j=1}\lambda^{p}_{j}}$
% \item Eigenvalue-entropy: $A_{p} = -{\sum^{3}_{j=1} (\lambda^{p}_{j} \ln{\lambda^{p}_{j}})}$
% \item 2D linearity: $L_{p, 2D} = \frac{\lambda^{p}_{2D, 2}}{\lambda^{p}_{2D, 1}}$
% \item Maximum height difference: $\Delta Z_{p, max}$
% \item Height variance: $\sigma Z_{p, var}$
% \end{itemize}

%Then we have
% \begin{equation}
% f(x,y) = f_1(x,y) \oplus f_2(x,y) \oplus ... \oplus f_6(x,y)
% \end{equation}
% where $\oplus$ is the concatenation operation. 

After feature extraction, we embark on feature aggregation, following the same pipeline demonstrated in Section~\ref{framework} to represent. One remaining problem is to aggregate the multi-channel feature into one for solving.

\begin{figure}[tp]
	\centering
	\includegraphics[width=9cm]{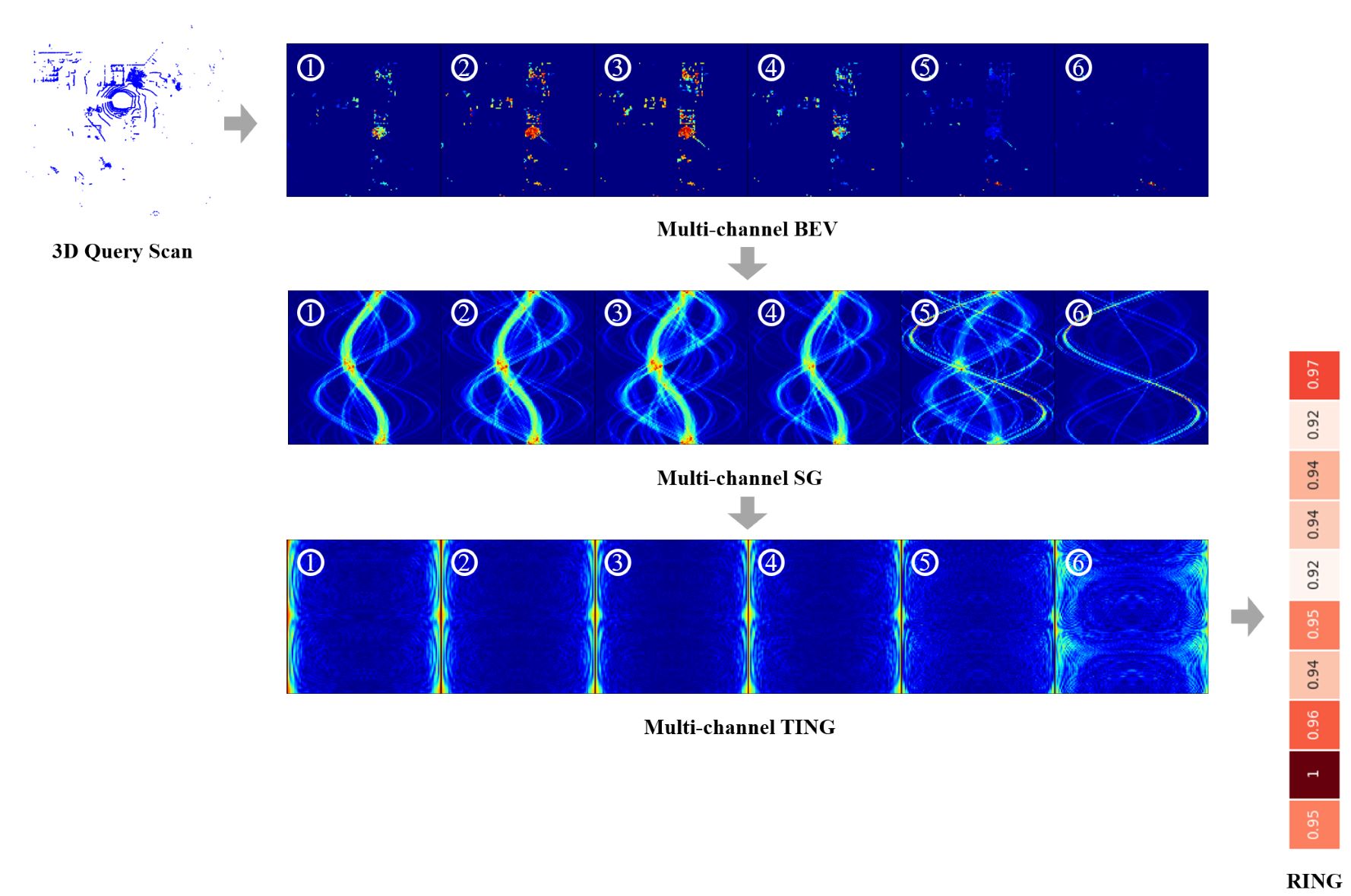}
	\caption{Visualization of the multi-channel representation pass. With six extracted local features, the corresponding BEV, SG and TING are 6-channel.}
	\label{fig:multi_representation}
    \vspace{-0.2cm}
\end{figure}

\subsection{Multi-channel Representation Pass}
% \ls{some formulas} In the representation pass, each kind of local feature of the point cloud is encoded to a 2D BEV channel, yielding a multi-channel BEV $f(x,y)$. 
For clear demonstration, we visualize the multi-channel representation pass in Fig.~\ref{fig:multi_representation}. Based on the 6-channel BEV $f(x,y)$, we apply Radon transform to each BEV channel $f_c(x,y)$, yielding a SG channel $R_{f_c}(\theta, \tau)$. By concatenating all $R_{f_c}(\theta, \tau)$ along  channel  dimension $c$, we can obtain a 6-channel SG $R_{f}(\theta, \tau)$.

% \begin{equation}
% R_{f}(x,y) = R_{f_1}(x,y) \oplus R_{f_2}(x,y) \oplus ... \oplus R_{f_6}(x,y)
% \end{equation}

After that, we employ row-wise DFT to $R_{f_c}(\theta, \tau)$ and take the magnitude spectrum of the result, denoted as $M_{f_c}(\theta, \omega)$, to achieve translation invariance. In the same manner, we concatenate all $M_{f_c}(\theta, \omega)$ along $c$ dimension, yielding a 6-channel TING $M_f(\theta, \omega)$. 
% \begin{equation}
% M_{f}(x,y) = M_{f_1}(x,y) \oplus M_{f_2}(x,y) \oplus ... \oplus M_{f_6}(x,y)
% \end{equation}

Next we implement 1D circular cross-correlation between query TING $M_{Q}(\theta, \omega)$ and all map TINGs $M_{i}(\theta, \omega)$. For each channel TING pair $(M_{Q_c}(\theta, \omega), M_{i_c}(\theta, \omega))$, we can get a 1D correlation map $\mathfrak{C}_{i_c}(k_{\theta})$:

\begin{equation}
\begin{aligned}
    \mathfrak{C}_{i_c}(k_{\theta})  &= M_{Q_c}(\theta, \omega) \star M_{i_c}(\theta, \omega) \\
    &= \sum_{\theta_j} M_{Q_c}(\theta_j + k_{\theta}) M_{i_c}(\theta_j)^T
\end{aligned}
\label{multicorrcirc}
\end{equation}
Then we sum all $\mathfrak{C}_{i_c}(k_{\theta})$ along $c$ dimension, generating a single-channel 1D correlation map $\mathfrak{C}_{i}(k_{\theta})$: 
\begin{equation}
 \mathfrak{C}_{i}(k_{\theta}) = \sum^{6}_{c=1} \mathfrak{C}_{i_c}(k_{\theta})
\end{equation}
Finally we take the maximum value of $\mathfrak{C}_{i}(k_{\theta})$ as the $i$th element of the final representation RING $N_{Q}$, which is the same as Eq.~\ref{corr}. 

\subsection{Multi-channel Solving Pass}
In the solving pass, we perform place recognition with roto-translation invariant global descriptor RING $N_{Q}$ first. Utilizing the same operation in Eq.~\ref{pr}, we retrieve the closest map scan $P_{n}$ from $\mathfrak{M}$ by finding the index of the maximum element in $\tilde{N}_{Q}$. 

Then we carry out pose estimation including three steps: rotation estimation, translation estimation and refinement. Taking advantage of the summed correlation map $\mathfrak{C}_{n}(k_{\theta})$ between the query TING $M_{Q}$ and the retrieved map scan TING $M_{n}$, the relative rotation can be easily estimated by Eq.~\ref{rotest}, which is the optimal shift when $M_{Q}$ best aligns with $M_{n}$. To solve the relative translation $d$, we begin with compensating the BEV $f_{Q}(x,y)$ via rotating $f_{n}(x,y)$ by $-\hat{\alpha}$. Under the multi-channel framework, we first calculate the correlation map of each channel $\mathfrak{C}_{n_c}(k_x,k_y)$ using Eq.~\ref{phasecorrelation} and then sum the correlation maps along the channel dimension to obtain the globally optimal translation $\hat{d}$:
\begin{equation}
\hat{d} =  \arg\max_{(k_x,k_y)} \sum^{6}_{c=1} \mathfrak{C}_{n_c}(k_x,k_y)
\label{multitau}
\end{equation}

% Under the multi-channel framework, the relationship between the shifted $R^{'}_{n_{c}}$ and $R_{Q_{c}}$ is described as:
% \begin{equation}
% {R}^{'}_{n_c}(\theta, \tau) \triangleq {R}_{Q_c}(\theta, \tau - (\cos(\theta + \hat{\alpha}), \sin(\theta + \hat{\alpha}))d)
% \end{equation}
% Then we follow Eq.~\ref{lstau} to build a linear equation. Different from Eq.~\ref{optimaltau}, as $R_{Q}$ and $R^{'}_{n}$ are multi-channel, we sum the correlation map of each channel to find the optimal shift $\Delta \hat{\tau}_{j}$:
% \begin{equation}
% \Delta \hat{\tau}_{j} = \arg\max_{k_{\tau}} \sum^{6}_{c=1} \sum_{\tau_{m}} {R}^{'}_{n_c}(\theta_{j}, \tau_{m}+k_{\tau}){R}_{Q_c}(\theta_{j}, \tau_{m})
% \label{multitau}
% \end{equation}
% By stacking the equation into a linear system as Eq.~\ref{les}, we solve the translation $\hat{d}$ with globally. 

The refinement for multi-channel framework is exactly the same as the single-channel one. The estimated 3-DoF relative transformation between $P_{Q}$ and $P_{n}$ is utilized as the initial value for ICP to acquire the refined 6-DoF pose $\hat{T}_{Q}$.

% The generated global descriptor RING by aggregating local features is more robust and discriminative since it contains sufficient neighbors information. 
% assign the maximum value of these local features of points within a bin as the encoding function to project a 3D LiDAR scan to a 2D BEV image. 

% To elaborate place retrieval with constructed RING. The specific procedures are developed in detail in Section \ref{framework}. Row-wise 1D DFT is applied on multi-layer RING to achieve translation invariance, generating  a translation-invariant representation, multi-layer TING. We employ circular cross-correlation for each feature layer of multi-layer TING, and concatenate each feature channel through summation operation. Then we can get the maximum correlation and corresponding cyclic shift for each pairwise correlation computation. For fair comparison, normalize the maximum correlation value in [0, 1] range and set a revisited threshold denoted by $d$. 

\ifx\allfiles\undefined
\end{document}
\fi
\ifx\allfiles\undefined

\begin{document}
\fi

\section{Dataset and Evaluation Criteria}
In this section, we describe the datasets and evaluation criteria that we employ for performance validation. 

\subsection{Dataset}
We perform our RING++ method on three widely used datasets for place recognition evaluation: NCLT \cite{carlevaris2016university}, MulRan \cite{kim2020mulran}, and Oxford \cite{barnes2020oxford} datasets. In order to validate the translation and rotation invariance of our approach, we select several trajectories with large translation and rotation changes. Furthermore, we utilize different sequences for multi-session place recognition evaluation. The characteristics of these sequences are detailed in the subsections. 

\subsubsection{NCLT Dataset}
A large-scale and long-term dataset collected by a Segway robot on the University of Michigan's North Campus. It contains 27 different sessions that captured biweekly from January 8, 2012 to April 5, 2013. Covering the same trajectory over 15 months, the dataset includes a large variety of environmental changes: dynamic objects like moving people, seasonal changes like winter and summer, and structural changes like the construction of buildings.
\subsubsection{MulRan Dataset}
A multimodal range dataset containing Radar and LiDAR data specially collected in the urban environment. It covers four different target environments: DCC, KAIST, Riverside, and Sejong City, providing both temporal and structural diversity for place recognition research.
\subsubsection{Oxford Radar RobotCar Dataset}
An extension to the \textit{Oxford RobotCar Dataset} \cite{maddern20171} for autonomous driving research. The data comprises 32 traversals of a central Oxford route in January 2019. A variety of weather and lighting conditions are encompassed in this dataset. A pair of Velodyne HDL-32E 3D LiDARs are mounted on the left and right sides of the vehicle to improve 3D scene understanding performance. For the convenience of place recognition evaluation, we concatenate point clouds collected by these two LiDARs into one single scan.

\subsection{Evaluation Metrics}
\subsubsection{Recall@1}
We utilize \textit{Recall@1} \cite{raghavan1989critical} metric to evaluate a place recognition system in terms of the number of true loop candidates. \textit{Recall@1} is defined as the ratio of top 1 true positives to total positives, which is formulated as
% \textit{Recall@1$\%$} is defined as the ratio of top 1$\%$ true positives to total positives. The mathematical expressions are
\begin{equation}
Recall@1 = \frac{TP_{top1}}{TP_{top1} + FN_{top1}}
\end{equation}
% \begin{equation}
% Recall@1\% = \frac{TP_{top1\%}}{TP_{top1\%} + FN_{top1\%}}
% \end{equation}
where $TP_{top1}$ denotes the number of true positives with top 1 retrieval, $FN_{top1}$ denotes the number of false negatives with top 1 retrieval.
% $TP_{top1\%}$denotes the number of true positives with top 1$\%$ retrievals, and $FN_{top1\%}$ denotes the number of false positives with top 1$\%$ retrievals.

\subsubsection{Precision-Recall Curve}
For place recognition task, \textit{precision} \cite{raghavan1989critical} is defined as the ratio of true positives to total matches, and \textit{recall} \cite{raghavan1989critical} is defined as the ratio of true positives to total positives. The mathematical expressions are
\begin{equation}
Precision = \frac{TP}{TP + FP}
\end{equation}

\begin{equation}
Recall = \frac{TP}{TP + FN}
\end{equation}
where \textit{TP} denotes the number of true positives, \textit{FP} denotes the number of false positives, and \textit{FN} denotes the number of false negatives.
\textit{Precision-Recall Curve} \cite{christopher1999foundations} is plotted under various thresholds. A point in the \textit{Precision-Recall Curve} depicts the \textit{precision} and \textit{recall} values corresponding to a specific threshold.

\subsubsection{F1 score-Recall Curve }
\textit{F1 score}\cite{schutze2008introduction} is the harmonic mean of {precision} and {recall}, combining {precision} and {recall} metrics into a single metric:
\begin{equation}
F1 \; score = \frac{2 \times Precision \times Recall}{Precision + Recall}
\end{equation}
which is a suitable metric to balance \textit{precision} and \textit{recall}.

\subsubsection{TE and RE}
We calculate average translation error \textit{TE} and rotation error \textit{RE}  following \cite{komorowski2021egonn, cattaneo2021lcdnet} for pose estimation evaluation. Since our method only yields 3-DoF poses, we evaluate 2-DoF translation and 1-DoF rotation error in the majority of evaluations. The mathematical formulas are
\begin{equation}
TE = \Vert \hat{d}-d^*\Vert
\end{equation}

\begin{equation}
% RE = \arccos{((Tr(\hat{R}^TR^*)-1)/2)}
RE = |\hat{\alpha} - \alpha^*|
\end{equation}
where $\hat{d}$ and $\hat{\alpha}$ are the estimated 2-DoF translation and 1-DoF rotation. $d^*$ and $\alpha^{*}$ are the ground-truth translation and rotation. Pose error calculation of incorrectly matched scans (i.e. not from the same place) is meaningless. Thus, we only compute the estimated pose errors of successfully matched pairs to compare the pose estimation performance.

\subsubsection{Success Rate}
To quantitatively evaluate the overall global localization performance, we leverage \textit{Success Rate} \cite{komorowski2021egonn, cattaneo2021lcdnet} metric which takes \textit{Recall@1}, \textit{TE} and \textit{RE} into account, with both place recognition and pose estimation involved. Following the definition of \textit{Success Rate} in \cite{komorowski2021egonn, cattaneo2021lcdnet}, \textit{Success Rate} is calculated by the percentage of successfully aligned matches whose \textit{TE} is below $2m$ and \textit{RE} is below $5^\circ$. The mathematical formula is
\begin{equation}
Success  \; Rate = \frac{TP_{TE <2 m \; \& \; RE < 5^\circ}}{TP + FP}
\end{equation}
where $TP_{TE <2 m \; \& \; RE < 5^\circ}$ denotes the number of true positives whose translation error is below $2m$ and rotation error is below $5^\circ$.

\subsubsection{ATE}
Absolute trajectory error (\textit{ATE}) is used to evaluate the performance of SLAM systems. It directly measures the difference between points of the true and the estimated trajectory.  The \textit{ATE} of $i_{th}$ frame is formulated as below
\begin{equation}
{ATE}_i = \Vert trans(Q_i^{-1}SP_i) \Vert
\end{equation}where $Q_i$ is the ground truth pose and $P_i$ is the estimated pose. $S\in SE(3)$ is the transfomation matrix to align the two poses, which is calculated by the least square method. $trans(\cdot)$ represents the translation part of the pose difference. In this paper, we use the average \textit{ATE} to evaluate the performance of SLAM systems.

\subsection{Comparative Methods}

In terms of place recognition and pose estimation, we compare our approach against state-of-the-art learning-free and learning-based methods:

\begin{itemize}
\item \textbf{M2DP} \cite{he2016m2dp} projects the point cloud to multiple 2D planes and leverages the singular vector of all signatures for points in each plane as the global descriptor to detect loop closure.
\item \textbf{Fast Histogram} \cite{rohling2015fast} utilizes the histogram of range distance directed extracted from the 3D point cloud as the global descriptor of the point cloud for loop closure detection.
\item \textbf{Scan Context} \cite{kim2018scan} encodes the 3D scan to a representation called Scan Context which contains 2.5D information, compared to histogram-based methods. It is invariant to rotation changes via a two-phase search algorithm.
\item \textbf{Scan Context++} \cite{kim2021scan} introduces two sub-descriptors to combine topological place recognition with 1-DoF semi-metric localization, so as to enhance the performance of Scan Context.
\item \textbf{PointNetVLAD} \cite{uy2018pointnetvlad} combines PointNet and NetVLAD to extract the global descriptor for a 3D point cloud by end-to-end training.
\item \textbf{DiSCO} \cite{xu2021disco} encodes a 3D scan to a scan context and then uses an encoder-decoder network to extract features. It constructs a rotation invariant place descriptor by taking the magnitude of the frequency spectrum for place recognition and designs a correlation-based rotation estimator for 1-DOF pose estimation.
\item \textbf{EgoNN} \cite{komorowski2021egonn} designs a fully convolutional architecture to extract global and local descriptors. It uses the global descriptor for coarse place recognition and the local descriptors for 6-DoF pose estimation.
\end{itemize}

% \subsection{Comparative Methods of Pose Estimation}
% In terms of the pose estimation, our proposed RING++ estimates a 3-DoF pose (yaw angle and translation along XY plane), while other methods like Scan Context++ \cite{kim2018scan} and DiSCO \cite{xu2021disco} only provide 1-DoF pose (yaw angle or lateral translation). For fair comparison with these approaches, we compare RING++ with these approaches in terms of the pose after point cloud registration by ICP \cite{xxxxxx}, and the above-mentioned place recognition methods followed by ICP. 

\subsection{Our Methods}
In Section~\ref{framework} and~\ref{multi-framework}, we solve place recognition problem utilizing RING representation constructed from TING and estimate the relative rotation based on TING. To validate the translation invariance of RING generated by TING, we directly construct RING representation based on SG representation skipping TING formation process for place recognition, and then estimate the relative rotation and translation based on SG, which serves as a variant of our method. Taking feature extraction into account, we then have four versions of our method: RING (SG), RING, RING++ (SG), and RING++. By comparing the four versions, we can figure out the effects of translation invariance design and feature extraction on place recognition and pose estimation results.
% Besides TING representation, SG representation can also be utilized for RING construction and rotation estimation. 
% The first two versions, RING (SG) and RING, encode occupancy information of a point cloud to different representations. The last two versions, RING++ (SG) and RING++, extract multiple local features, following a multi-channel framework. RING (SG) and RING++ (SG) construct RING representation based on SG without TING formation. RING and RING++ generate RING representation based on TING, which achieves translation invariance.
\begin{itemize}
\item \textbf{RING (SG)} encodes occupancy information of a point cloud to SG and RING representations. It utilizes RING representation for place recognition, SG representation for rotation estimation, and single-channel BEV representation for translation estimation.
\item \textbf{RING} encodes occupancy information of a point cloud to SG, TING and RING representations. It leverages RING representation for place recognition, TING representation for rotation estimation, and single-channel BEV representation for translation estimation.
\item \textbf{RING++ (SG)} extracts multiple local features of a point cloud, following a multi-channel framework. It utilizes RING representation for place recognition, SG representation for rotation estimation, and multi-channel BEV representation for translation estimation.
\item \textbf{RING++} extracts multiple local features of a point cloud, following a multi-channel framework. It leverages RING representation for place recognition, TING representation for rotation estimation, and multi-channel BEV representation for translation estimation.
\end{itemize}

\subsection{Implementation Details}
For a fair comparison, we remove the ground plane of the raw point cloud and filter it to the same range $[-70m, 70m]$ for all methods. For both Scan Context and our method, we set the grid size of BEV to $120 \times 120$, so the translation resolution is $140/120 = 1.17m/pixel$ and the rotation resolution is $360/120 = 3^{\circ}/pixel$. The number of candidates of Scan Context++ is 10. The parameters and settings of other compared methods are kept the same as those in the original papers.

PointNetVLAD and DiSCO are retrained on the MulRan dataset. The output dimension of these two methods is set to $1024$. In training step, places within $10m$ are regarded as positive pairs, while negative pairs are at least $20m$ apart. Other parameter configurations are similar to those in the original paper. For EgoNN, we use the publicly available pre-trained model for better performance. The output global descriptor dimension of EgoNN is $256$, while the local descriptor dimension is $128$. We use ICP implementation from Open3D \cite{Zhou2018} in the experiments with pose refinement. The parameters of ICP are set according to common practice: \textit{max correspondence distance} is 1.5m and \textit{iterations} is 100.

\ifx\allfiles\undefined
\end{document}
\fi
\ifx\allfiles\undefined

% \usepackage{xr}
% \externaldocument{4-method.tex}
\begin{document}
\fi

\section{Experimental Evaluation}
In this section, we design some experiments to verify that the proposed approach:
\begin{itemize}
\item has strong translation invariance and rotation invariance that is independent of translational difference.
\item detects the loops successfully when the pose difference between query and map point clouds is large.
\item estimates a 3-DoF pose as a qualified initial guess for further metric refinement (ICP alignment).
\item is computation-efficient with a compact representation for real-time applications.
\item is easily pluggable into SLAM systems for loop closure detection and relocalization.
\end{itemize}

\subsection{Illustrative Toycase Study}

We present two toycases to validate the effectiveness of our approach in terms of global localization. Specifically, we design a toycase to verify the roto-translation invariance of our representation RING for place recognition. Moreover, we investigate the impact of translation motion on relative rotation estimation utilizing various representations (SC (Scan Context), SG and TING), in order to highlight the advantage of our representation for pose estimation.

\subsubsection{Place Recognition Illustration}
\begin{figure}[tbp]
	\centering
    \includegraphics[width=9cm]{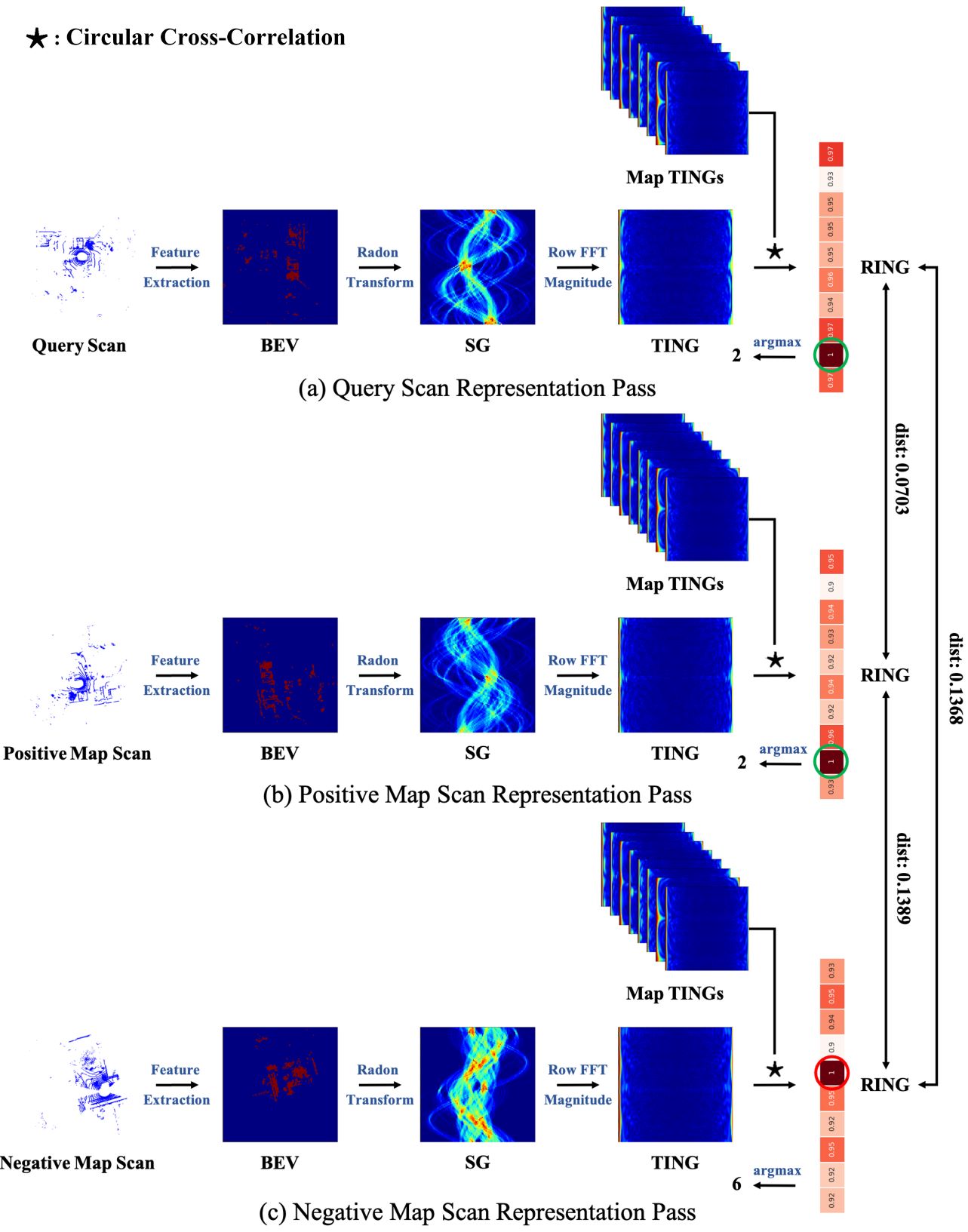}
	\caption{Visualization of the representation pass for query scan, positive map scan and negative map scan, to verify the feasibility of utilizing Eq.~\ref{pr} to equivalently replace Eq.~\ref{prt} for place recognition.
% 	the maximum value of RING for place recognition, which leads to the same result with Euclidean distance. 
	(a) The maximum value of RING representation points to the second scan in the scan map, which is the positive map scan shown in (b). (b) The RING representation of the positive map scan is the closest to the RING representation of query scan in terms of Euclidean distance. (c) The Euclidean distance between query RING and negative map RING is nearly twice of that between query RING and positive map RING.}
	\label{fig:case study 1}
	\vspace{-0.4cm}
\end{figure}

We begin with a toycase study to illustrate the feasibility of our method for place recognition. In this toycase, we perform place retrieval in a scan map with $10$ map scans for simplification. To better illustrate the proposed method, we visualize all the intermediate representations, as shown in Fig.~\ref{fig:case study 1}. The top block (a) of Fig.~\ref{fig:case study 1} shows that the query scan is located at the same place where the second map scan lies according to Eq.~\ref{pr}. The retrieved map scan, i.e., the second map scan follows the representation pass which is depicted in the bottom block (b) of Fig.~\ref{fig:case study 1}. Furthermore, the Euclidean distance between query RING and positive map RING is the smallest, which satisfies Eq.~\ref{prt}, and is also consistent with the retrieved result above by Eq.~\ref{pr}.

% \begin{figure}[htbp]
% 	\centering
%     \subfloat[Query Scan Representation Pass]{
% 		\includegraphics[width=8cm]{figs/case_query.png}}
%     \subfloat[Retrieved Scan Representation Pass]{
% 		\includegraphics[width=8cm]{figs/case_map.png}}
% 	\caption{Visualization of the representation pass for query scan and the top1 retrieved scan. (a) The maximum value of RING representation points to the second scan in the scan map, which is the retrieved scan shown in (b). (b) The RING representation of the retrieved scan is the closest to the RING representation of query scan in terms of Euclidean distance. }
% 	\label{fig:case study 1}
%     \vspace{-0.5cm}
% \end{figure}

\begin{figure}[htbp]
	\centering
    \subfloat[SC]{
		\includegraphics[width=8.8cm]{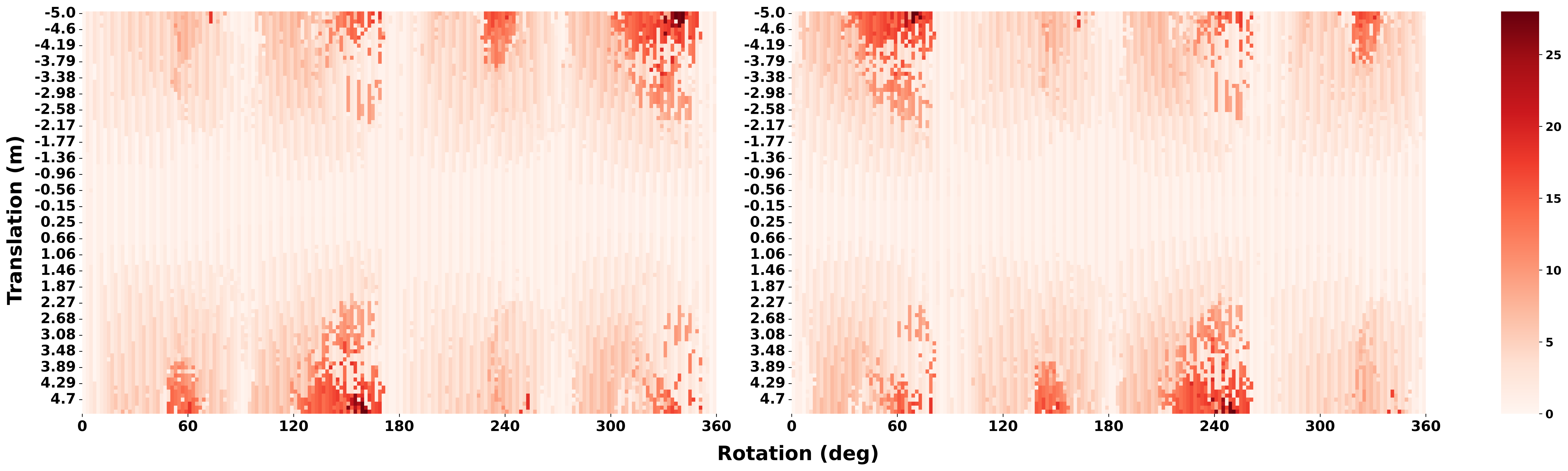}}\\
    \subfloat[SG]{
		\includegraphics[width=8.8cm]{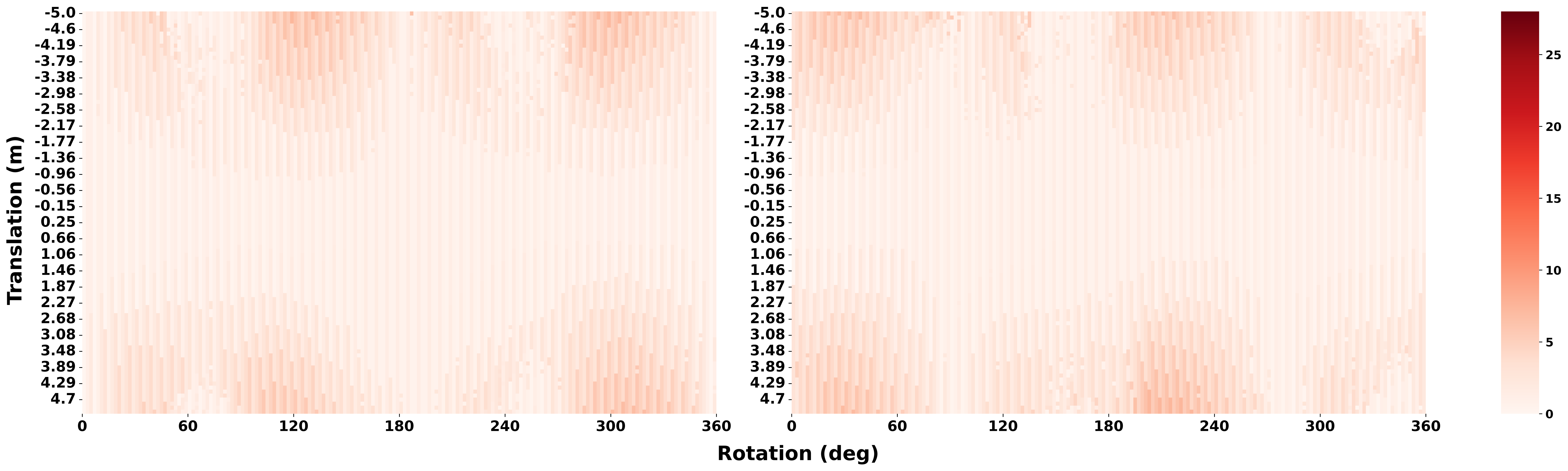}}\\
    \subfloat[TING]{
		\includegraphics[width=8.8cm]{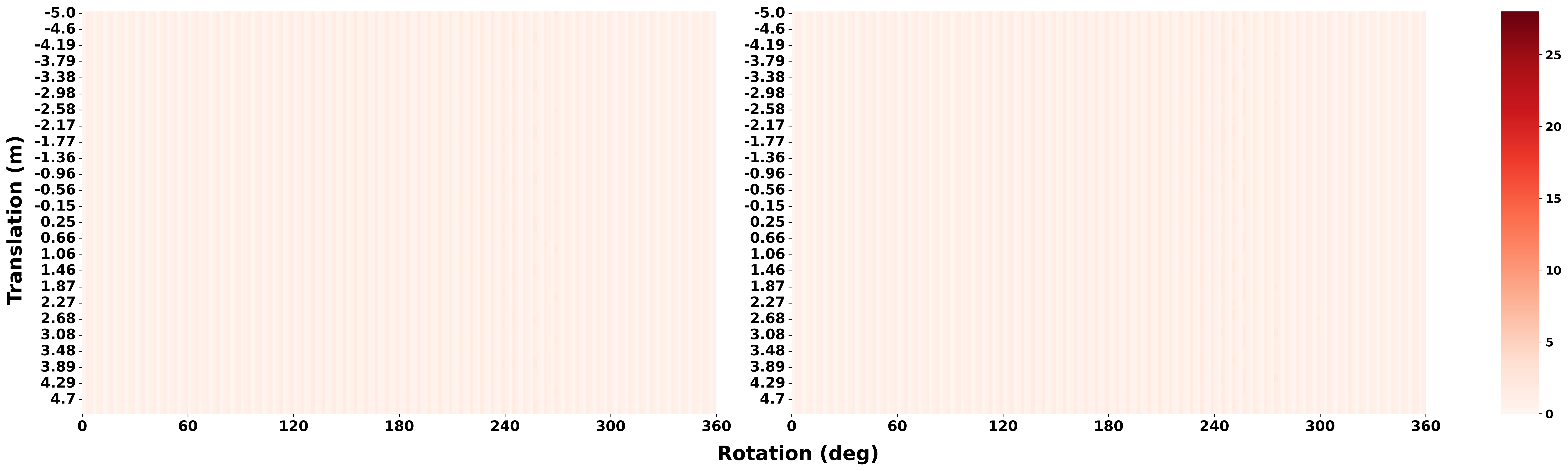}}        
	\caption{Rotation estimation error using different representations. The left column shows the rotation estimation error with translation along the x-axis, and the right column shows the rotation estimation error with translation along the y-axis.}
	\label{fig:case_study_2}
	\vspace{-0.4cm}
\end{figure}

\subsubsection{Pose Estimation Illustration}
Similar to Scan Context series \cite{kim2018scan,kim2021scan}, we leverage rotation-equivariant representations SG and TING to estimate the relative rotation, referring to rotation equivariance illustrated in Lemma~\ref{lemma 1} and~\ref{lemma 2}. To distinguish our representation from others, we rotate the point cloud within $[0^{\circ}, 360^{\circ})$ and translate the point cloud within $[-5 m, 5 m]$. After that, we employ the three aforementioned representations to estimate the relative rotation between the original point cloud and the transformed point cloud. The resolution of all representations for rotation estimation is the same, $3^{\circ}/pixel$. As we can easily see in Fig.~\ref{fig:case_study_2}, SC representation is strongly affected by translation difference. When the relative translation is beyond $1m$, the rotation estimation error rises to about $30^{\circ}$. In contrast, our representations SG and TING are much more robust to translation perturbation, showing smaller rotation estimation error. In addition, TING equipped with translation invariance is almost not influenced by translation disturbance, whose rotation estimation error keeps nearly $0^{\circ}$ for all translation within $5m$. 

\subsection{Evaluation of Place Recognition}

For comprehensive place recognition evaluation, we evaluate the proposed method regarding both online loop closure detection (single-session scenarios) and long-term localization (multi-session scenarios). 

\subsubsection{Single-session Scenarios}
\begin{figure}[tbp]
	\centering
		\includegraphics[width=9cm]{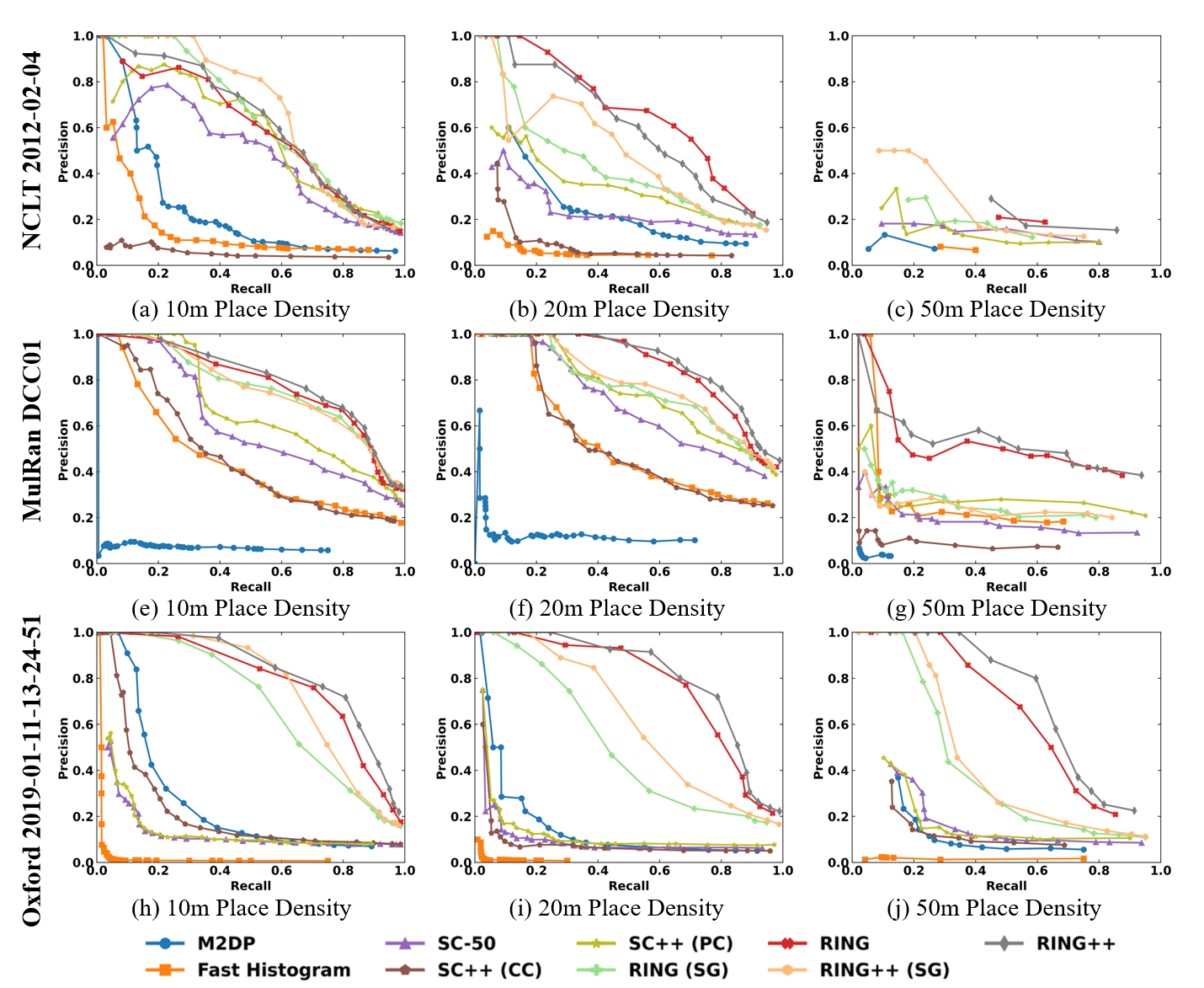}
	\caption{Precision-Recall Curve for NCLT, MulRan and Oxford datasets under single-session scenarios.}
	\label{fig:single_PR}
    \vspace{-0.3cm}
\end{figure}

\begin{figure}[tbp]
	\centering
		\includegraphics[width=9cm]{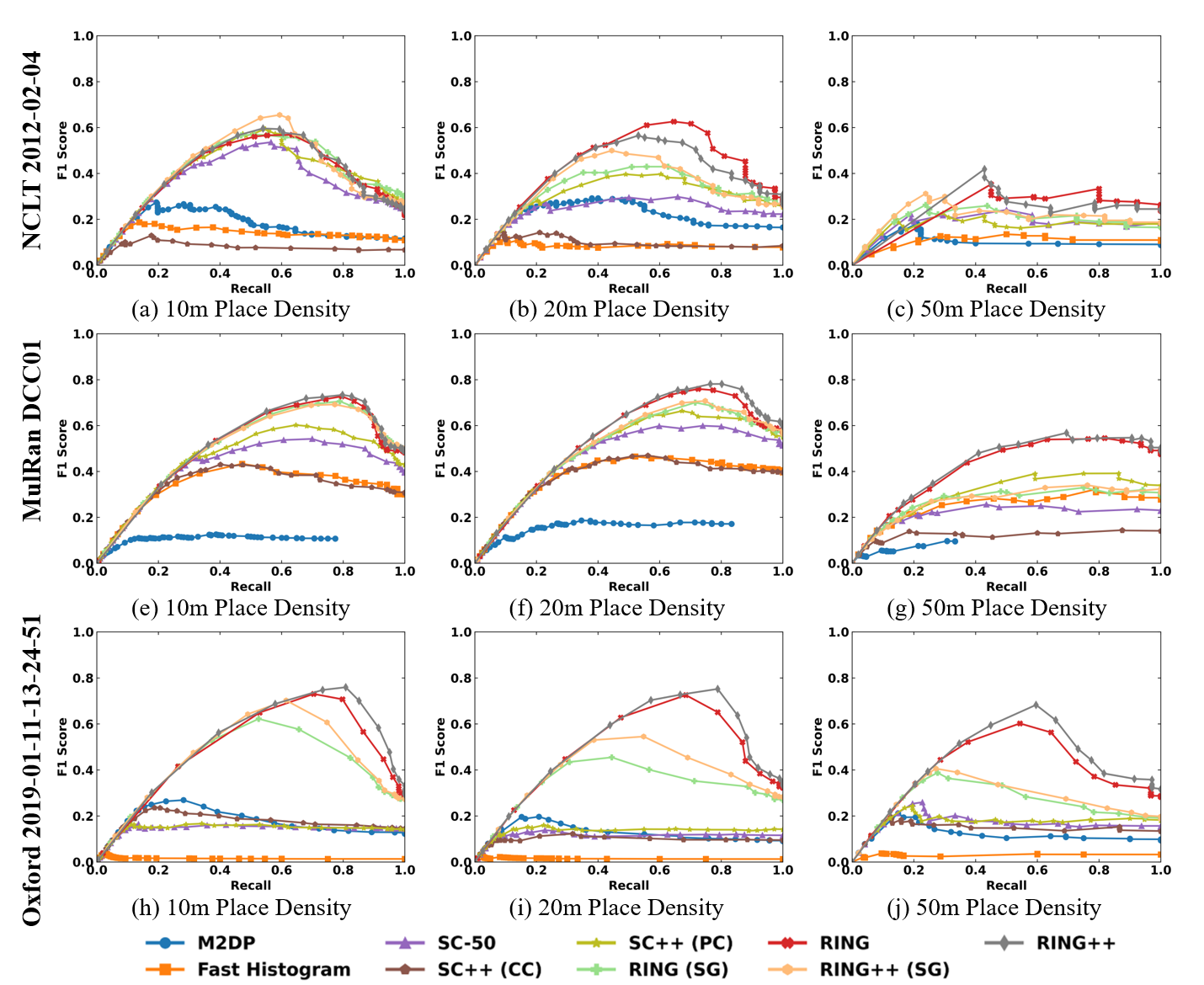}
	\caption{F1 score-Recall Curve for NCLT, MulRan and Oxford datasets under single-session scenarios.}
	\label{fig:single_F1}
    \vspace{-0.3cm}
\end{figure}

Under single-session scenarios, we select ``2012-02-04" sequence in NCLT dataset, ``DCC01" sequence in MulRan dataset, and ``2019-01-11-13-24-51" sequence in Oxford dataset for place recognition evaluation. To verify the advantage of our approach in the sparse scan map, we perform all methods to detect loop closure online at different place density ($10m$, $20m$ and $50m$). \textit{Precision-Recall Curve} and \textit{F1 score-Recall Curve} are utilized as the evaluation metrics of single-session scenarios, as depicted in Fig.~\ref{fig:single_PR} and~\ref{fig:single_F1}. As we can see, our RING-based methods are less affected by place density than others. At $10m$ place density, RING-based methods show competitive performance with SC-based methods. However, with the decrease of place density, the translation between the query scan and the retrieved map scan enlarges, so our methods outperform them by an increasing margin. Compared with RING (SG) and RING++ (SG), RING and RING++ remain high-performance even at $50m$ place density, which validates the strong translation invariance of RING representation based on TING. The comparison result is consistent with the toycase result illustrated in Fig.~\ref{fig:case_study_2}, validating the robustness to translation of our methods and SC-based methods. Compared with RING (SG) and RING respectively, RING++ (SG) and RING++ achieve better performance, which is more obvious at lower place density. It demonstrates that the extracted local features are beneficial for place representation, thereby improving discrimination. In terms of Fast Histogram, the histogram aggregation is invariant to the rotation difference in the dense place representation. However, without the design of translation invariance, the performance on sparse place representation is limited. In terms of M2DP, the projection strategy improve discrimination but also cannot handle the scenario in which point cloud centerings are not well aligned.

\subsubsection{Multi-session Scenarios}
\begin{figure}[tbp]
	\centering
		\includegraphics[width=9cm]{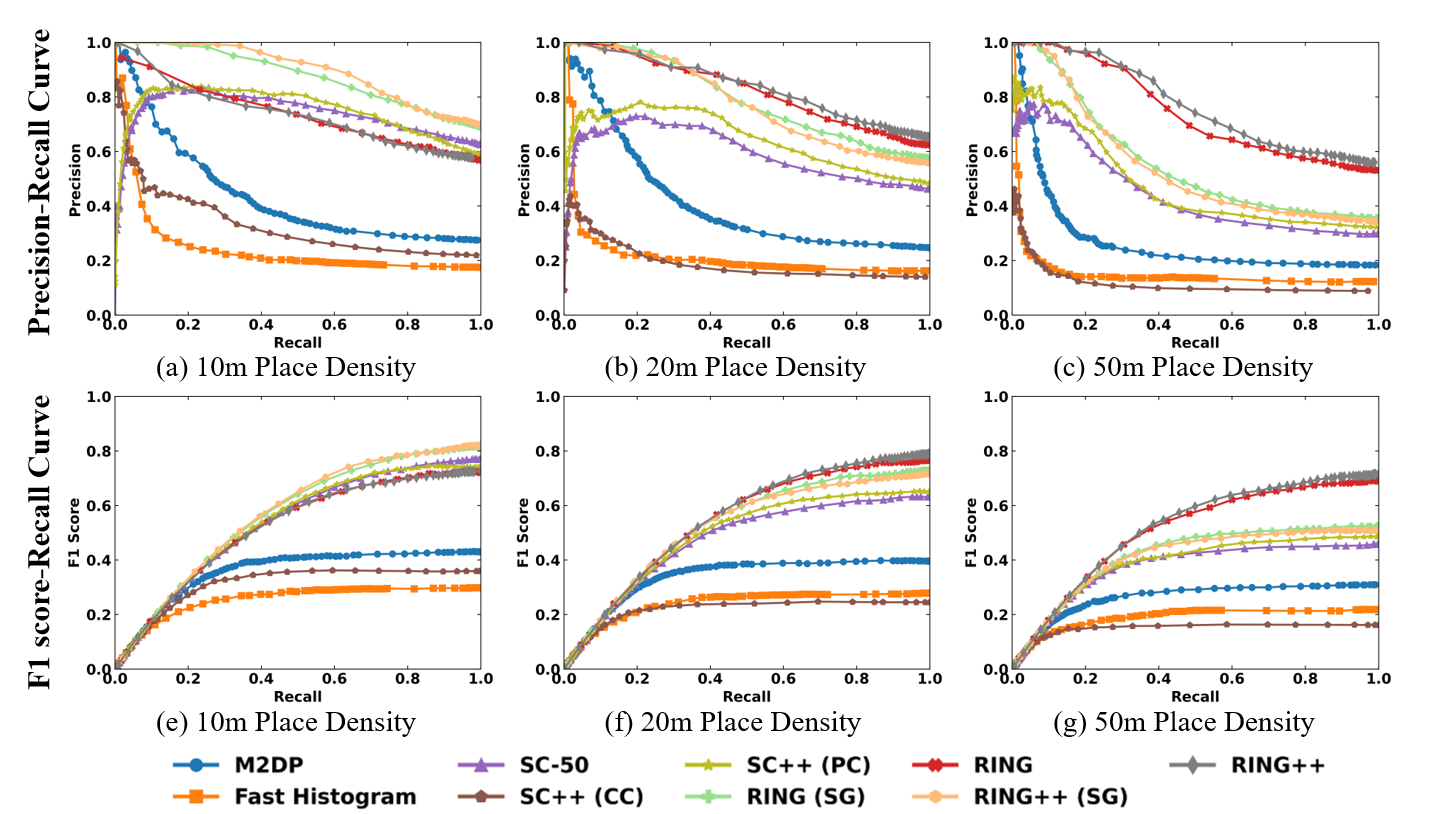}
	\caption{Precision-Recall Curve (top) and F1 score-Recall Curve (bottom) for "2012-03-17" query sequence to "2012-02-04" map sequence in NCLT dataset.}
	\label{fig:multi_session_nclt}
    \vspace{-0.2cm}
\end{figure}

\begin{figure}[tbp]
	\centering
		\includegraphics[width=9cm]{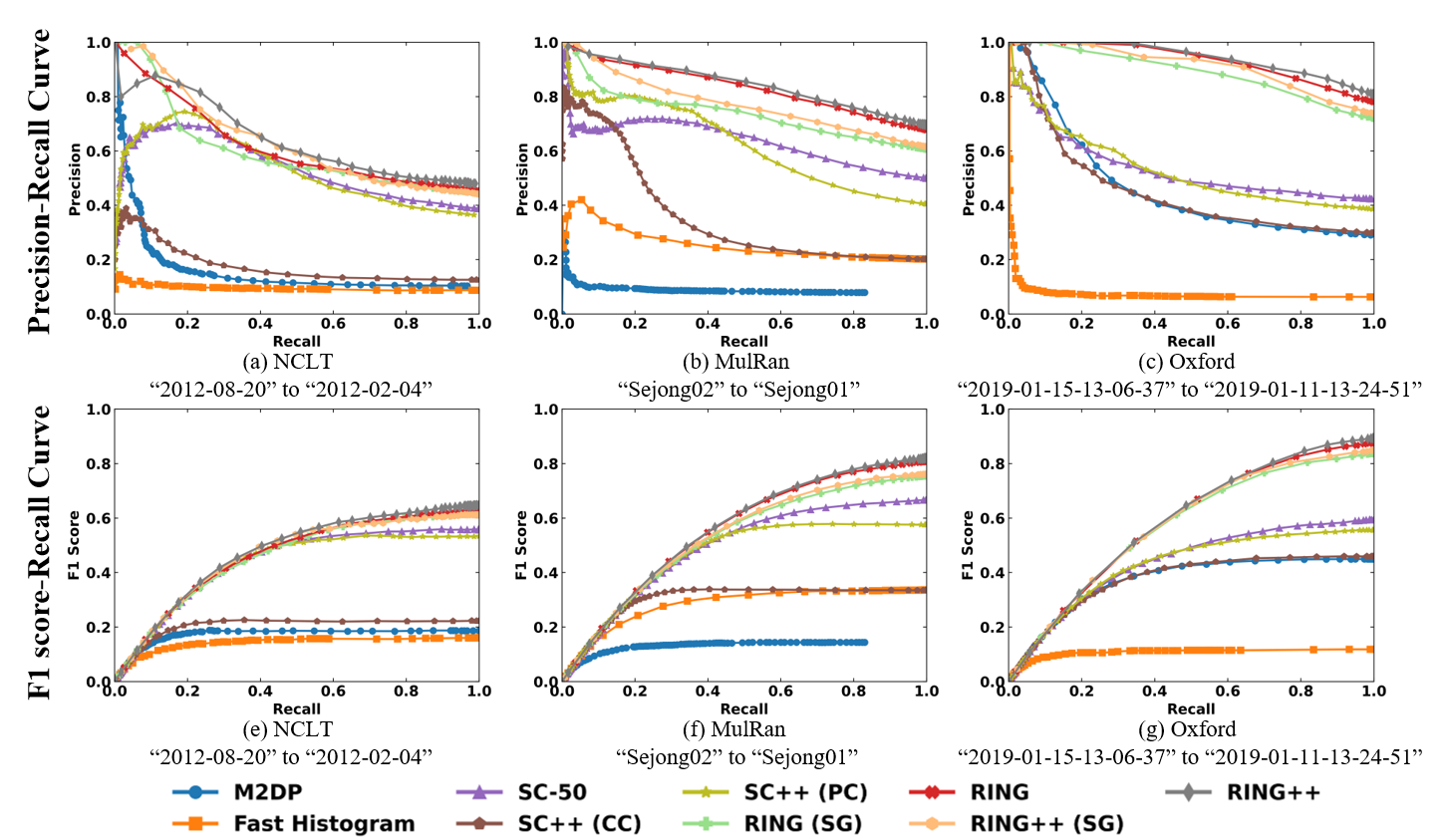}
	\caption{Precision-Recall Curve (top) and F1 score-Recall Curve (bottom) for NCLT, MulRan and Oxford datasets at 20m place density.}
	\label{fig:multi_session_20m}
    \vspace{-0.3cm}
\end{figure}

\begin{figure*}[htb]
	\centering
    \subfloat[M2DP]{
		\includegraphics[width=3.6cm]{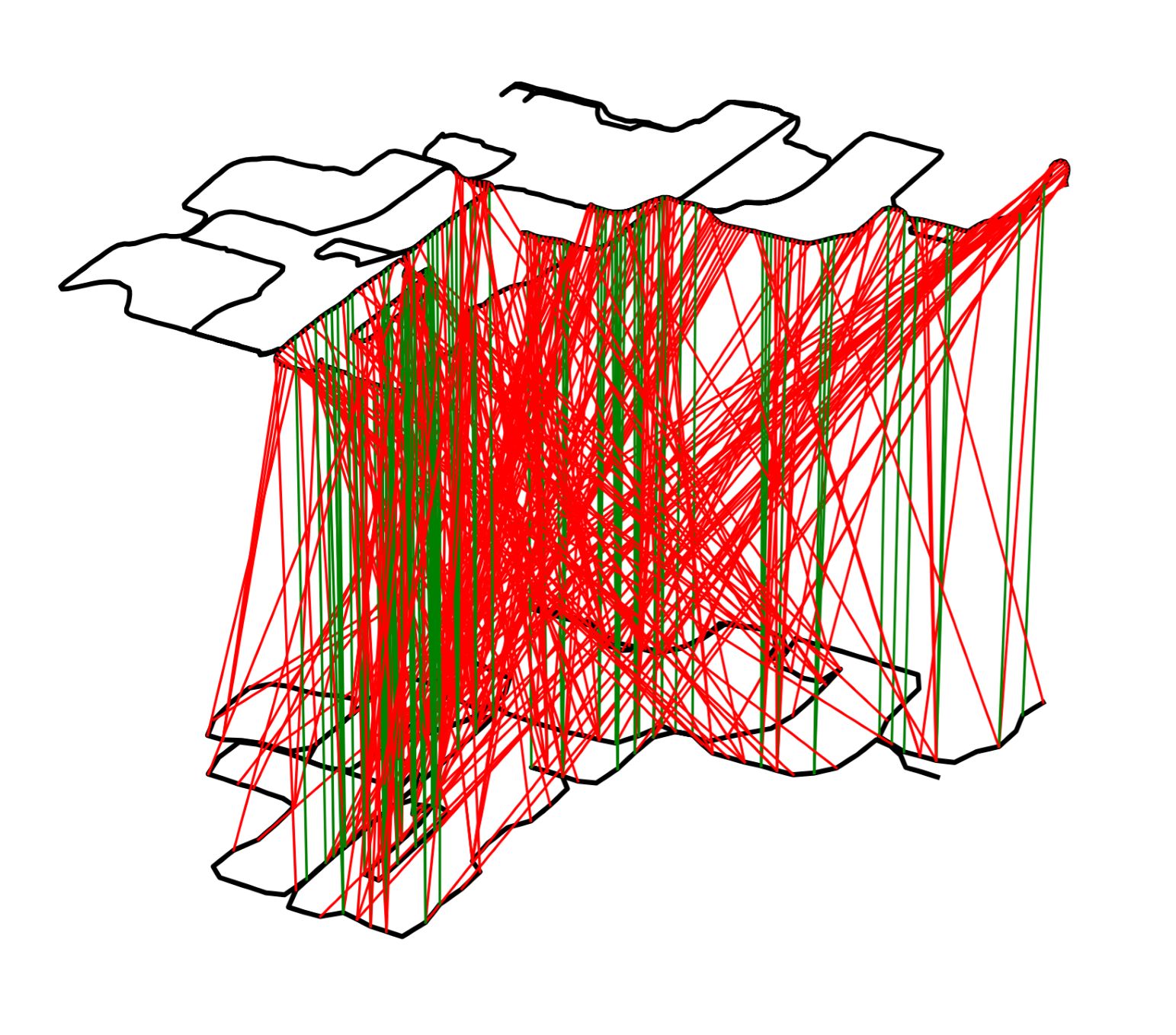}}
	\hspace{-0.6cm}	
    \subfloat[Fast Histogram]{
		\includegraphics[width=3.6cm]{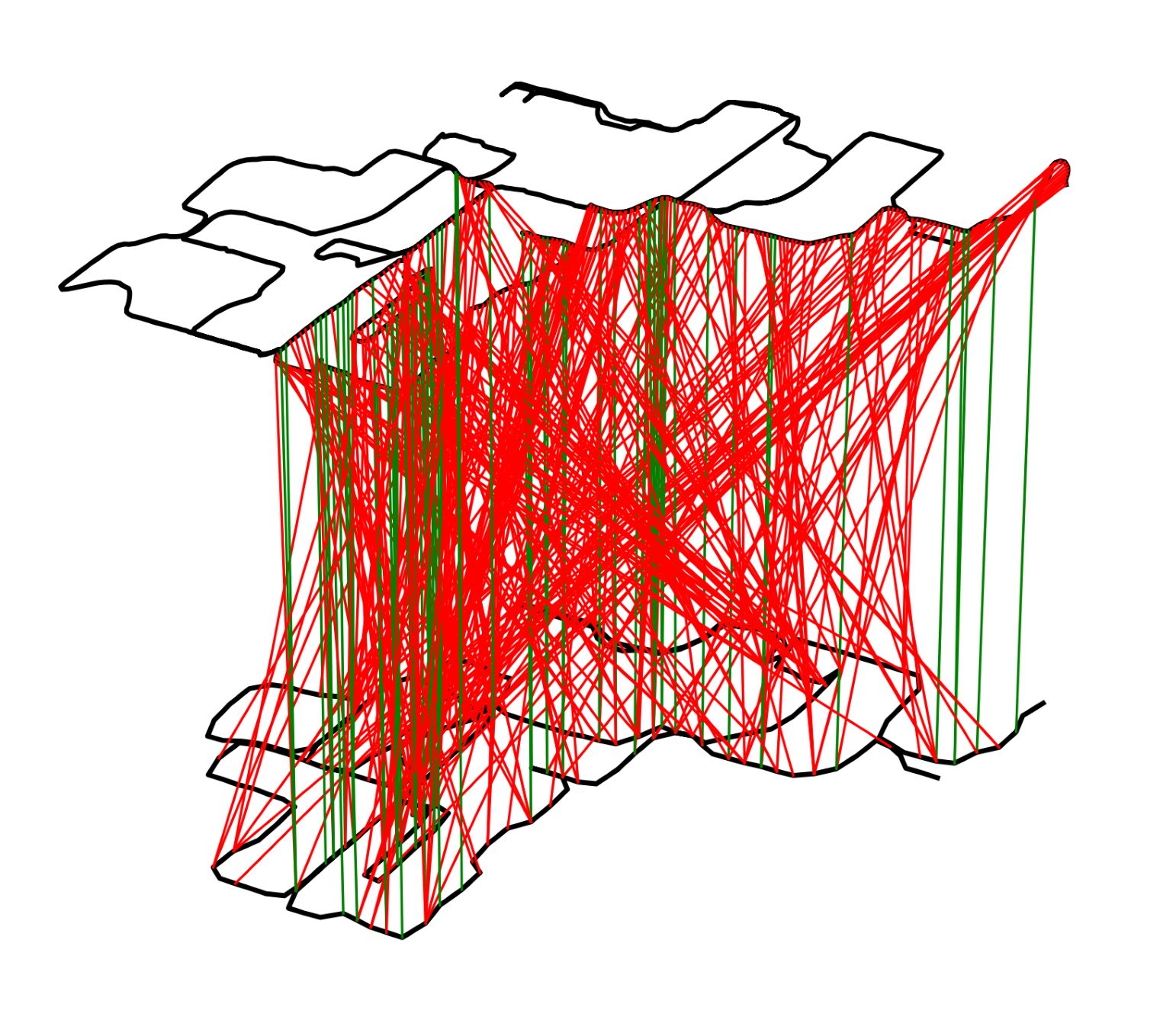}}
    \hspace{-0.6cm}			
    \subfloat[SC]{
		\includegraphics[width=3.6cm]{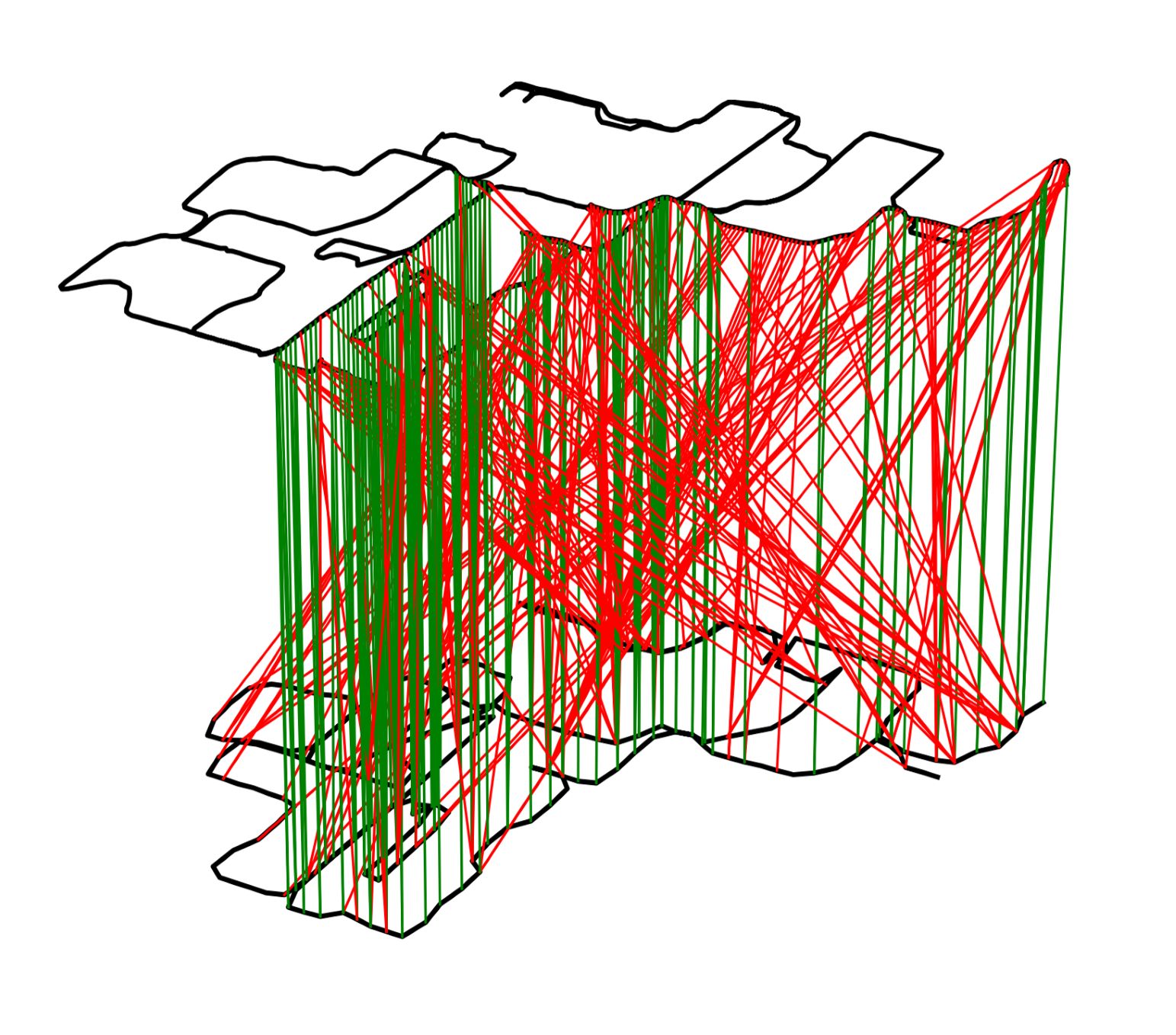}}		
	\hspace{-0.6cm}	
    \subfloat[PC]{
		\includegraphics[width=3.6cm]{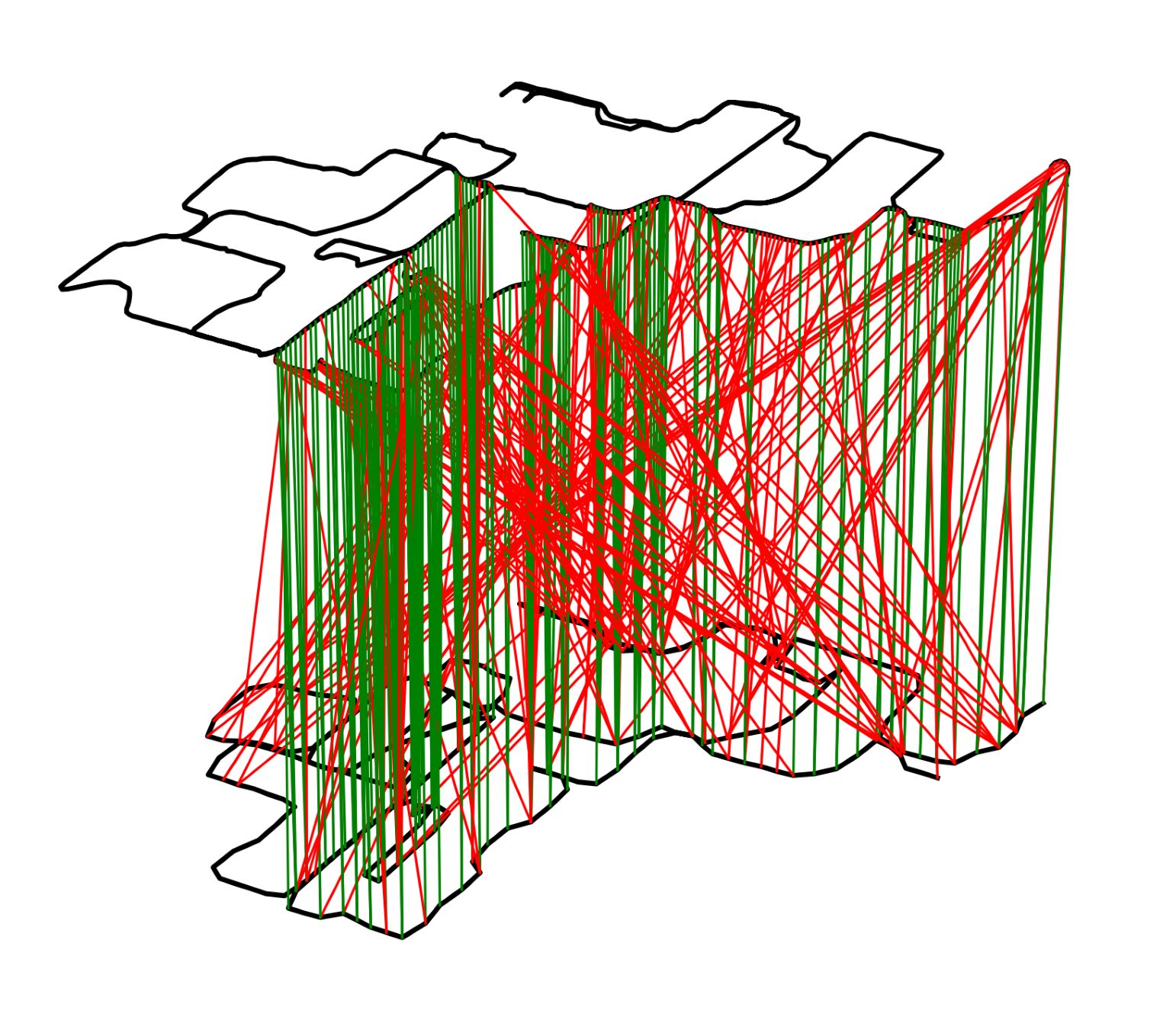}}
	\hspace{-0.6cm}	
    \subfloat[CC]{
		\includegraphics[width=3.6cm]{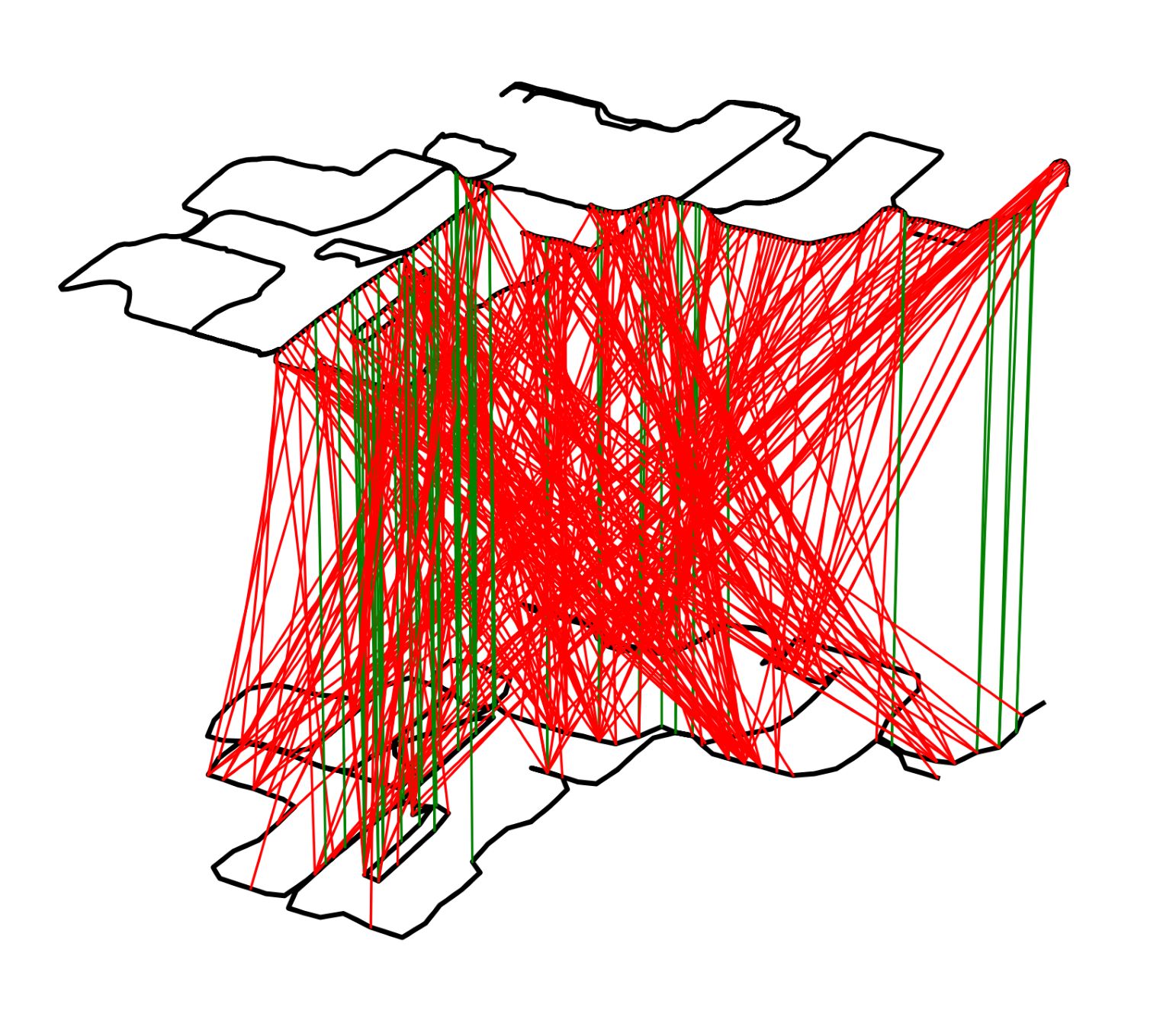}}	
	\vspace{-0.3cm}
    \subfloat[RING (SG)]{
		\includegraphics[width=3.6cm]{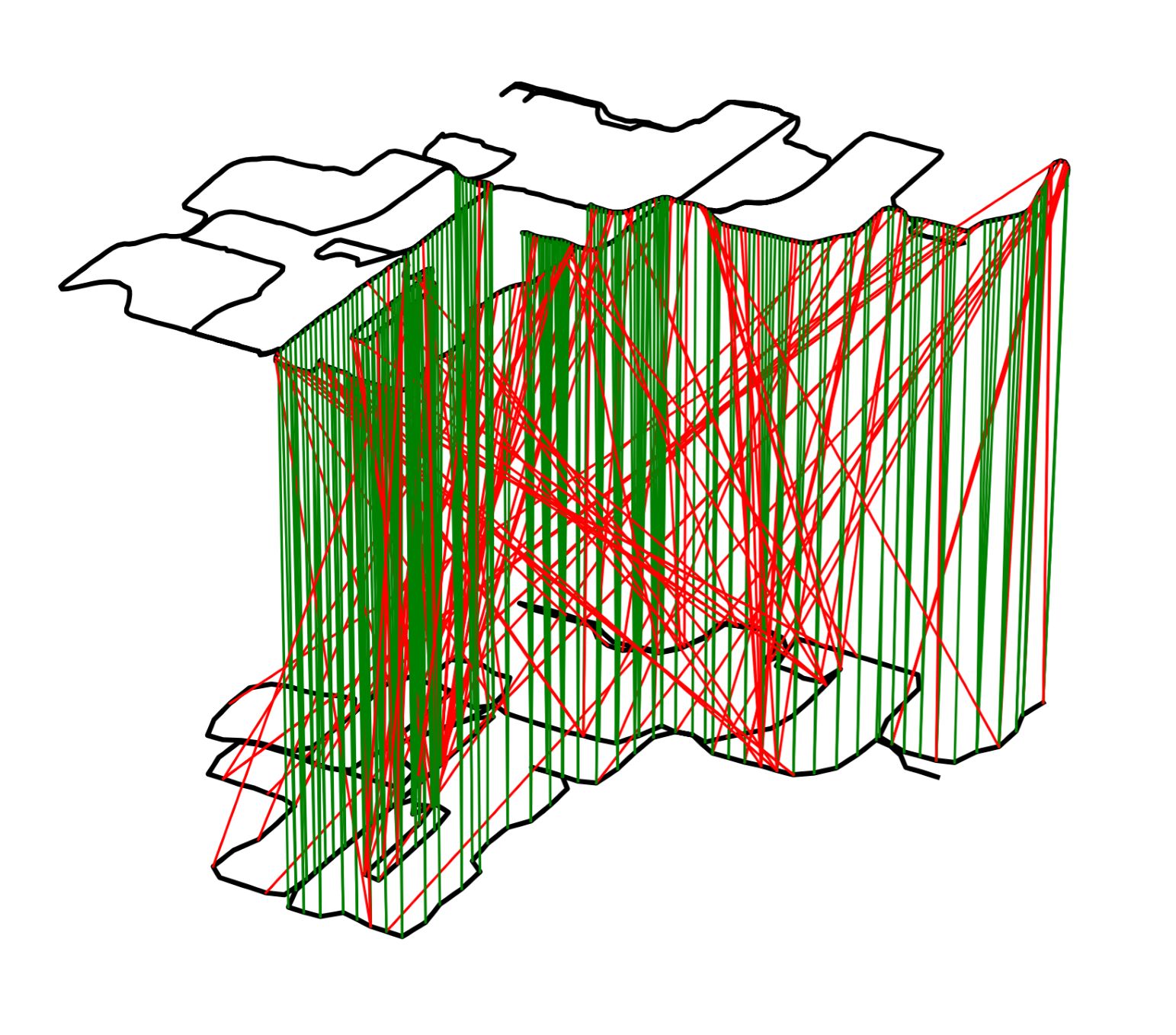}}%
    \hspace{-0.6cm}		
    \subfloat[RING]{
		\includegraphics[width=3.6cm]{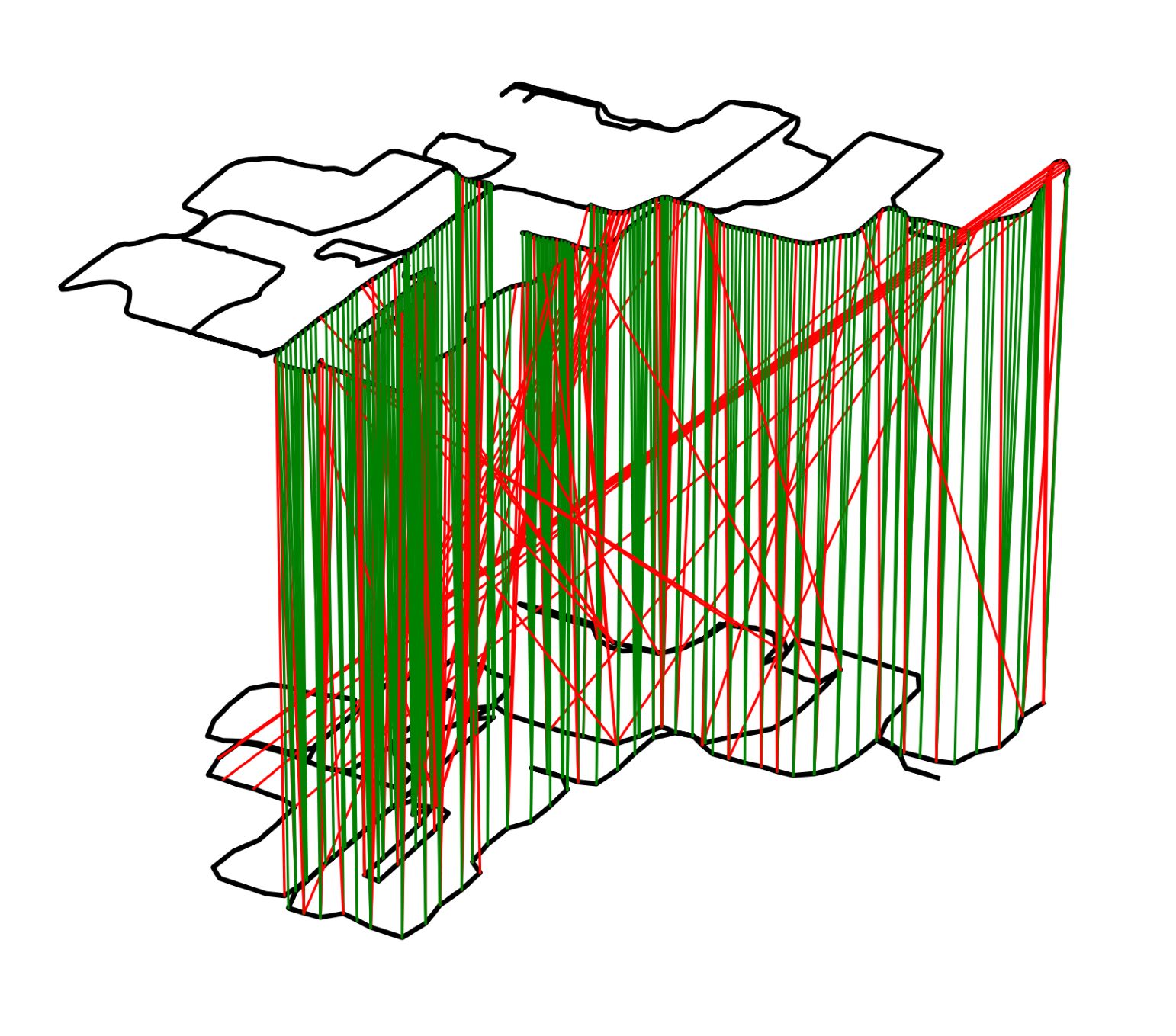}}	
	\hspace{-0.6cm}	
    \subfloat[RING++ (SG)]{
		\includegraphics[width=3.6cm]{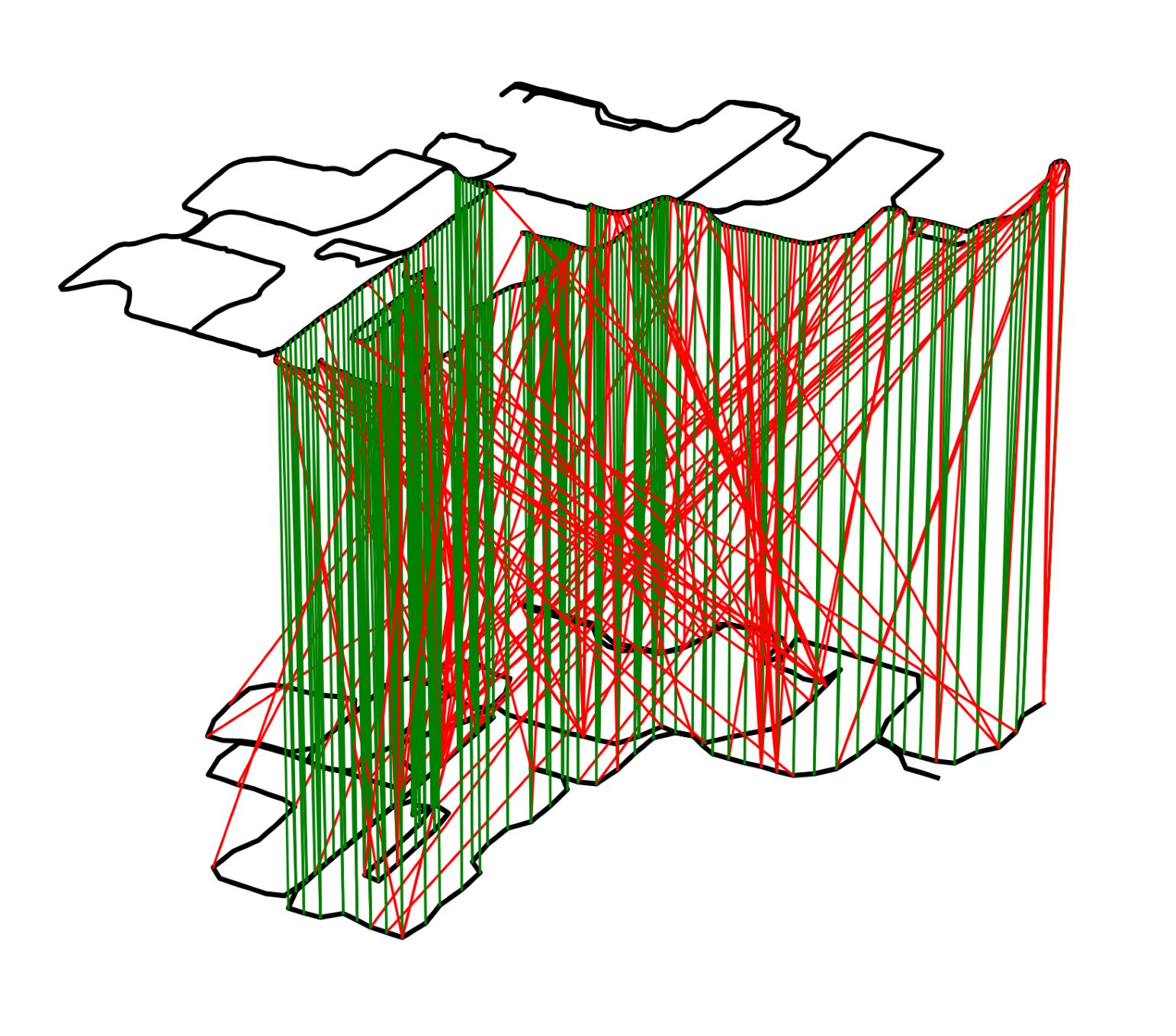}}
    \hspace{-0.6cm}		
    \subfloat[RING++]{
		\includegraphics[width=3.6cm]{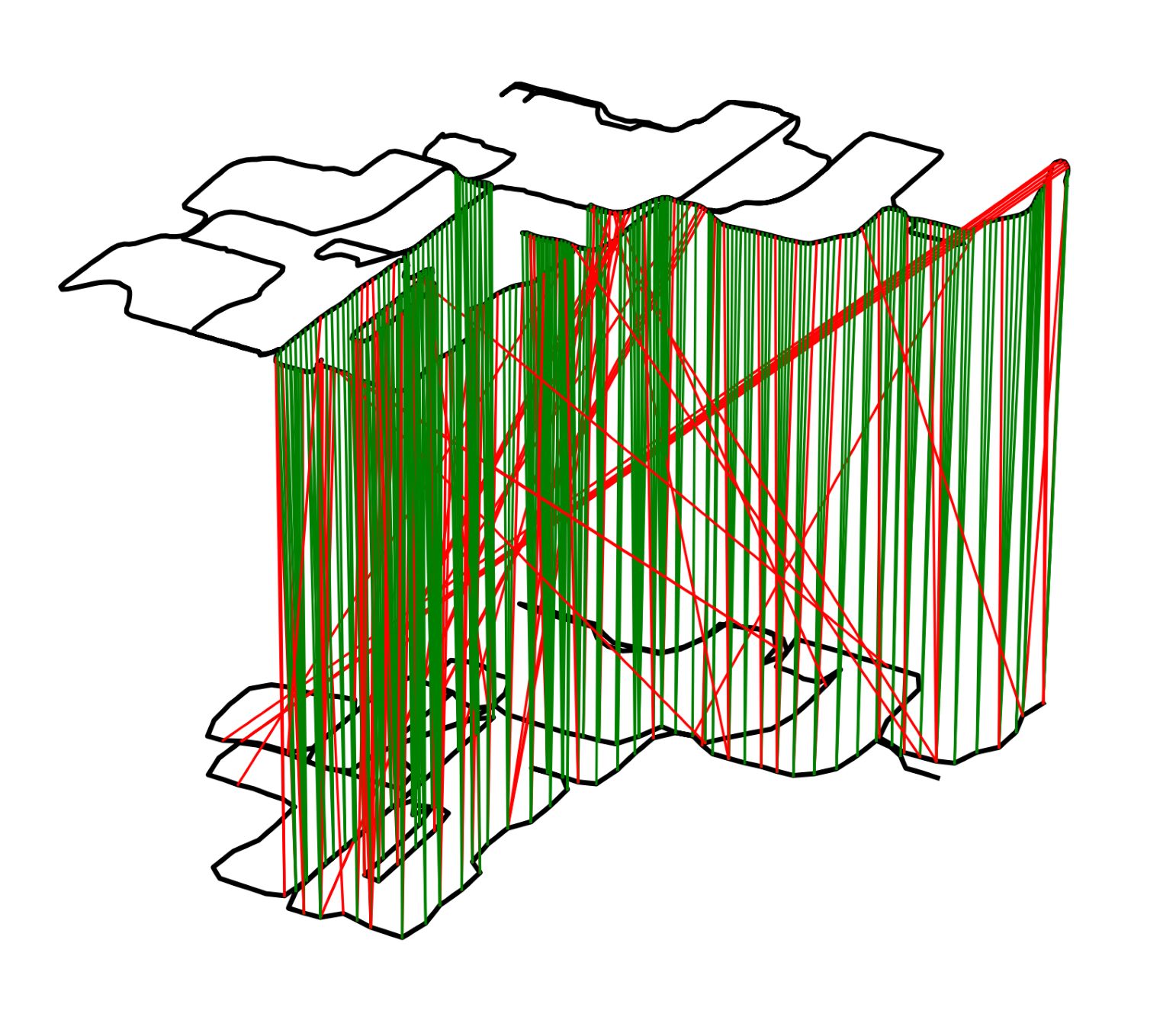}}		
	\caption{Qualitative comparison of match graph for NCLT dataset at 20m place density, where black lines visualize trajectories, green lines visualize correctly retrieved pairs, and red lines visualize mistakenly retrieved pairs.}
	\label{fig:nclt match graph}
    \vspace{-0.3cm}
\end{figure*}

\begin{figure*}[htbp]
	\centering
    \subfloat[M2DP]{
		\includegraphics[width=3.6cm]{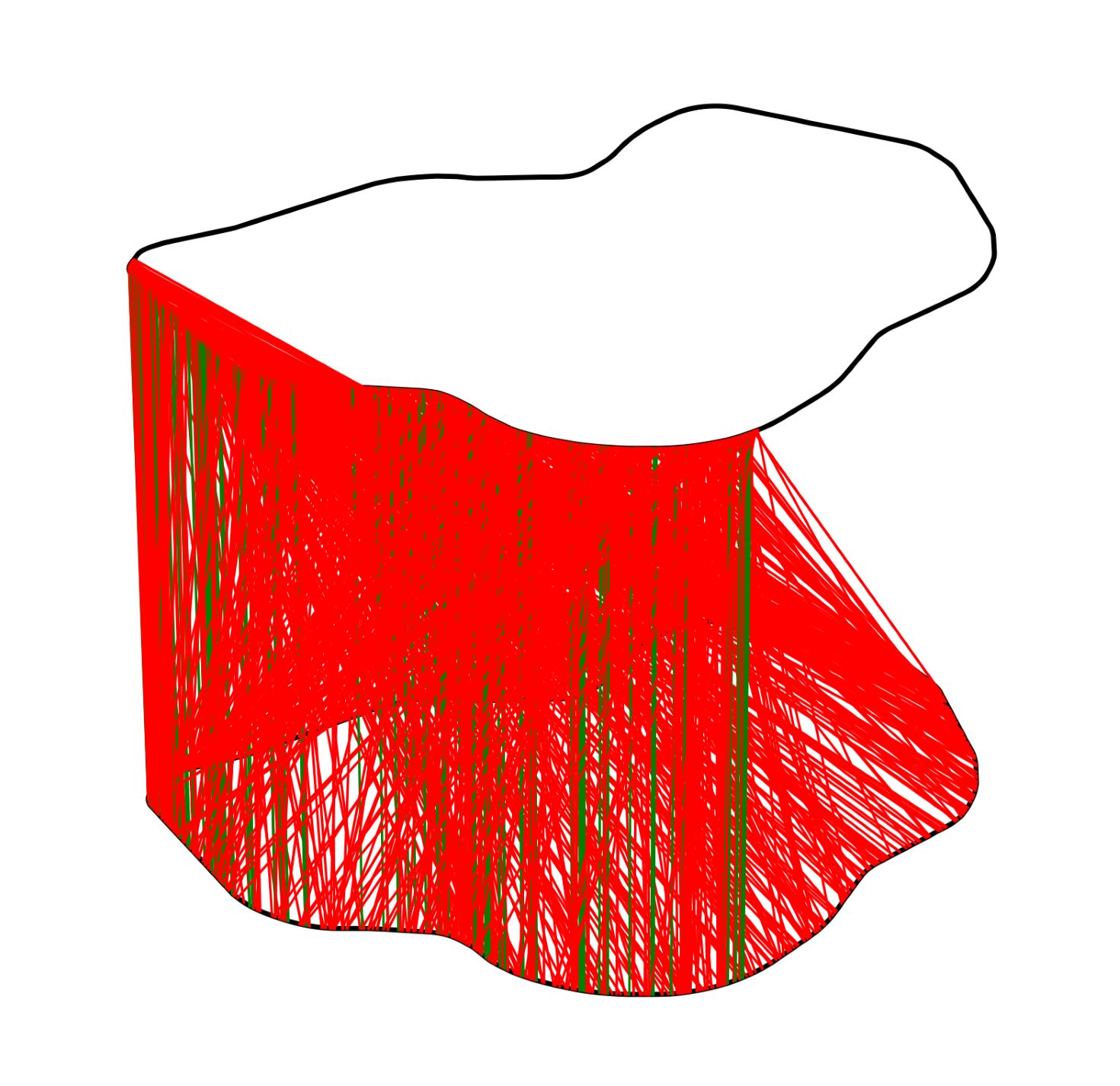}}
	\hspace{-0.6cm}	
    \subfloat[Fast Histogram]{
		\includegraphics[width=3.6cm]{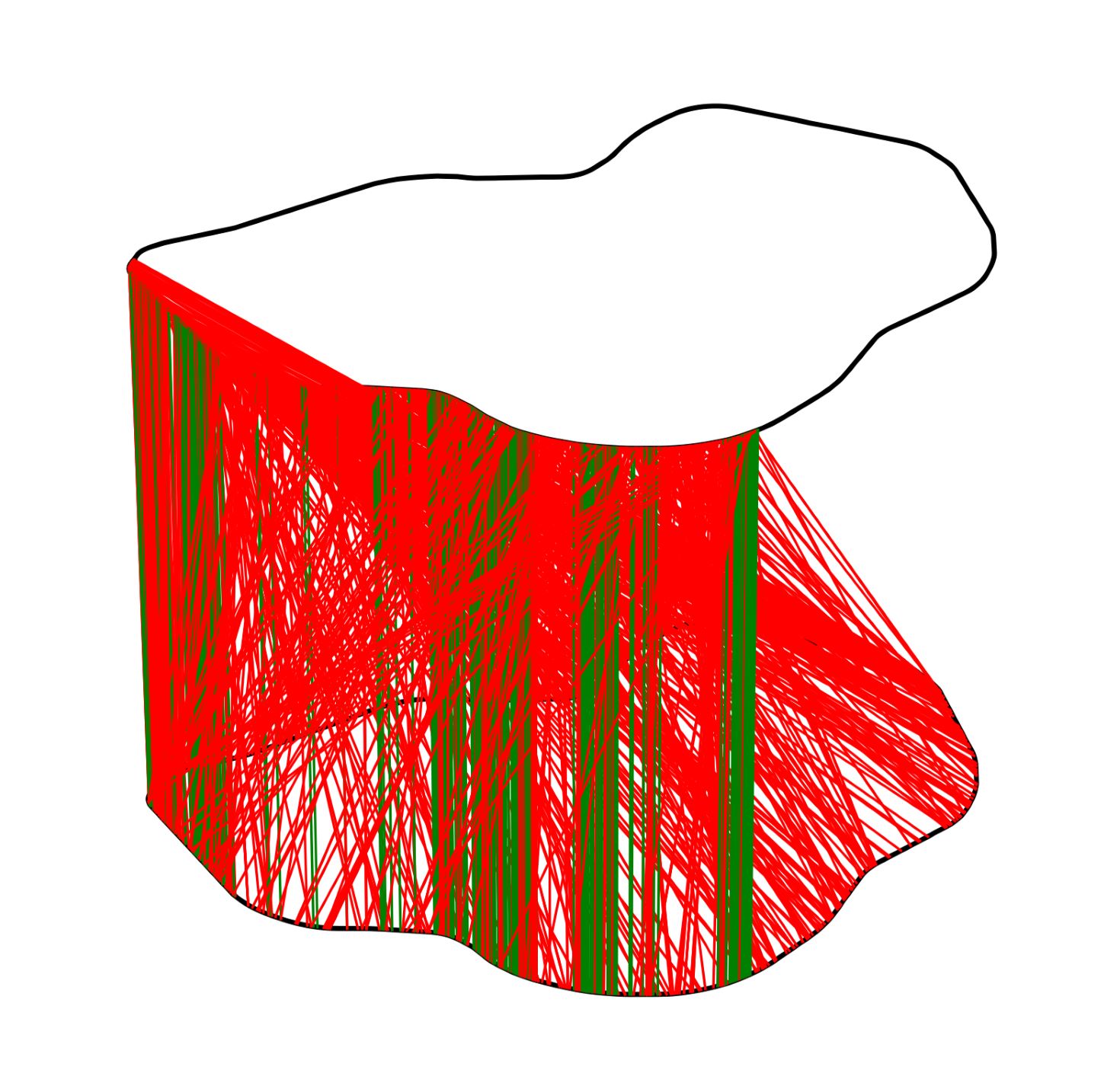}}
    \hspace{-0.6cm}			
    \subfloat[SC]{
		\includegraphics[width=3.6cm]{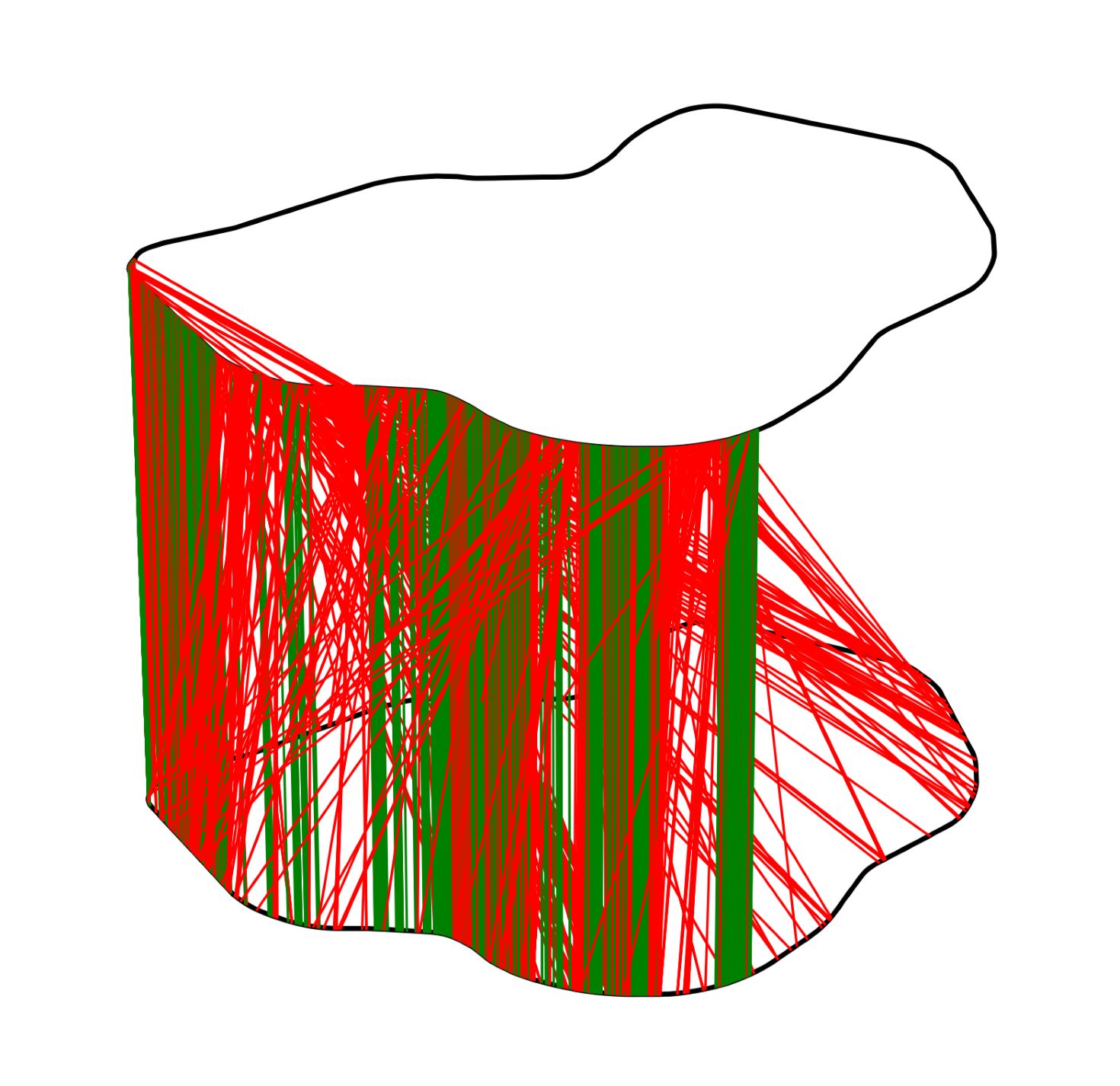}}		
	\hspace{-0.6cm}	
    \subfloat[PC]{
		\includegraphics[width=3.6cm]{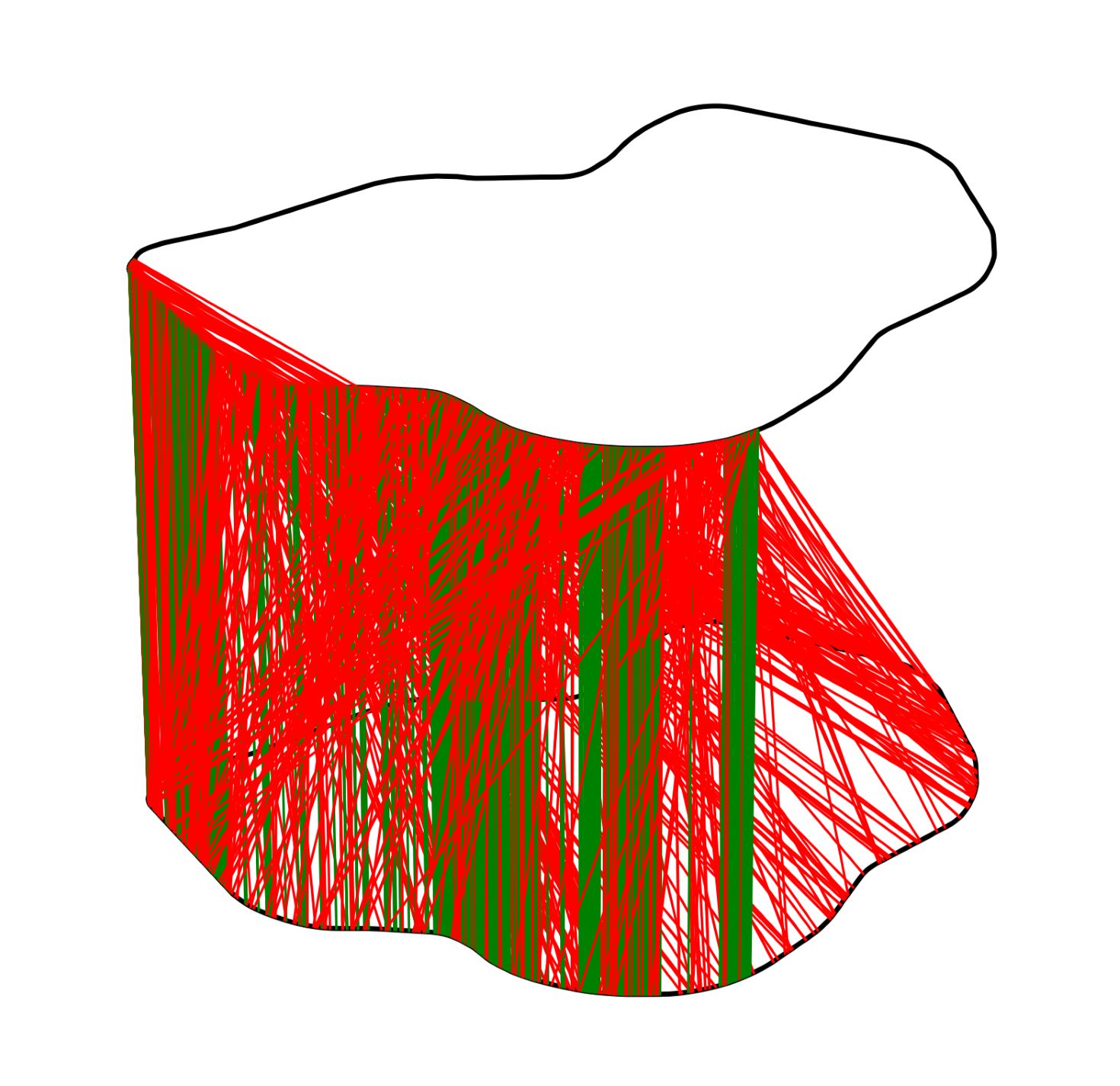}}
	\hspace{-0.6cm}	
    \subfloat[CC]{
		\includegraphics[width=3.6cm]{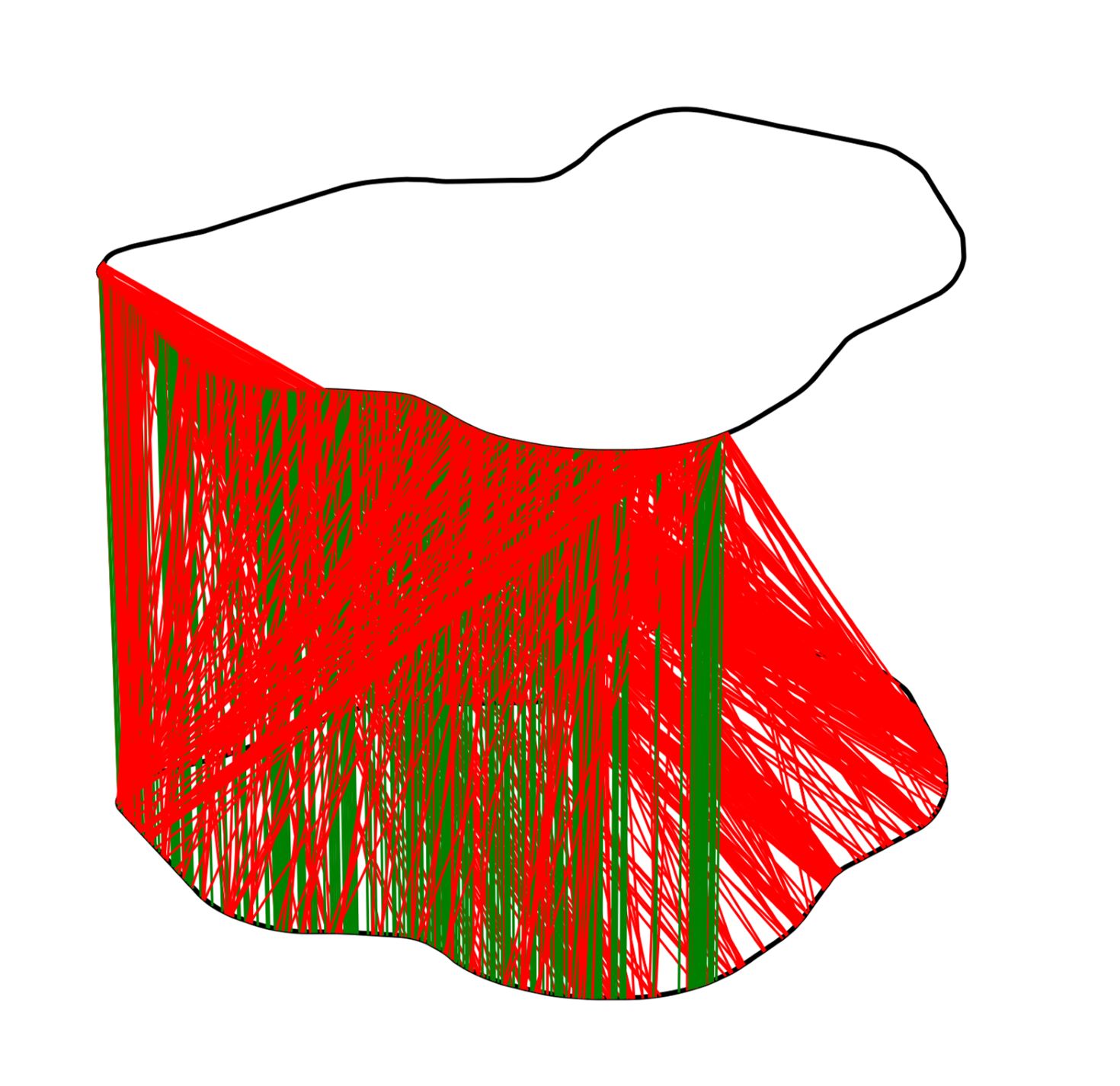}}	
    \subfloat[RING (SG)]{
		\includegraphics[width=3.6cm]{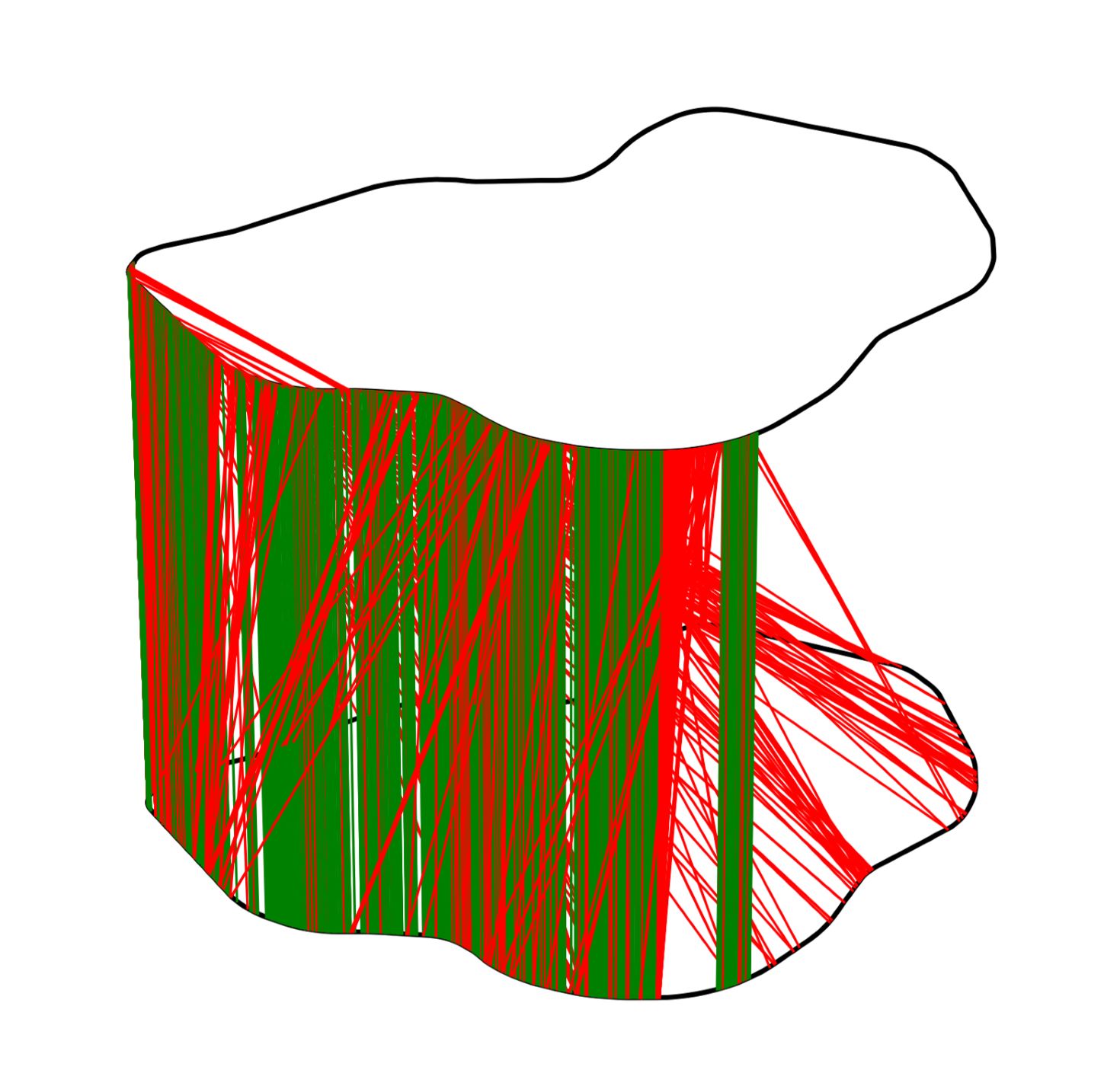}}
    \hspace{-0.6cm}		
    \subfloat[RING]{
		\includegraphics[width=3.6cm]{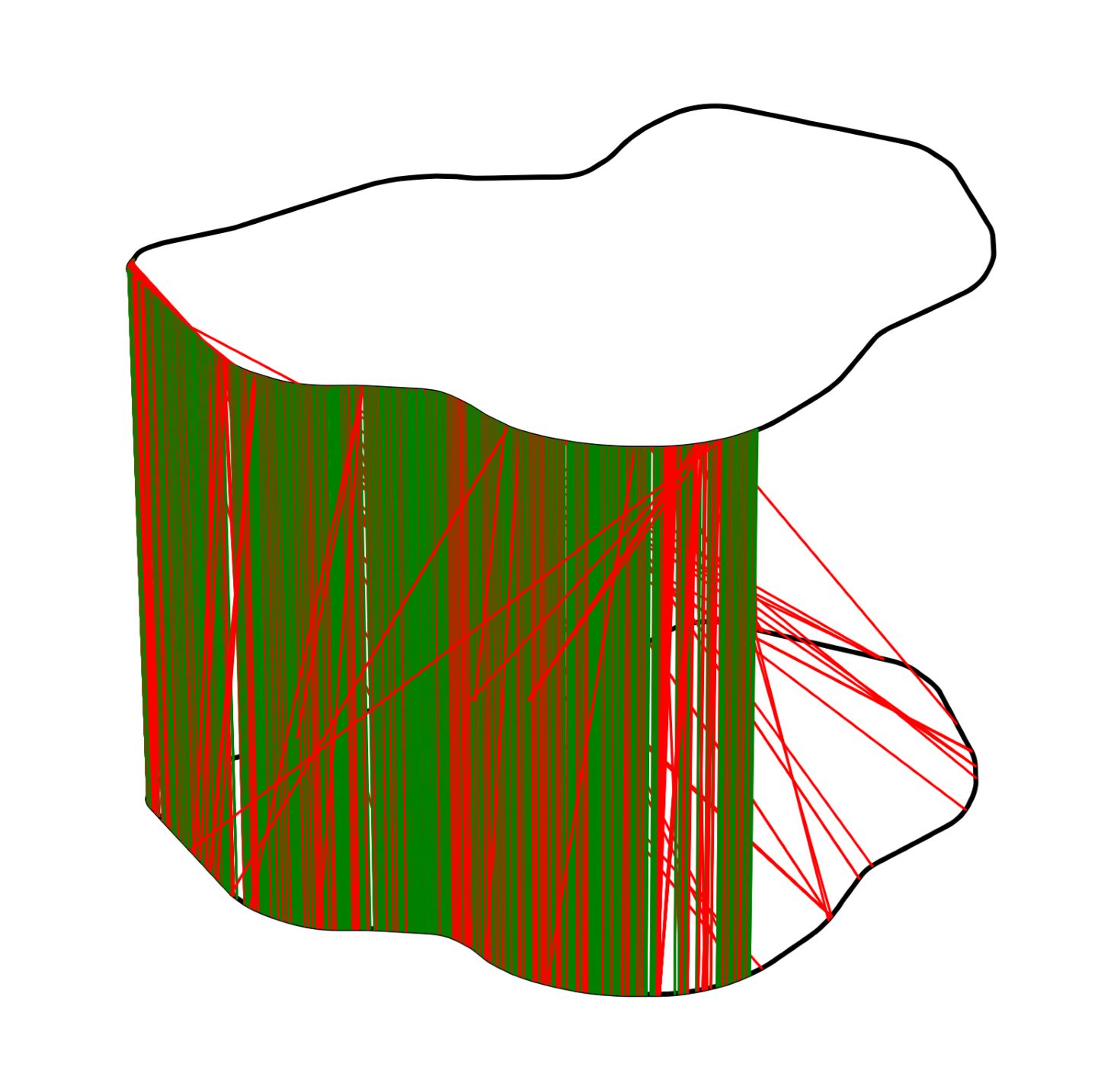}}	
	\hspace{-0.6cm}	
    \subfloat[RING++ (SG)]{
		\includegraphics[width=3.6cm]{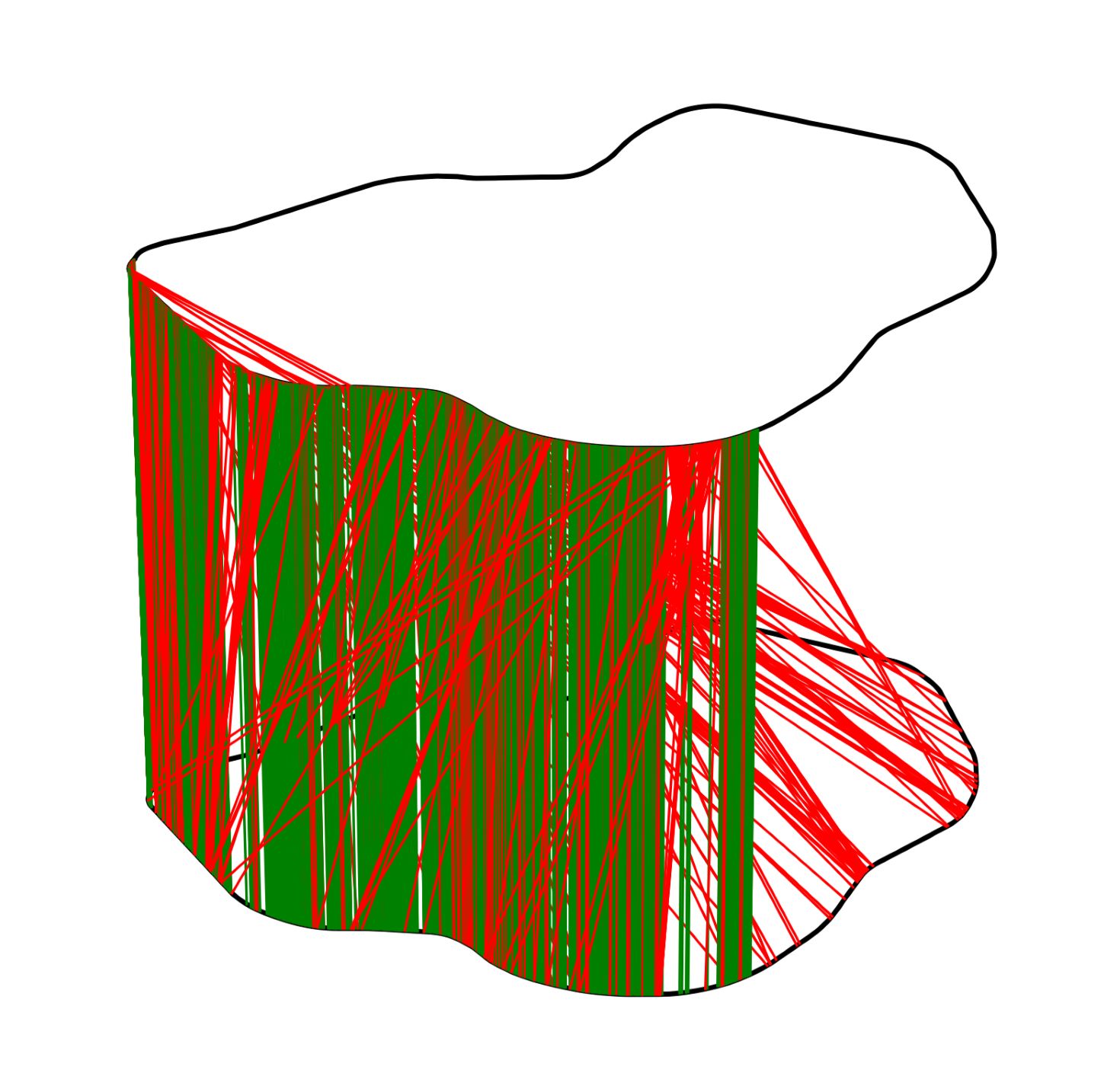}}
    \hspace{-0.6cm}		
    \subfloat[RING++]{
		\includegraphics[width=3.6cm]{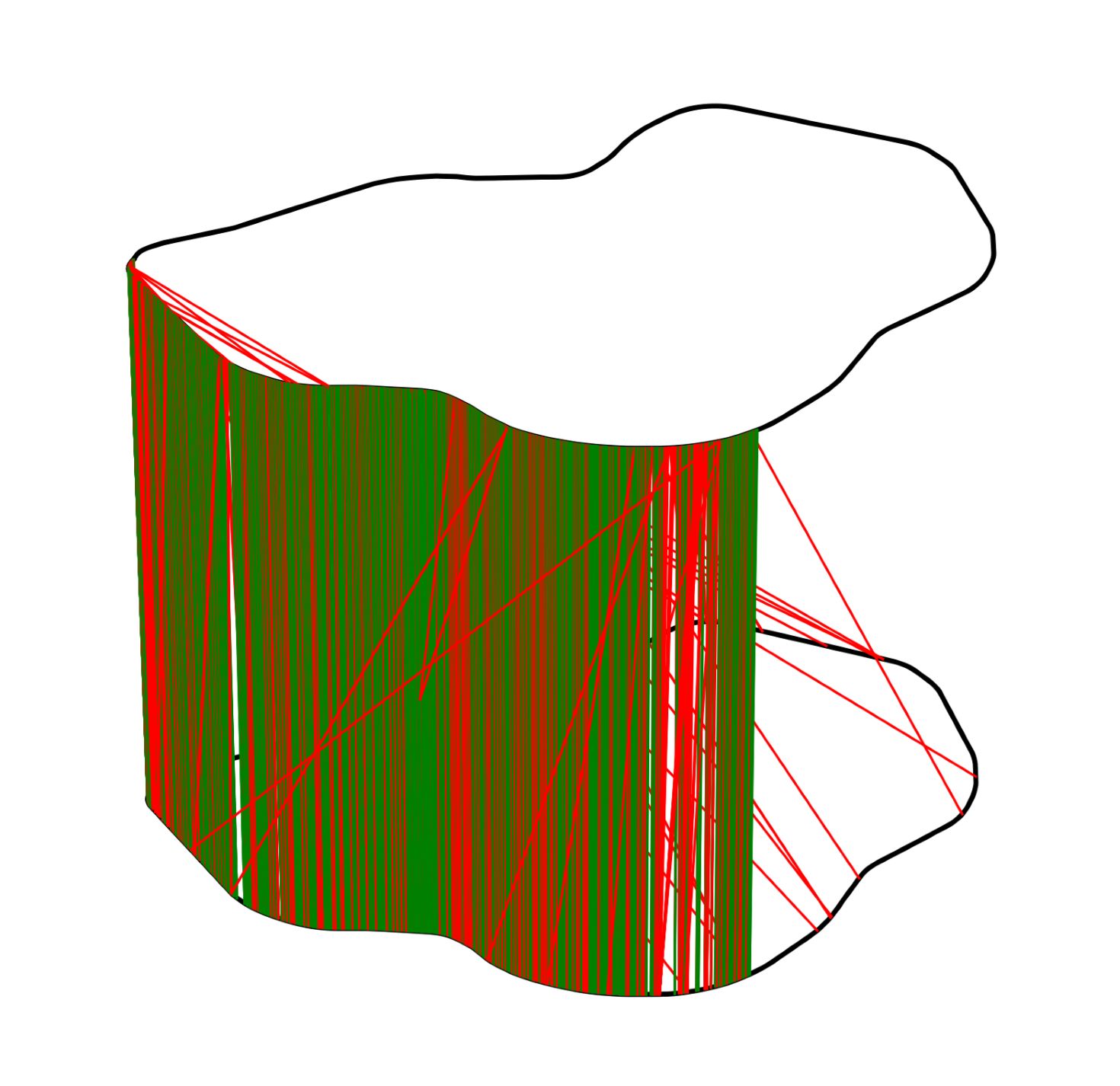}}		
	\caption{Qualitative comparison of match graph for MulRan dataset at 20m place density, where black lines visualize trajectories, green lines visualize correctly retrieved pairs, and red lines visualize mistakenly retrieved pairs.}
	\label{fig:mulran match graph}
    \vspace{-0.2cm}
\end{figure*}

From the viewpoint of long-term autonomy, a robust place recognition system should work well when the surroundings change as time passes. To compare long-term place recognition performance, we select several inter-sessions covering the same region to serve as map sequence and query sequence respectively. Under multi-session scenarios, the sampling interval of query data is fixed to $5m$ for all datasets. In the same manner, we compare the proposed approach against the state-of-the-art methods at different map place density ($10m$, $20m$ and $50m$) on NCLT dataset, with the results depicted in Fig.~\ref{fig:multi_session_nclt}. \textit{Precision-Recall Curve} and \textit{F1 score-Recall Curve} show the same trend as that in Fig.~\ref{fig:single_PR} and~\ref{fig:single_F1}. RING (SG), RING++ (SG) and SC-based methods show competitive performance at high place density ($10m$) while RING and RING++ achieve better performance at low place density ($20m$ and $50m$) due to rigorous roto-translation invariance design. Using six extracted local features, RING++ (SG) and RING++ perform better than RING (SG) and RING, proving the validity of feature extraction for place recognition. To validate the cross-dataset consistency, we evaluate all methods on different datasets at $20m$ place density. Besides ``2012-03-17" query sequence to ``2012-02-04" map sequence for NCLT datsaset, we perform long-term place recognition on ``2012-08-20" to ``2012-02-04" pair for NCLT dataset, ``Sejong02" to ``Sejong01" pair for MulRan dataset, and ``2019-01-15-13-06-37" to ``2019-01-11-13-24-51" pair for Oxford dataset, as shown in Fig.~\ref{fig:multi_session_20m}. Via comparison, four versions of our approach all outperform other approaches at $20m$ place density, especially for Oxford dataset, which presents the strength of roto-translation invariance for place recognition. In Fig.~\ref{fig:nclt match graph} and Fig.~\ref{fig:mulran match graph}, we visualize the revisited pairs of \textit{top1 retrieval} using different methods on NCLT and MulRan datasets at $20m$ place density. It can be easily seen that RING++ surpasses other methods by a lot, further validating the effectiveness of our approach in sparse places.
% which is able to detect more correct loops for backend optimization of SLAM systems.

% In MulRan dataset, we choose "Sejong01" as map sequence and "Sejong02" as query sequence, whose results are visualized in Fig. \ref{fig:Sejong02_Sejong01_PR} and Fig. \ref{fig:Sejong02_Sejong01_F1}. For Oxford Radar RobotCar dataset, Fig. \ref{fig:2019-01-11-13-24-51_2019-01-15-13-06-37_PR} and Fig. \ref{fig:2019-01-11-13-24-51_2019-01-15-13-06-37_F1} show the inter-session place recognition performance of "2019-01-15-13-06-37 (query)" to "2019-01-11-13-24-51 (map)" localization. Based on precision-recall and F1 score-recall metrics, four versions of our approach all outperform other approaches especially at $20m$ and $50m$ place density. 

In addition to qualitative comparison, we provide the quantitative results of the handcrafted methods for long-term place recognition, as listed in Tab.~\ref{multi-session performance}. Compared with other handcrafted methods, our method and its variants are capable of recognizing many more revisited places. Compared with RING (SG) and RING++ (SG), RING and RING++ make obvious improvements thanks to the translation invariance construction of RING representation based on TING. 

\begin{table}[tbp]
\renewcommand\arraystretch{1.2}
\centering
\caption{Quantitative Results of Multi-session Scenarios}
\label{multi-session performance}
\begin{tabular}{clccc}
% \hline
\toprule[1pt]
\textbf{Dataset} & \textbf{Approach} & \textbf{Recall@1} & \textbf{F1 score} & \textbf{AUC} \\ \hline
& \textbf{RING++ (Ours)} & \textbf{0.7321} & \textbf{0.7890} & \textbf{0.8374} \\ 
& RING++ (SG) (Ours) & 0.6309 & 0.7150 & 0.7776 \\
& RING (Ours) & \underline{0.7098} & \underline{0.7658} & \underline{0.8246} \\
& RING (SG) (Ours) & 0.6582 & 0.7274 & 0.7392 \\ 
NCLT & M2DP & 0.2811 & 0.3966 & 0.4127 \\ 
& Fast Histogram & 0.1840 & 0.2785 & 0.2136 \\ 
& SC-50 & 0.5248 & 0.6314 & 0.5934 \\ 
& SC++ (PC) & 0.3862 & 0.6495 & 0.6465 \\
& SC++ (CC) & 0.1355 & 0.2449 & 0.1895 \\ \hline
& \textbf{RING++ (Ours)} & \textbf{0.7451} & \textbf{0.8267} & \textbf{0.8380} \\ 
& RING++ (SG) (Ours) & 0.6592 & 0.7625 & 0.7701 \\
& RING (Ours) & \underline{0.7256} & \underline{0.8079} & \underline{0.8329} \\
& RING (SG) (Ours) & 0.6487 & 0.7549 & 0.7370 \\ 
MulRan & M2DP & 0.0831 & 0.1437 & 0.0750 \\ 
& Fast Histogram & 0.2163 & 0.3363 & 0.2551 \\ 
& SC-50 & 0.5342 & 0.6658 & 0.6359 \\ 
& SC++ (PC) & 0.3740 & 0.6260 & 0.6260 \\
& SC++ (CC) & 0.1829 & 0.3347 & 0.3572 \\ \hline
& \textbf{RING++ (Ours)} & \textbf{0.8274} & \textbf{0.8937} & \textbf{0.9438} \\ 
& RING++ (SG) (Ours) & 0.7612 & 0.8492 & 0.8992 \\
& RING (Ours) & \underline{0.8013} & \underline{0.8772} & \underline{0.9234} \\
& RING (SG) (Ours) & 0.7425 & 0.8366 & 0.8787 \\ 
Oxford & M2DP & 0.2997 & 0.4496 & 0.4638 \\ 
& Fast Histogram & 0.0642 & 0.1181 & 0.0741 \\ 
& SC-50 & 0.4375 & 0.5935 & 0.5366\\ 
& SC++ (PC) & 0.3993 & 0.5561 & 0.5295 \\
& SC++ (CC) & 0.3084 & 0.4596 & 0.4540 \\ 
% \hline
\bottomrule[1pt]
\end{tabular}
\vspace{-0.3cm}
% \vspace{-0.7cm}"
\end{table}

\subsubsection{Generalization}
Unlike deep learning representations, our representation is training-free, so we can easily generalize to new scenes without re-training or fine-tuning. We compare RING++ with some deep learning methods across different datasets to show our superior generalization ability among deep learning methods. For deep learning methods, we train the model on MulRan dataset (``Sejong01" trajectory serves as map sequence, ``Sejong02" trajectory serves as query sequence) and validate it on NCLT dataset (``2012-02-04" trajectory serves as map sequence, ``2012-03-17" trajectory serves as query sequence) and Oxford dataset (``2019-01-11-13-24-51" trajectory serves as map sequence, ``2019-01-15-13-06-37" trajectory serves as query sequence) for generalization ability evaluation, where the equidistant sampling gap between query data is $5m$ and that of map data is $20m$ with $10m$ revisited distance threshold.

As shown in Tab.~\ref{generalization}, RING++ has excellent performance across different datasets. DiSCO and EgoNN maintain most of their place recognition performance on generalized datasets due to their rotation invariant structures, but the polar transform still suffers from translation variance.

\begin{table}[tbp]
\renewcommand\arraystretch{1.2}
\centering
\caption{Generalization Evaluation$^{*}$}
\label{generalization}
\begin{threeparttable}
\begin{tabular}{clccc}
% \hline
\toprule[1pt]
\textbf{Dataset} & \textbf{Approach} & \textbf{Recall@1} & \textbf{F1 score} & \textbf{AUC} \\ \hline
\multirow{4}{*}{MulRan\tnote{*}} & RING++ (Ours) & 0.6941 & 0.7983 & 0.8337 \\ 
& PointNetVLAD & 0.6968 & 0.8214 & 0.8436 \\ 
& \textbf{DiSCO} & \textbf{0.7425}  & \textbf{0.8630} & \underline{0.8973}\\ 
& EgoNN & \underline{0.7260} &  \underline{0.8412} & \textbf{0.9060} \\ \hline
\multirow{4}{*}{NCLT} & \textbf{RING++ (Ours)} & \textbf{0.7321} & \textbf{0.7890} & \textbf{0.8374} \\ 
& PointNetVLAD & 0.3691 & 0.5391 & 0.6260 \\ 
& DiSCO & \underline{0.6036} & \underline{0.7816} & 0.7722\\ 
& EgoNN & 0.5620 & 0.6965 & \underline{0.7988} \\ \hline
\multirow{4}{*}{Oxford} & \textbf{RING++ (Ours)} & \textbf{0.8274} & \textbf{0.8937} & \textbf{0.9438} \\ 
& PointNetVLAD & 0.5247 & 0.7825 & 0.6114 \\ 
& DiSCO & \underline{0.6562} & \underline{0.7890} & \underline{0.8566}\\ 
& EgoNN & 0.5920 & 0.7438 & 0.8471 \\ 
% \hline
\bottomrule[1pt]
\end{tabular}
\begin{tablenotes}
        \footnotesize
        \item[*] The compared learning-based methods are trained on MulRan dataset and evaluated on the other two datasets for generalization evaluation. For MulRan dataset, we perform place recognition evaluation on the test dataset which is used in EgoNN \cite{komorowski2021egonn}.
      \end{tablenotes}
\end{threeparttable}
\vspace{-0.3cm}
% \vspace{-0.7cm}
\end{table}

\subsection{Evaluation of Pose Estimation}

At the pose estimation stage, our method yields a 3-DoF pose: 1-DoF rotation and 2-DoF translation. We perform pose estimation on NCLT and MulRan datasets, which provides accurate 6-DoF ground truth poses. We utilize the same trajectories used in place recognition evaluation to compare the pose estimation performance for complete global localization evaluation.

\begin{figure}[t]
	\centering
    \subfloat[Rotation Error]{
		\includegraphics[width=4.3cm]{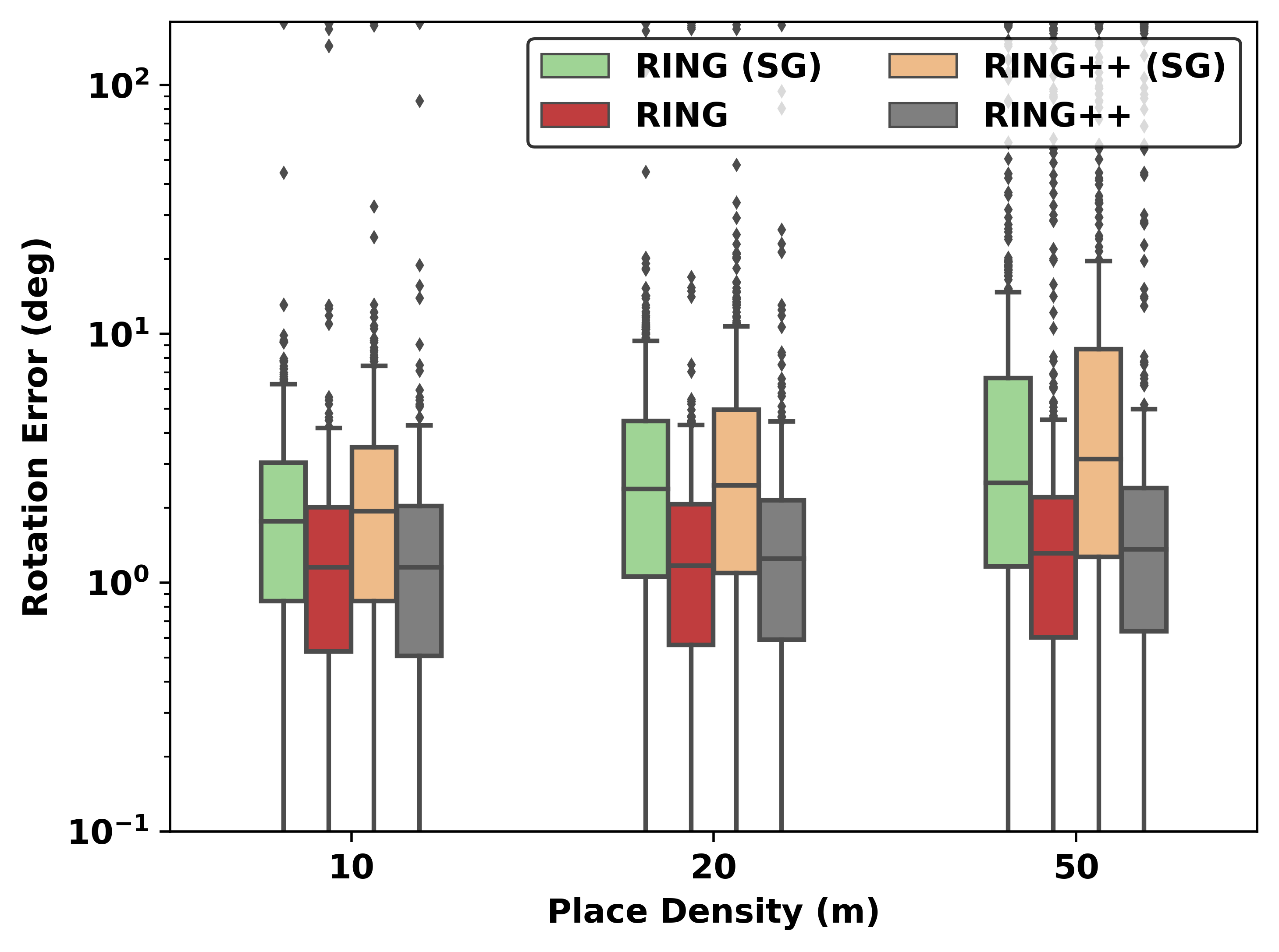}}
    \hspace{-0.4cm}
    \subfloat[Translation Error]{
		\includegraphics[width=4.3cm]{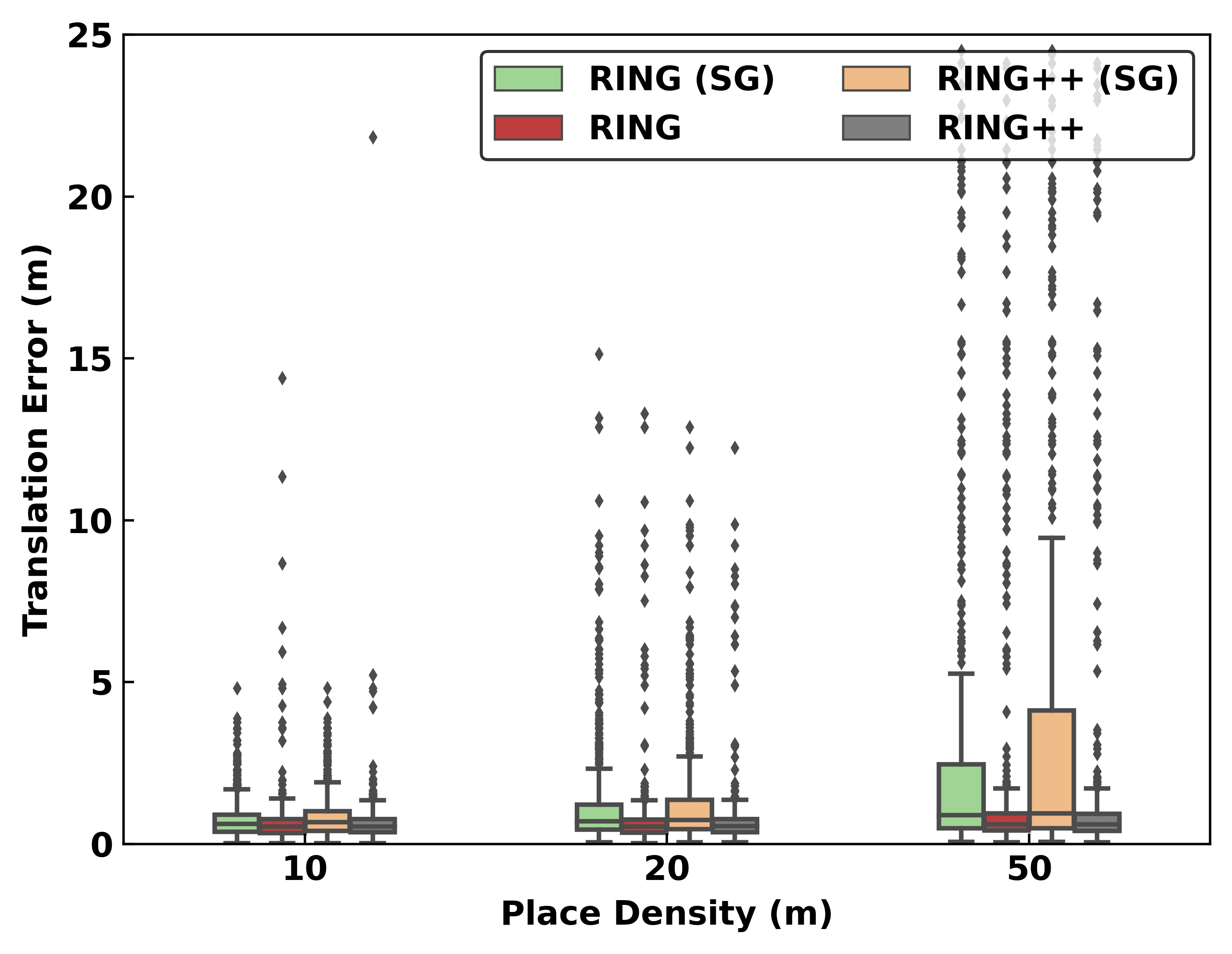}}
	\caption{Pose estimation error on NCLT dataset. (a) Relative 1-DoF rotation estimation error. (b) Relative 2-DoF translation estimation error.}
	\label{fig:nclt_pe}
    \vspace{-0.5cm}
\end{figure}

We first compare the proposed four variants. The pose estimation error on ``2012-03-17" to ``2012-02-04" for NCLT dataset at different place density is visualized by box plot shown in Fig.~\ref{fig:nclt_pe}. Together with pose estimation on NCLT dataset, we provide Tab.~\ref{pose error 1} which presents \textit{Success Rate}, \textit{TE} and \textit{RE} of our approach on ``Sejong02" to ``Sejong01" for MulRan dataset with a place density of $20m$. As the results show, RING and RING++ estimate more accurate rotation and translation than RING (SG) and RING++ (SG), which indicates that translation invariance benefits pose estimation. By comparing RING with RING++, we find that the extracted six local features of RING++ do not obviously facilitate pose estimation performance. In general, RING++ performs the best among all variants.

% \begin{table}
% \renewcommand\arraystretch{1.5}
% \centering
% \caption{Pose Estimation Error on MulRan Dataset}
% \label{pose error 1}
% \begin{threeparttable}
% \begin{tabular}{lccc}
% \toprule[1pt]
% \multirow{2}{*}{Approach} & \multicolumn{3}{c}{MulRan} \\ \cline{2-4}
%  &  \multicolumn{1}{c}{Success ($\%$)} & \multicolumn{1}{c}{TE (m)} & \multicolumn{1}{c}{RE (\textdegree)} \\ \hline
% RING (SG) & 34.34 & 1.71/4.07/7.81 & 0.61/2.68/6.02 \\
% RING & \underline{44.65} & \underline{1.11}/\underline{3.48}/\underline{7.59} & \textbf{0.37}/\textbf{0.81}/\underline{2.59} \\
% RING++ (SG) & 35.93 & 1.64/3.80/7.67 & 0.73/2.91/6.27 \\
% \textbf{RING++} & \textbf{48.89} & \textbf{0.92}/\textbf{3.06}/\textbf{7.46} & \textbf{0.37}/\underline{0.82}/\textbf{2.58} \\
% \bottomrule[1pt]
% \end{tabular}
% \begin{tablenotes}
%     \footnotesize
%     \item[*] Our four versions without ICP refinement only estimate 3-DoF relative pose between two point clouds, therefore TE represents 2-DoF translation error and RE represents 1-DoF rotation error here. We list 50\%, 75\% and 95\% quantiles of TE and RE in this table.
% \end{tablenotes}
% \end{threeparttable}
% \end{table}

\begin{table}
\renewcommand\arraystretch{1.5}
\centering
\caption{Pose Estimation Error on MulRan Dataset$^{*}$}
\label{pose error 1}
\begin{threeparttable}
\begin{tabular}{lccc}
\toprule[1pt]
\multirow{2}{*}{Approach} & \multicolumn{3}{c}{MulRan} \\ \cline{2-4}
 &  \multicolumn{1}{c}{Success ($\%$)} & \multicolumn{1}{c}{TE (m)} & \multicolumn{1}{c}{RE (\textdegree)} \\ \hline
RING (SG) & 53.01 & 0.58/\underline{1.08}/4.52 & 0.51/1.77/5.94 \\
RING & \underline{63.47} & \underline{0.51}/\textbf{0.70}/\underline{1.81} & \underline{0.35}/\underline{0.73}/\underline{1.52} \\
RING++ (SG) & 53.01 & 0.60/1.17/4.54 & 0.62/2.54/6.24 \\
\textbf{RING++} & \textbf{65.76} & \textbf{0.50}/\textbf{0.70}/\textbf{1.79} & \textbf{0.34}/\textbf{0.72}/\textbf{1.51} \\
\bottomrule[1pt]
\end{tabular}
\begin{tablenotes}
    \footnotesize
    \item[*] We list 50\%, 75\% and 95\% quantiles of TE and RE in this table.
\end{tablenotes}
\end{threeparttable}
\vspace{-0.1cm}
\end{table}

\begin{table*}
\renewcommand\arraystretch{1.5}
\centering
\caption{Quantitative Results of Pose Error Evaluation$^{*}$}
\label{pose error 2}
\begin{threeparttable}
\begin{tabular}{clcccccc}
\toprule[1pt]
\multicolumn{2}{c}{\multirow{2}{*}{Approach}} & \multicolumn{3}{c}{NCLT} & \multicolumn{3}{c}{MulRan} \\ \cline{3-8}
\multicolumn{2}{c}{} & \multicolumn{1}{c}{Success ($\%$)} & \multicolumn{1}{c}{TE$^{\dagger}$ (m)} & \multicolumn{1}{c}{RE$^{\dagger}$ (\textdegree)} & \multicolumn{1}{c}{Success ($\%$)} & \multicolumn{1}{c}{TE$^{\dagger}$ (m)} & \multicolumn{1}{c}{RE$^{\dagger}$ (\textdegree)} \\ \hline

\multirow{3}{*}{Handcrafted}  & SC-50 + ICP & 24.97/47.58  & \textbf{0.30}/\underline{1.02}/\underline{7.67} & 4.14/\textbf{6.50}/\underline{18.11} & 35.82/65.45 & 0.45/6.54/8.17 & 0.52/0.95/2.01 \\
                                & SC++ (PC) + ICP & 25.68/46.95 & 0.33/1.87/8.78 & 4.18/6.70/19.81 & 36.25/66.94 & 0.38/6.12/\underline{8.10} & \underline{0.51}/\underline{0.94}/1.96 \\
                                & SC++ (CC) + ICP & 8.09/\underline{50.98}  & 0.36/14.30/139.55 & \textbf{3.67}/7.06/177.01 & 18.19/62.23 & 0.48/138.85/140.63 & \textbf{0.47}/\textbf{0.80}/\underline{1.48} \\ \hline
\multirow{2}{*}{Learning-based} & DiSCO + ICP & \underline{28.31}/46.91 & 0.35/2.73/8.73 & 6.09/\underline{6.52}/34.69 & 38.36/51.66 & 1.23/7.45/8.25 & 0.71/1.02/1.84  \\
                                & EgoNN + ICP & 6.07/10.80 & 3.48/8.12/15.51 & 12.32/27.45/134.95 & \underline{62.00}/\underline{85.40} & \underline{0.28}/\underline{0.52}/13.04 & 0.72/1.14/1.98 \\ \hline
\textbf{Ours} & \textbf{RING++ + ICP} & \textbf{42.37}/\textbf{57.88} & \underline{0.31}/\textbf{0.53}/\textbf{1.36} & \underline{4.13}/6.74/\textbf{12.52} & \textbf{66.09}/\textbf{95.22} & \textbf{0.27}/\textbf{0.42}/\textbf{1.82} & \underline{0.51}/\textbf{0.80}/\textbf{1.19} \\ 
\bottomrule[1pt]
\end{tabular}
\begin{tablenotes}
    \footnotesize
    \item[$\dagger$] As ICP refinement is applied, TE here represents 3-DoF translation error and RE represents 3-DoF rotation error.
    \item[*] We list 50\%, 75\% and 95\% quantiles of TE and RE in this table. The second value of \textit{Success Rate} is the percentage of successfully aligned matches $(TE <2 m \; \& \; RE < 5^\circ)$ among all correctly retrieved pairs. For MulRan dataset, we perform place recognition and pose estimation evaluation on the test dataset which is used in EgoNN \cite{komorowski2021egonn}.
\end{tablenotes}
\end{threeparttable}
\vspace{-0.4cm}
\end{table*}

\begin{figure*}[tbp]
	\centering
    \subfloat[Source]{
		\includegraphics[width=3.7cm]{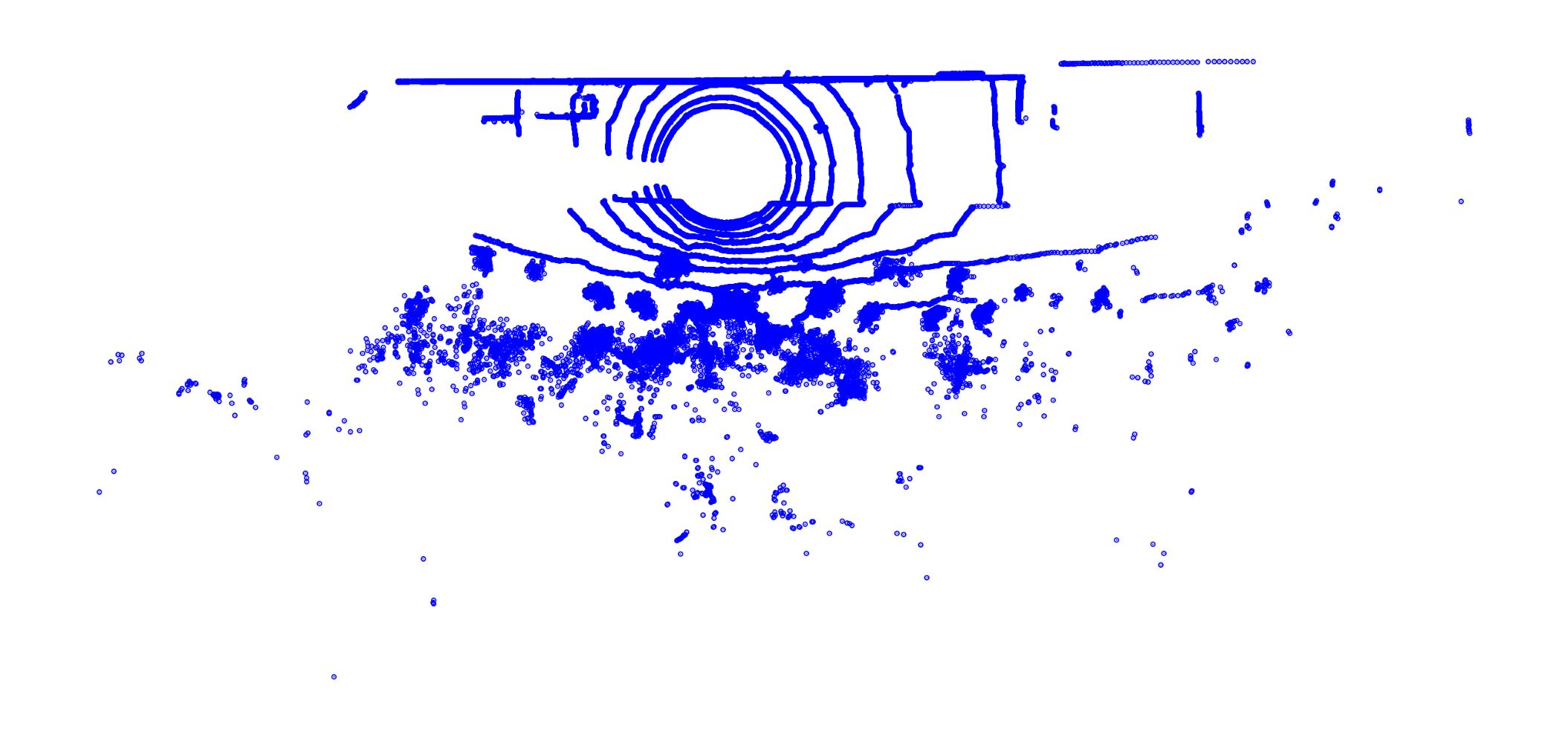}}
    \hspace{-0.5cm}			
    \subfloat[Target]{
		\includegraphics[width=3.7cm]{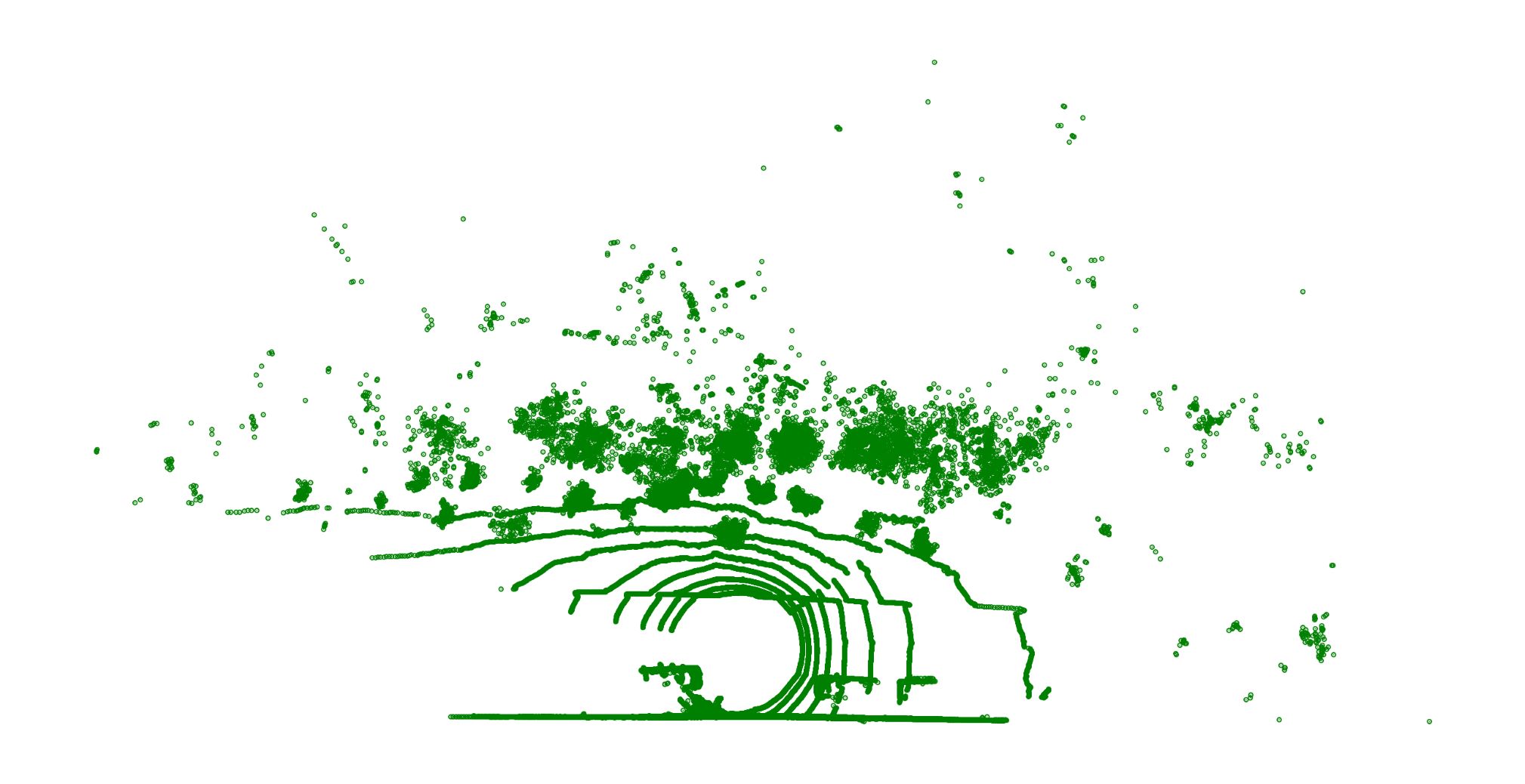}}
    \hspace{-0.5cm}			
    \subfloat[ICP without initial pose]{
		\includegraphics[width=3.7cm]{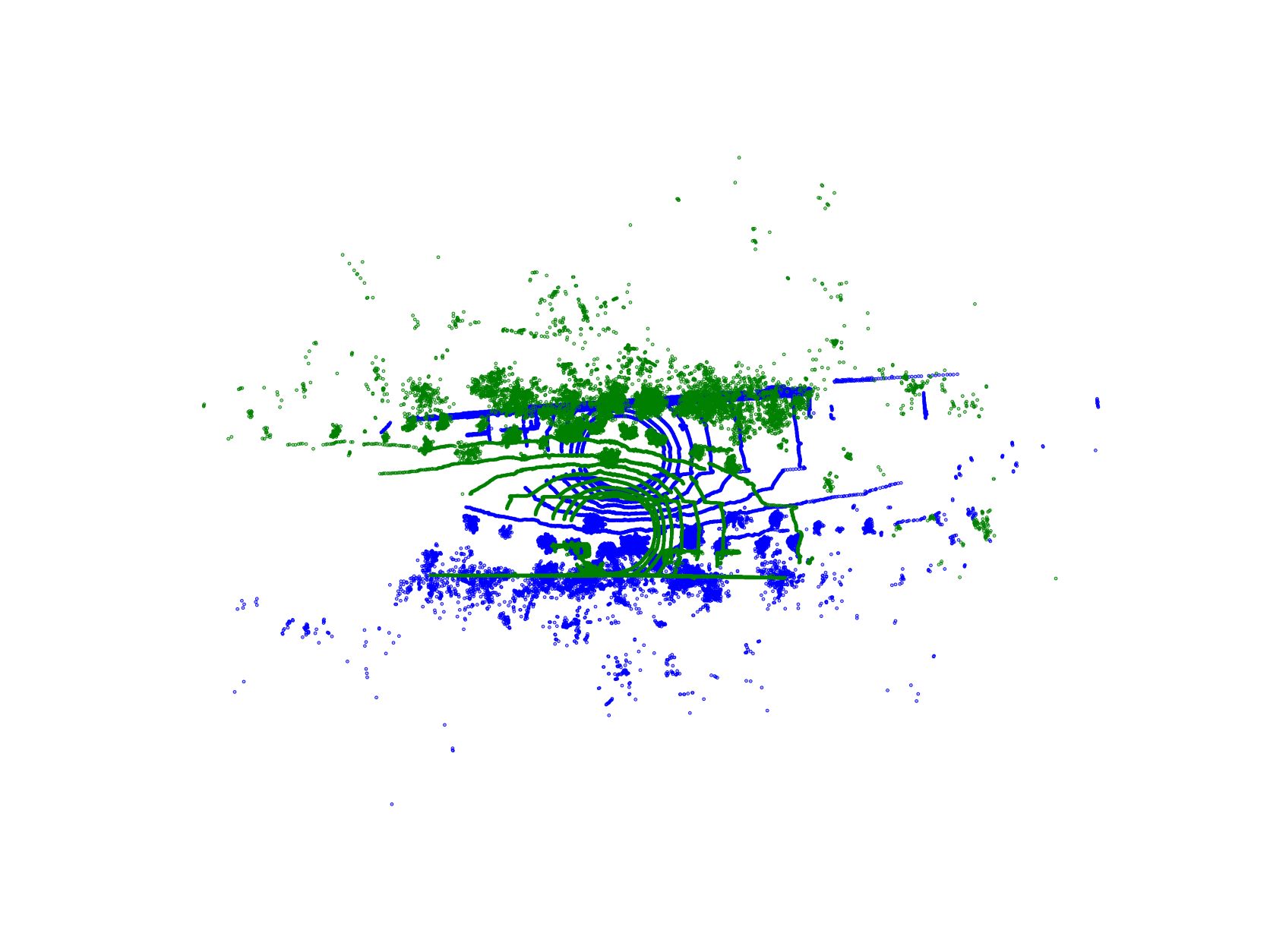}}
	\hspace{-0.5cm}	
    \subfloat[Alignment by RING++]{
		\includegraphics[width=3.7cm]{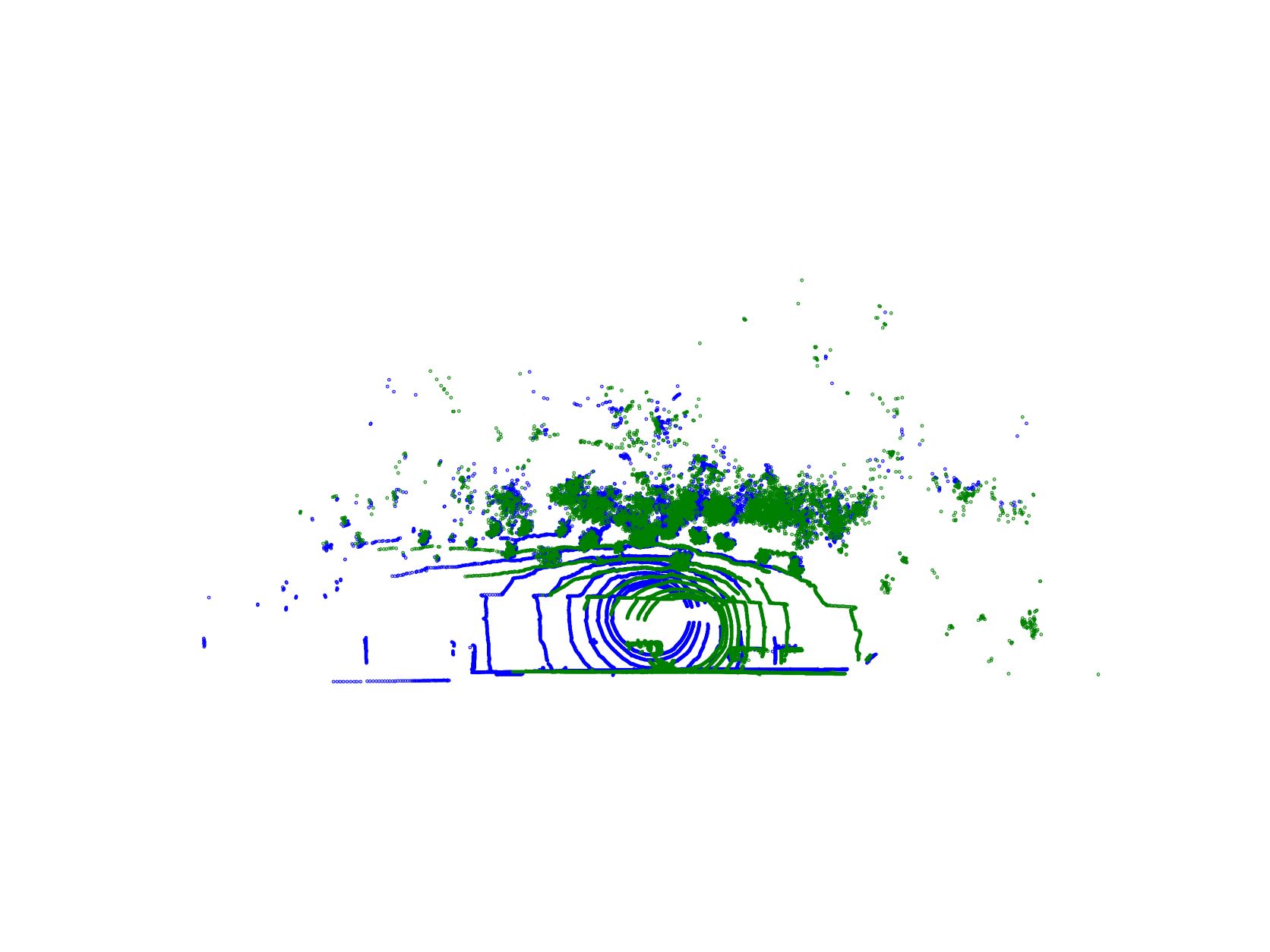}}	
    \hspace{-0.5cm}			
    \subfloat[ICP with RING++ initial pose]{
		\includegraphics[width=3.7cm]{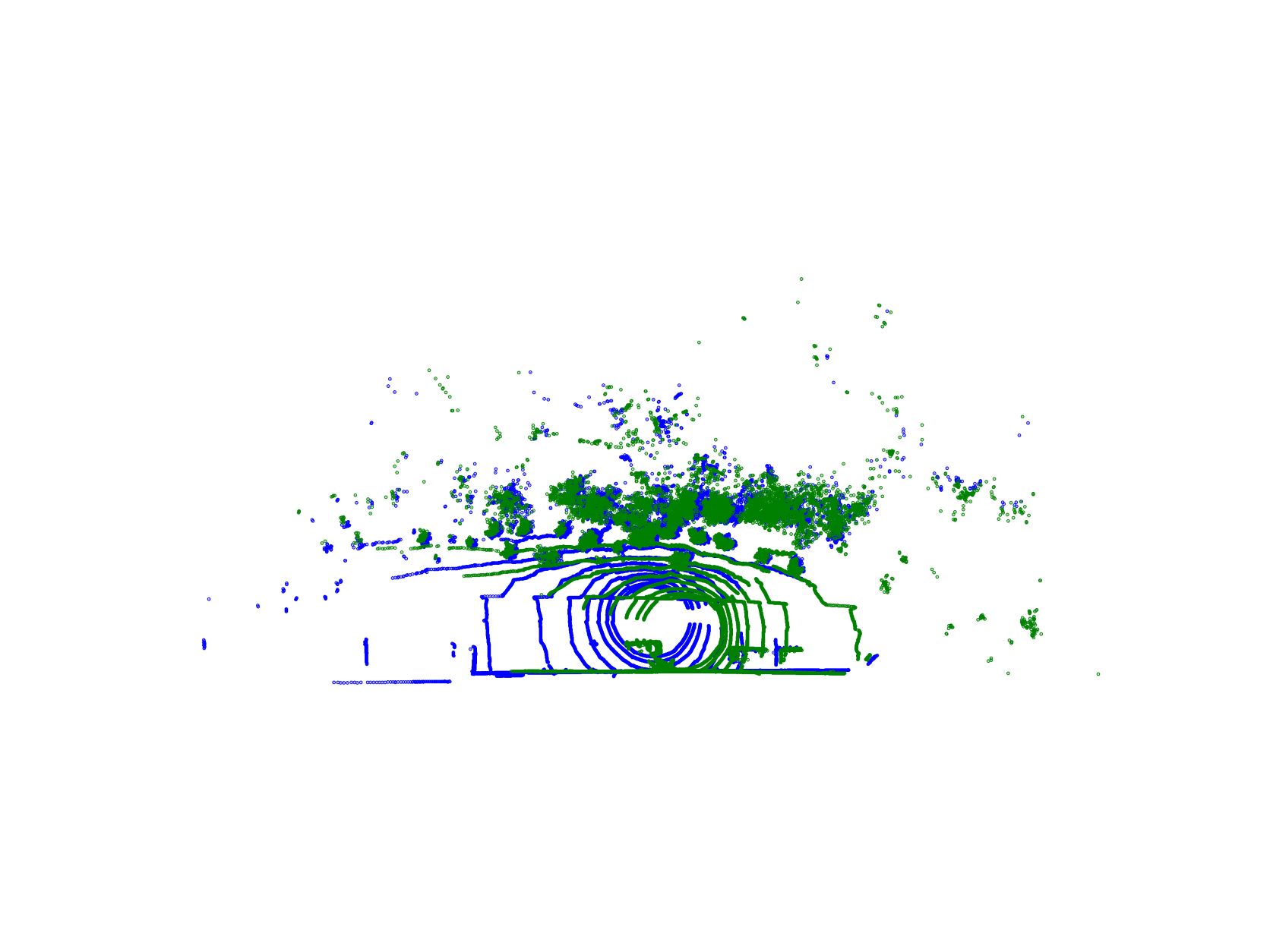}}	
	\caption{ICP alignment with and without initial pose provided by RING++.}
	\label{fig:icp}
    \vspace{-0.2cm}
\end{figure*}

To verify the quality of the 3-DoF pose estimated by our method, we employ ICP with and without the initial pose provided by the proposed method to align two matched scans after place recognition. The case results are illustrated in Fig.~\ref{fig:icp}. As we can easily see, ICP fails without an initial guess, especially in the case of large rotation variance. Our approach offers a qualified initial guess for scan matching, close to the best alignment. With the estimated pose of RING++ (SG) and RING++, ICP converges to global minima rather than traps in local optima. Furthermore, the time cost of ICP with the initial guess is much less than without any initial poses.

After that, we compare RING++ with both handcrafted and learning-based methods on NCLT and MulRan datasets. For each dataset, we evaluate these approaches in terms of \textit{Success Rate}, \textit{TE} and \textit{RE}. To demonstrate the effectiveness of better initial guesses, we also provide an additional pose estimation successful rate, which is calculated by the percentage of successfully aligned matches $(TE <2 m \; \& \; RE < 5^\circ)$ among all correctly retrieved pairs.
% We evaluate these methods on NCLT and MulRan datasets since they provide accurate 6-DoF ground truth poses while the GPS/INS data of Oxford dataset is not precise enough. 

\subsubsection{Comparison with Learning-free Methods}
Among the compared hand-engineering methods, M2DP and Fast Histogram are only capable of place recognition, while Scan Context and Scan Context++ produce 1-DoF relative pose (i.e., relative rotation or lateral translation) together with place recognition, leaving the remaining dimension to ICP. In contrast, our method yields a relative rotation that is a by-product in the process of place recognition and estimates a 2-DoF relative translation successively in the process of pose estimation. For fair comparison, we evaluate the pose estimation performance of all methods after refinement by ICP. As can be seen in Tab.~\ref{pose error 2}, we calculate \textit{Success Rate} for global localization evaluation and \textit{TE} and \textit{RE} at $50\%$, $75\%$ and $95\%$ for pose estimation evaluation. Compared with SC-based methods, RING++ has better performance on both NCLT and MulRan datasets, arriving at high \textit{Success Rate} for global localization. Referring to the pose estimation success rate, RING++ has a better chance of achieving satisfied alignment by providing the qualified 3-DoF initial guesses, which makes refinement more easily converge to the global optimal pose. Our method achieves less \textit{TE} and \textit{RE} at most quantiles, which verifies the efficacy of better initial guesses. 

\subsubsection{Comparison with Learning-based Methods}
Learning-based methods, DiSCO and EgoNN, are trained on MulRan dataset and tested on both MulRan and NCLT datasets for generalization evaluation. For MulRan dataset, EgoNN shows great performance of pose estimation thanks to the extracted local features. RING++ presents competitive performance ($66.09\%$ \textit{Success Rate}) to EgoNN ($62.00\%$ \textit{Success Rate}). The overall global localization is mainly limited by the place recognition performance which we discuss in the Section~\ref{limitations}.  In terms of pose estimation error, RING++ has lower \textit{TE} and \textit{RE} than DiSCO and EgoNN. For NCLT dataset, the distribution of points is different from that of MulRan dataset due to the different LiDAR sensor, which makes it hard for generalization. Therefore, the global localization performance of learning-based methods on NCLT dataset drops a lot, especially for EgoNN. The underlying reason may be explained by the lack of explicit invariance design in EgoNN, making the feature correlated and fail to generalize to unseen datasets. Unlike EgoNN, DiSCO constructs a rotation invariant place descriptor by transforming features to the frequency domain and then taking the magnitude of the frequency spectrum, which explains the relatively smaller decline of \textit{Success Rate} value from MulRan dataset generalizing to NCLT dataset. Through comparison, RING++, as a learning-free method arrives at the top performance on NCLT dataset, with high \textit{Success Rate} and small \textit{TE} and \textit{RE}.

\begin{table}[tbp]
\renewcommand\arraystretch{1.2}
\centering
\caption{Ablation Study on the Resolution$^{*}$}
\label{resolution}
\begin{threeparttable}
\begin{tabular}{cccc}
\toprule[1pt]
Resolution & Recall@1 (\%) & TE (m) & RE (\textdegree) \\ \hline
40$\times$40 & \multirow{2}{*}{62.18} & \multirow{2}{*}{1.43/2.08/3.73} & \multirow{2}{*}{2.45/4.14/7.27} \\ 
(3.50 m, 9\textdegree) & & & \\ 
60$\times$60 & \multirow{2}{*}{68.76} &  \multirow{2}{*}{0.98/1.37/2.52} &  \multirow{2}{*}{1.86/3.19/5.96} \\ 
(2.33 m, 6\textdegree) & & & \\ 
80$\times$80 & \multirow{2}{*}{71.59} &  \multirow{2}{*}{0.79/1.10/1.95}&  \multirow{2}{*}{1.53/2.70/5.32} \\ 
(1.75 m, 4.5\textdegree) & & & \\ 
120$\times$120 & \multirow{2}{*}{73.21} &  \multirow{2}{*}{0.56/0.77/1.31}&  \multirow{2}{*}{1.25/2.14/4.07} \\ 
(1.17 m, 3\textdegree) & & & \\ 
160$\times$160 & \multirow{2}{*}{75.83} &  \multirow{2}{*}{0.48/0.70/1.30} &  \multirow{2}{*}{1.08/1.98/3.59} \\ 
(0.88 m, 2.25\textdegree) & & & \\ 
200$\times$200 & \multirow{2}{*}{76.54}&  \multirow{2}{*}{0.41/0.60/1.10} &  \multirow{2}{*}{1.08/1.93/3.26} \\ 
(0.70 m, 1.8\textdegree) & & & \\ 
300$\times$300 & \multirow{2}{*}{77.65} &  \multirow{2}{*}{0.32/0.48/1.13} &  \multirow{2}{*}{1.07/1.79/3.13} \\ 
(0.47 m, 1.2\textdegree) & & & \\
\bottomrule[1pt]
\end{tabular} 
\begin{tablenotes}
    \footnotesize
    \item[*] We list 50\%, 75\% and 95\% quantiles of TE and RE in this table.
\end{tablenotes}
\end{threeparttable}
\vspace{-0.3cm}
\end{table}

\subsection{Ablation Study}
\label{ablation}
To investigate the influence of resolution on place recognition and pose estimation, we present ablation studies on the grid size and corresponding resolution of RING++. We carry out experiments utilizing the same multi-session pair ``2012-03-17" to ``2012-02-04" in NCLT dataset, with the results displayed in Tab.~\ref{resolution}. As the grid size/resolution increases, \textit{Recall@1} of our method increases fast at first and then approaches a constant gradually. Additionally, translation estimation error declines slightly, while rotation estimation benefits greatly from the increased resolution.

The other ablation study is carried out to confirm RING++'s effectiveness as a feature aggregation method. Using the same multi-session pair from the NCLT dataset, we test RING++ using several local features with the resolution set to $120 \times 120$. The results are shown in Tab.~\ref{feature_ablation}. We utilize FPFH \cite{rusu2010fast} implemented in Open3D \cite{Zhou2018}, which extracts 33-channel point features. Regarding SHOT \cite{tombari2011combined, SALTI2014251}, we use its Python implementation and change the bin value to 2, which results in the final 64-channel features. Although the bin value can be increased to 11 as in the original work, we only set it to 2 because of the linear memory consumption growth during the evaluation. The results show that our RING++ can also perform well with only two signature bins. Compared to the features we selected, FPFH and SHOT show competitive performance, validating the RING++ framework as a generic feature aggregation method. To strike a balance between precision and storage, we use 6-channel features introduced in Section~\ref{feat_extract} in all other experiments.

\begin{table}[t]
\renewcommand\arraystretch{1.2}
\centering
\caption{Ablation Study on Local Features$^{*}$}
\label{feature_ablation}
\begin{threeparttable} 
\begin{tabular}{cccc}
\toprule
Local Features & Recall@1 (\%)&TE (m) & RE (\textdegree) \\ \hline
FPFH\cite{rusu2010fast} &  \multirow{2}{*}{75.03} &  \multirow{2}{*}{0.56/0.79/1.30} &  \multirow{2}{*}{1.19/2.03/4.03} \\ 
 w/ RING++ & & & \\
SHOT\cite{SALTI2014251} &  \multirow{2}{*}{63.73} &  \multirow{2}{*}{0.56/0.79/1.39} &  \multirow{2}{*}{1.16/2.03/4.08}  \\
 w/ RING++ & & & \\
\bottomrule
\end{tabular}
\begin{tablenotes}
    \footnotesize
    \item[*] We list 50\%, 75\% and 95\% quantiles of TE and RE in this table.
\end{tablenotes}
\end{threeparttable}
\vspace{-0.3cm}
\end{table}

% \begin{table}[htbp]
% \renewcommand\arraystretch{1.2}
% \centering
% \caption{Performance Comparison of Resolution Analysis at MulRan DCC02 (query) to DCC01 (map)}
% \label{resolution}
% \begin{tabular}{c|c|c|c}
% \hline
% \hline
% Resolution & Recall@1 & TE (m) & RE (\textdegree) \\ \hline
% 40$\times$40 (3.50 m, 9\textdegree) & 0.6806 & 2.7462 & 1.1406 \\ \hline
% 60$\times$60 (2.33 m, 6\textdegree) & 0.6840 & 3.0007 & 1.0619 \\ \hline
% 80$\times$80 (1.75 m, 4.5\textdegree) & 0.6736 & 3.0027 & 1.0095 \\ \hline
% 120$\times$120 (1.17 m, 3\textdegree)  & 0.6771 & 3.0123 & 1.0754 \\ \hline
% 160$\times$160 (0.88 m, 2.25\textdegree)  & 0.6684 & 3.0313 & 1.0713 \\ \hline
% 400$\times$400 (0.35 m, 0.9\textdegree)  & 0.6615 & 3.0429 & 1.2254 \\ \hline
% 500$\times$500 (0.28 m, 0.72\textdegree)  & 0.6563 & 2.9972 & 1.3296 \\ 
% \hline
% \hline
% \end{tabular}
% \end{table}

\begin{figure}[t]
	\centering
	\includegraphics[width=8.5cm]{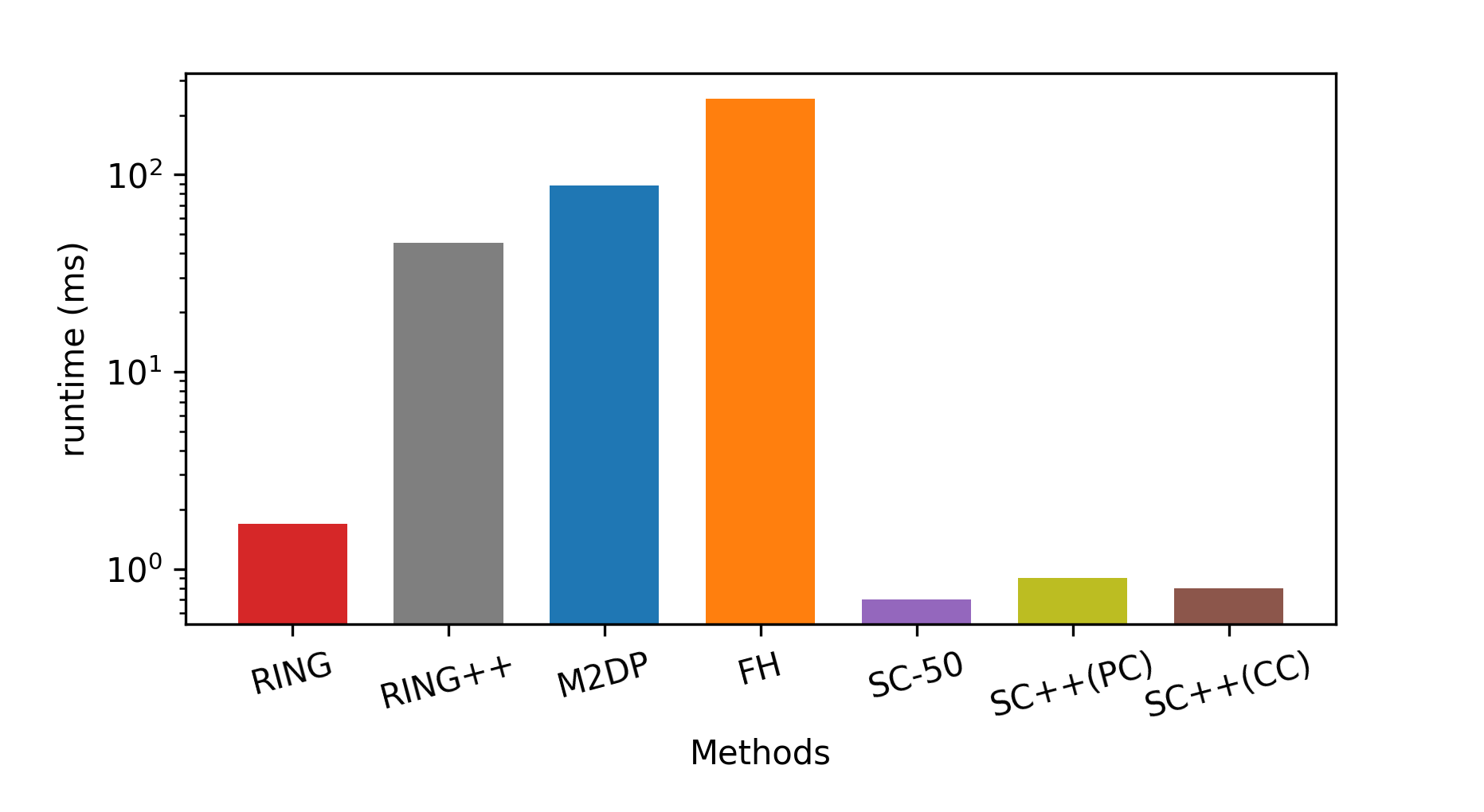}
	\caption{Average computational time when tested on NCLT Dataset. FH represents Fast Histogram.}
	\label{fig:computation_cost}
    \vspace{-0.3cm}
\end{figure}

\subsection{Computational Cost}
Considering the real-time constraint, we calculate the computation cost of our algorithm. All methods are implemented in Python and tested on a system equipped with an Intel i7-10700 (2.9GHz) and an Nvidia GeForce RTX 2060 SUPER with 8G memory. As can be seen in Fig.~\ref{fig:computation_cost}, the average computational time of RING++ is less than $50ms$. The most time-consuming part is the calculation of local features. It is worth noting that the time consumption varies according to the number of channels in the local feature. It grows linearly as the feature channel increases.

\subsection{Limitations and Potential Extension}
\label{limitations}
\subsubsection{Limited Evaluation Performance}
RING++ can retrieve places with large viewpoint differences thanks to its roto-translation invariance property. As a result, the retrieved places may not be close enough to satisfy revisited criteria during evaluation, leading to a reduced place recognition success rate. However, in real-world applications, the success rate of place recognition is not a primary goal. Despite the fact that some of the retrieved places are not close to the current one, our estimators can provide satisfactory initial poses, resulting in improved overall global localization performance.

\subsubsection{Leveraging Deep learning for Feature Extraction}
We utilize handcrafted local features to validate our multi-channel framework; however, the representation ability of these features is limited. To achieve better performance, a possible way is to utilize learning-based local features with an end-to-end learning framework that takes RING++ as an aggregation.

%\subsubsection{Application to Other Range Sensors}
%The proposed framework can be regarded as an general aggregation method which is not limited to LiDAR sensors but also applicable to other range sensors.

% \begin{table*}[htbp]
% \renewcommand\arraystretch{1.2}
% \centering
% \caption{Time Cost (ms) and Memory Usage (KB) of Different Methods}
% \label{computation cost}
% \begin{tabular}{lccccccc}
% \toprule
% Approach & RING (ours) & RING++ (ours) & M2DP & Fast Histogram & SC-50 & SC++ (PC) & SC++ (CC) \\ \hline
% Description Time & 1.7 & 45.2 & 88.1 & 243.5 & 0.7 & 0.9 & 0.8 \\
% Memory Usage & 56.3  & 337.5 & 1.5 & 0.8 & 56.3 & 57.2 & 57.2 \\ 
% \bottomrule
% \end{tabular}
% \end{table*}  

\ifx\allfiles\undefined
\end{document}
\fi
\ifx\allfiles\undefined

\begin{document}
\fi

\begin{figure}[t]
	\centering
	\subfloat[SC++ SLAM]{
		\includegraphics[width=\linewidth]{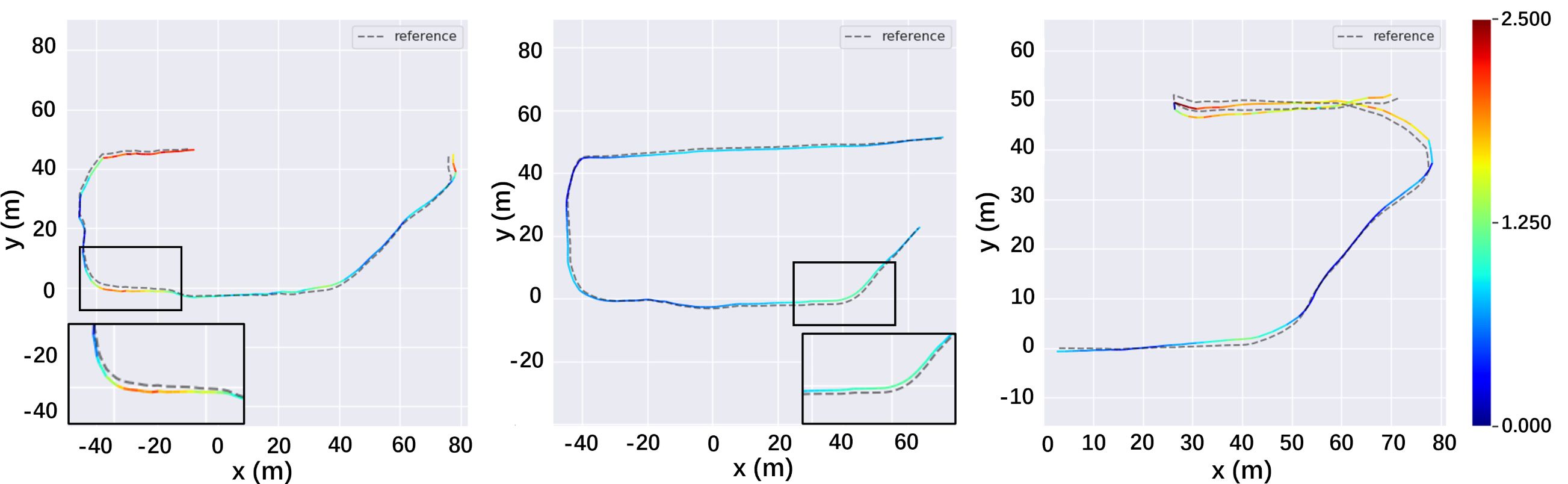}}\\
	\subfloat[RING++ SLAM]{
		\includegraphics[width=\linewidth]{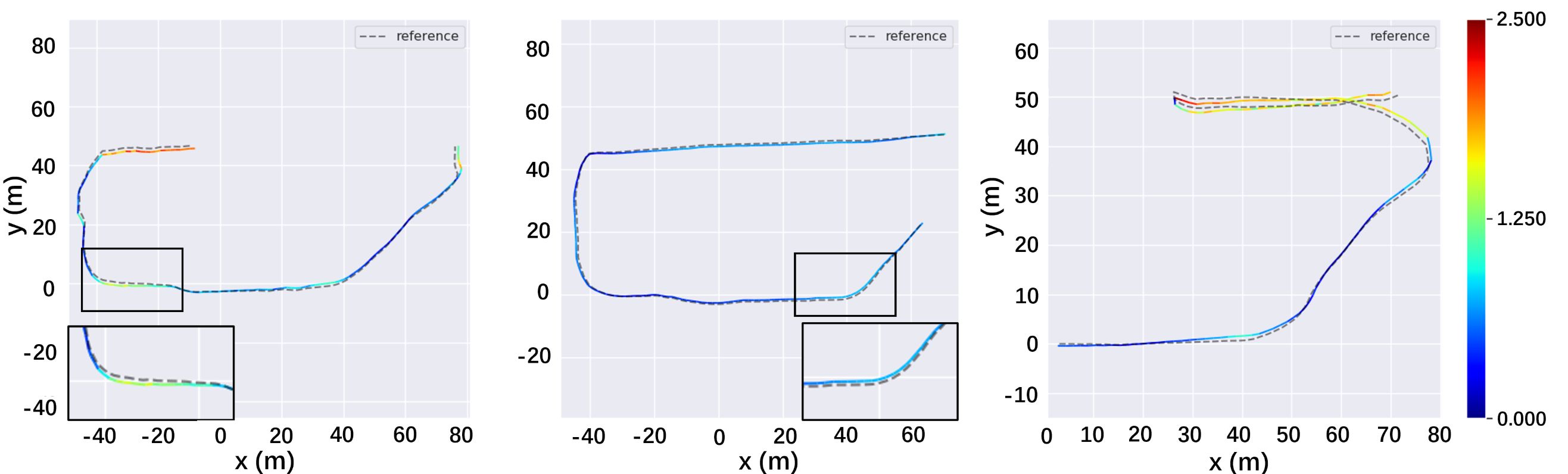}}
	\caption{Error visualization of Scan Context++ (PC) integrated SLAM system (top) compared to our approach (bottom) on legged robot dataset.}
	\label{fig:dog_slam}
\vspace{-0.2cm}
\end{figure}

\section{System Application}
As a lightweight framework, we implement the proposed RING++ into a stand-alone module without any prior information. We also integrate a real-time back-end manager with the pose graph optimizer implemented by GTSAM \cite{dellaert2012factor}. With the optimization results acquired, we rearrange the keyframes to generate a global map. These components, together with any front-end odometry, can form a complete SLAM system. To validate the performance of our approach in real-world applications, we integrate FAST-LIO2 \cite{xu2022fast} into our SLAM system and compare RING++ and Scan context++ (PC) within this SLAM system.

\subsection{Multi-robot SLAM System}
\begin{figure}[tbp]
	\centering
	\includegraphics[width=8.5cm]{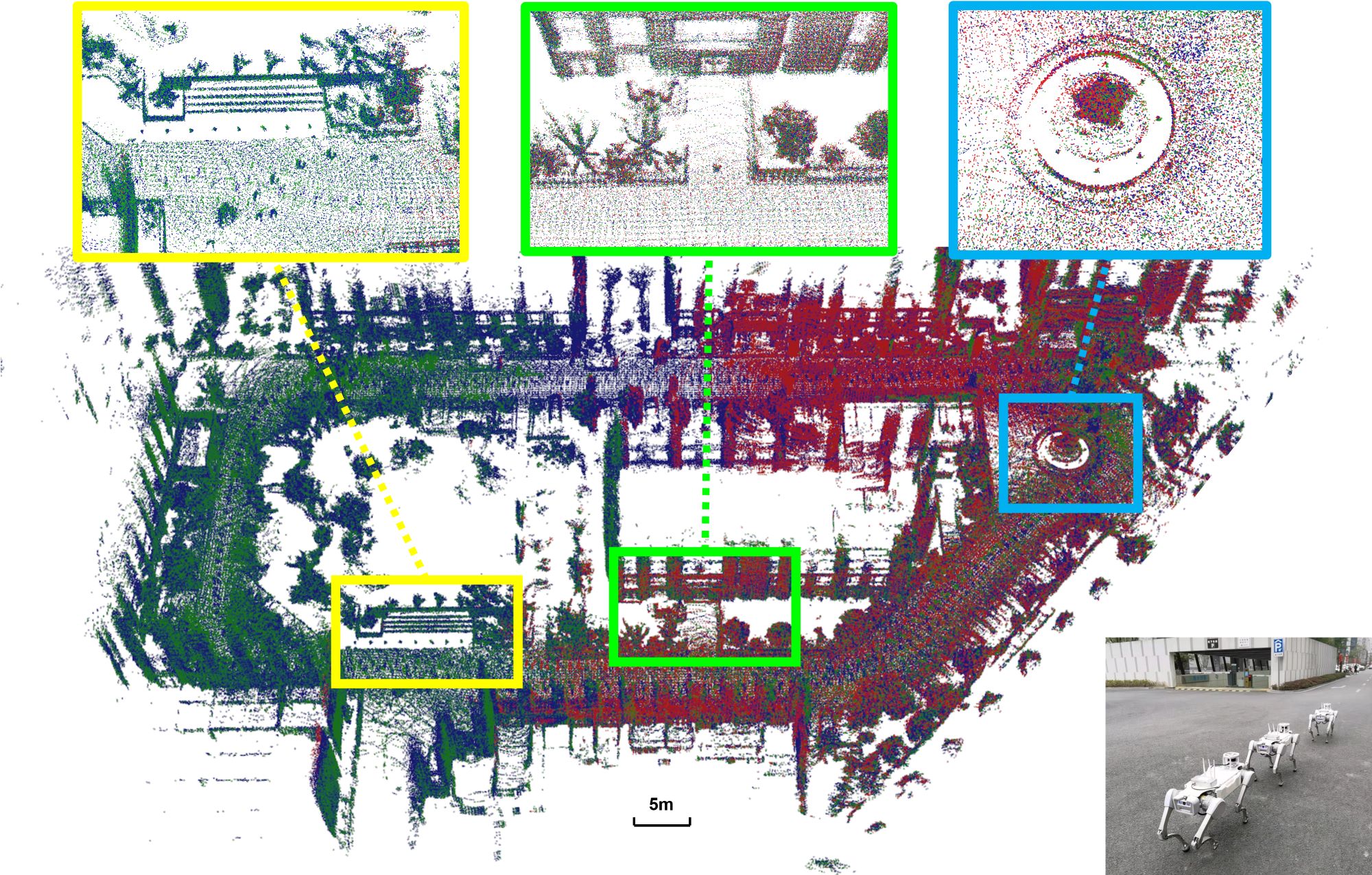}
	\caption{Qualitative results of our RING++ SLAM on legged robot dataset. Different colored points are generated from different legged robots.}
	\label{fig:dog_slam_cloud}
\end{figure}

In real applications such as exploration and rescue, multi-robot SLAM systems are expected to quickly build a precise environment map. The precision of the map is largely dependent on the correct alignment between keyframes. Because of the motion flexibility, legged robots are frequently used in such scenarios \cite{bellicoso2018advances}. With this background, we collected data from three legged robots outfitted with an IMU-integrated Ouster64 LiDAR and a Jetson Xavier NX. The ground truth is acquired by the offline interactive SLAM \cite{koide2020interactive} with automatically added and handpicked edges. In this setup, keyframes are generated at $2m$ intervals. The results are shown in Tab.~\ref{SLAM_performance} and Fig.~\ref{fig:dog_slam}, RING++ provides more successfully aligned loops for back-end pose optimization, which leads to lower drifts. The point cloud built by our system can be seen in Fig.~\ref{fig:dog_slam_cloud}, points generated by different robots are well aligned.

\begin{table}[t]
 \centering
 \renewcommand\arraystretch{1.2}
\caption{ATE and successfully aligned loop number of Scan Context++ (PC) and RING++ integrated SLAM systems.}
\label{SLAM_performance}
\setlength{\tabcolsep}{2mm}{
\begin{tabular}{lcccc}
\toprule
\multirow{2}{*}{Approach} & \multicolumn{2}{c}{Legged Robot Dataset} & \multicolumn{2}{c}{NCLT} \\ \cline{2-5}
& ATE (m) & Succ. Loop  & ATE (m) & Succ. Loop \\  \hline
SC++ SLAM     & 1.56   &   151     &  7.89    &  66 \\ 
\textbf{RING++ SLAM} & \textbf{0.72}   &   \textbf{167}     &  \textbf{6.43}    &  \textbf{126} \\ 
\bottomrule
\end{tabular}}
\end{table}

\subsection{Multi-session SLAM System}
\begin{figure}[tbp]
	\centering
	\subfloat[SC++ SLAM]{
		\includegraphics[width=\linewidth]{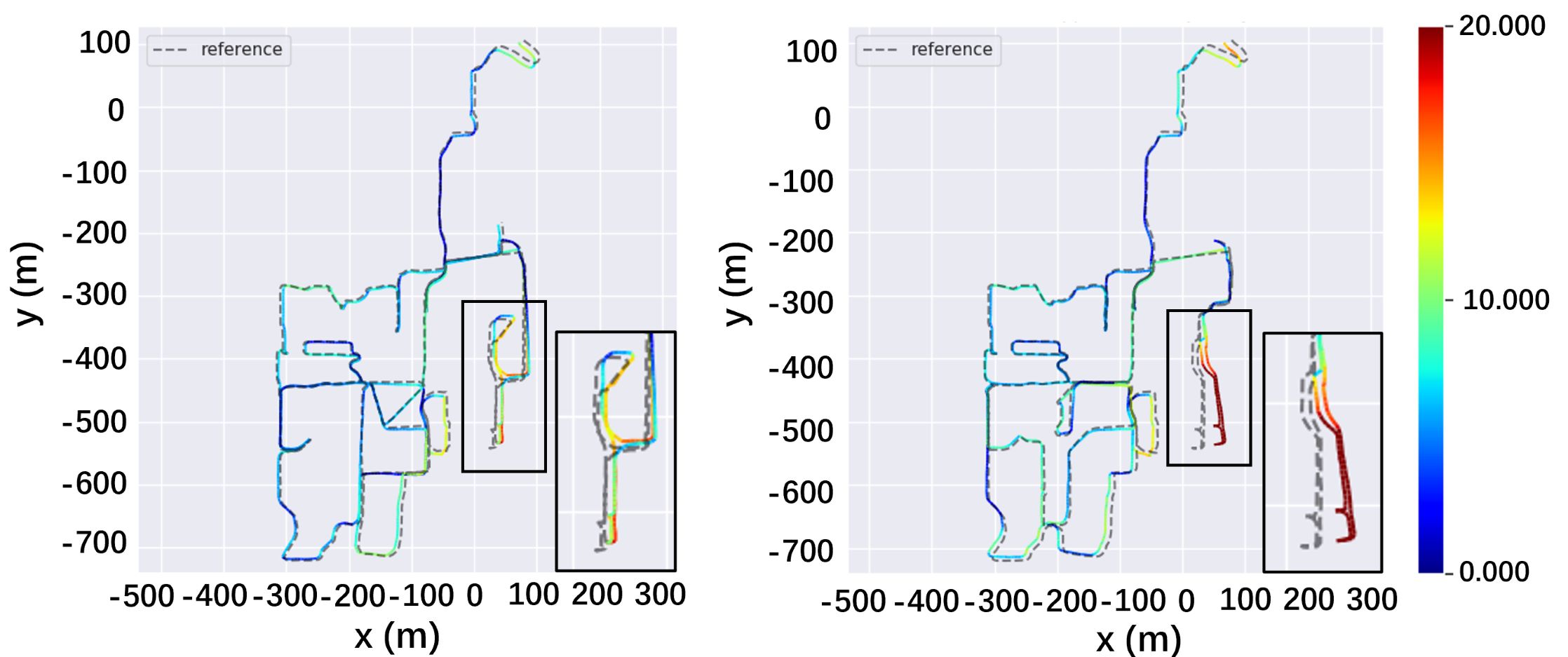}}\\
    \subfloat[RING++ SLAM]{
		\includegraphics[width=\linewidth]{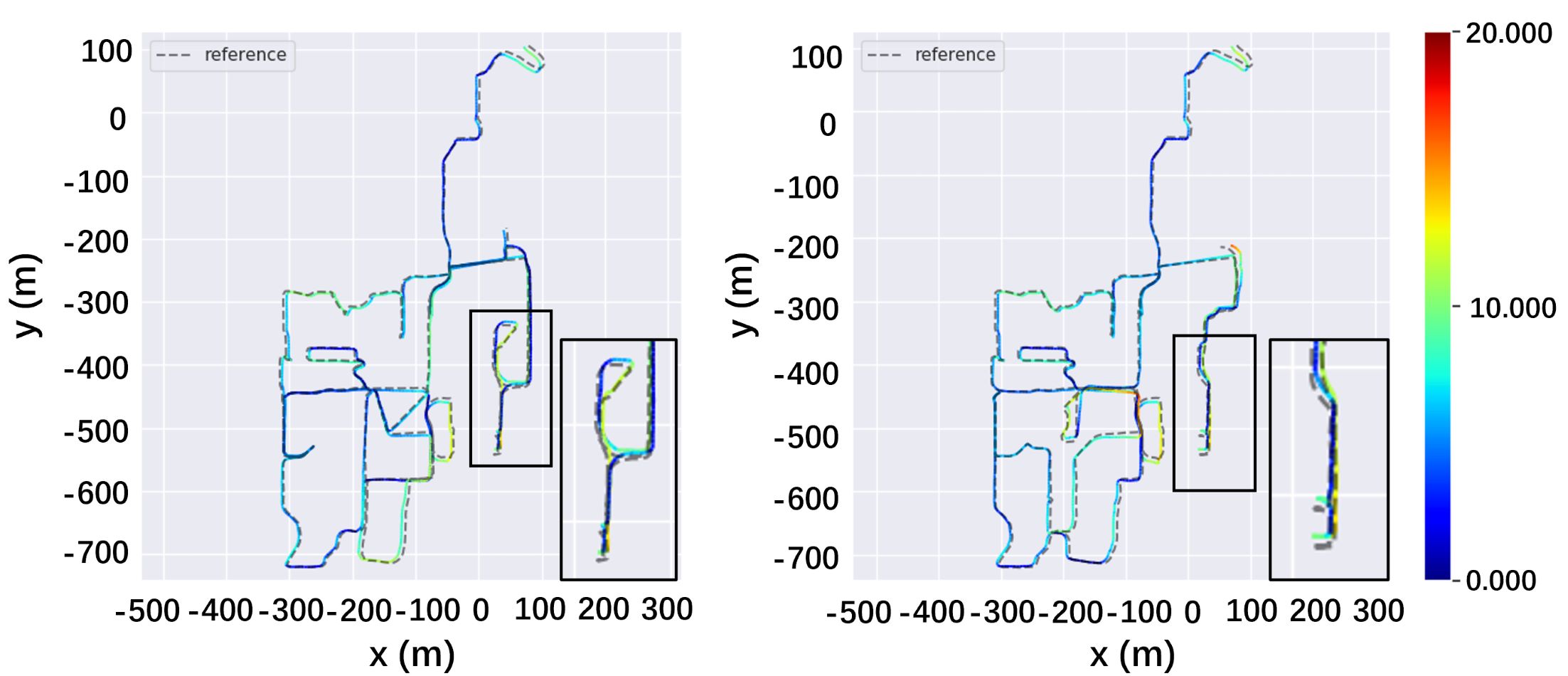}}		
	\caption{Error visualization of Scan Context++ (PC)  integrated SLAM system (top) compared to our approach (bottom) on two sequences of NCLT dataset. (left:``2012-05-26", right:``2012-03-17")}
	\label{fig:nclt_slam}
	\vspace{-0.2cm}
\end{figure}

\begin{figure}[tp]
	\centering
	\includegraphics[width=8.5cm]{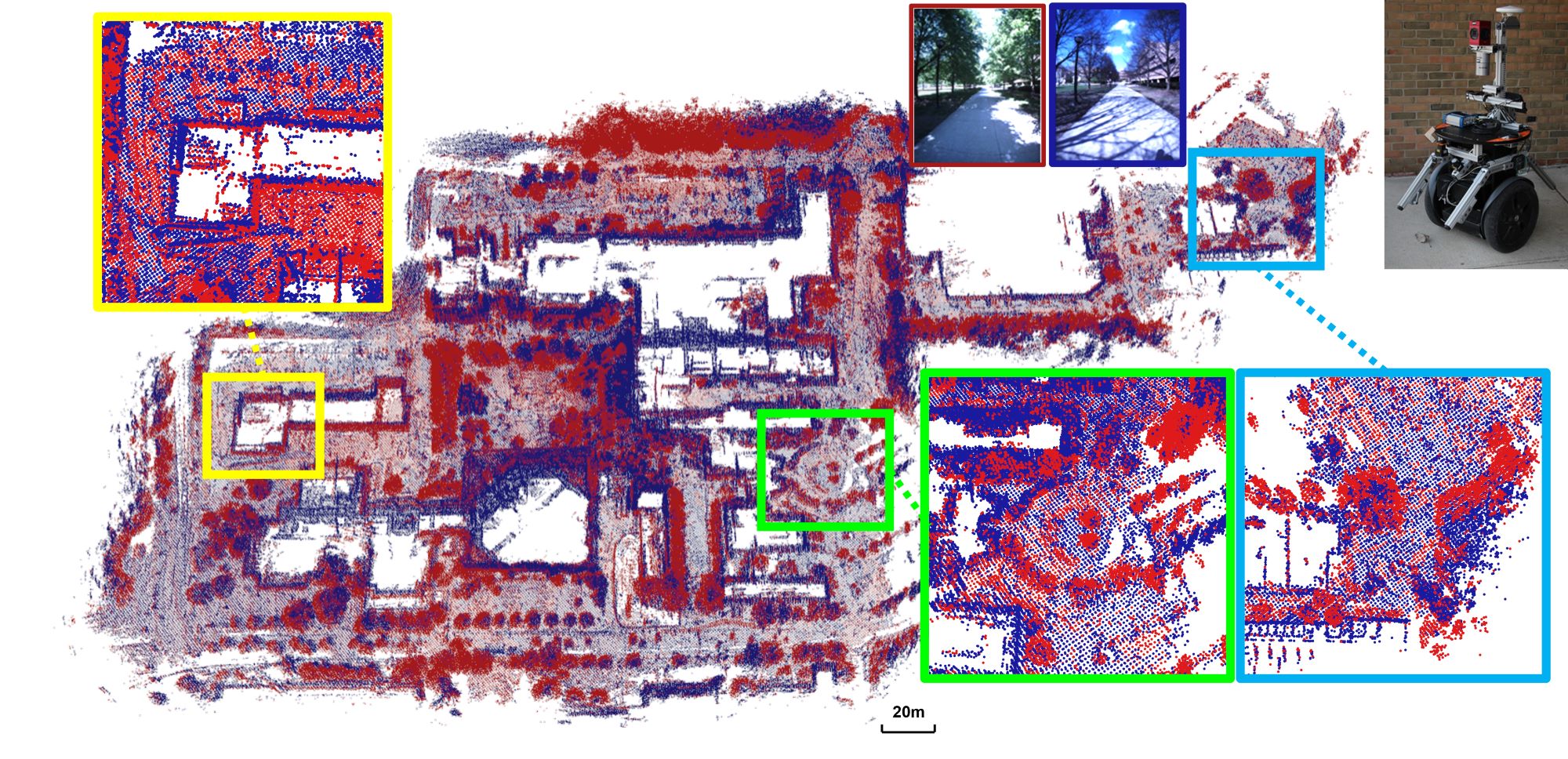}
	\caption{Qualitative results of our RING++ SLAM on NCLT Dataset. Red points are generated from ``2012-05-26" while blue points are generated from ``2012-03-17".}
	\label{fig:nclt_slam_cloud}
	\vspace{-0.2cm}
\end{figure}

Another application of RING++ is multi-session SLAM. It differs from multi-robot SLAM system in the changing of environment. The chosen NCLT dataset contains a wide range of environmental changes, such as dynamic objects, seasonal changes like winter and summer, and structural changes like building construction. We evaluate our system on ``2012-05-26" and ``2012-03-17". The data from these sessions is processed online at the same time, indicating a multi-robot setup. The LiDAR points replayed online are from different sessions, forming a challenging temporal/spatial multi-agent setup. To maintain a sparse representation for the very large environment, we generate keyframes at $5m$ intervals. The performance can be seen in Tab.~\ref{SLAM_performance} and Fig.~\ref{fig:nclt_slam}, RING++ provides almost twice the number of loop closures than Scan Context++ (PC), thus has better performance. The mapping result is shown in Fig.~\ref{fig:nclt_slam_cloud}, where two different colored points represent two NCLT dataset sequences. The overlapping landmarks in magnified figures indicate the low drift of our system. It should be noted that FAST-LIO2 failed to provide acceptable odometry in the final minutes of the NCLT dataset, so those parts were discarded in the evaluation.

% \begin{figure}[htbp]
% 	\centering
% 	\includegraphics[width=8.5cm]{figs/slam_error.png}
% 	\caption{Performance of Scan Context integrated SLAM system (left) compared to our approach (right) on NCLT dataset}
% 	\label{fig:nclt_slam}
%     \vspace{-0.5cm}
% \end{figure}

\ifx\allfiles\undefined
\end{document}
\fi
\ifx\allfiles\undefined

\begin{document}
\fi

\section{Conclusion}
In this paper, we present a roto-translation invariant framework RING++ for global localization on the sparse scan map, including representation pass and solving pass. Specifically, we extract six local features with geometry information and aggregate the features into three representations in the representation pass: rotation equivariant SG, translation invariant and rotation equivariant TING, and roto-translation invariant RING. In the solving pass, we perform place recognition, rotation estimation, translation estimation, and pose refinement. Thanks to roto-translation invariant representation, RING++ achieves superior performance on benchmark datasets, outperforming the state-of-the-art methods. By integrating RING++ as a stand-alone loop closure detection module into SLAM systems, we validate the effectiveness of our approach without any prior knowledge in real scenarios.

\ifx\allfiles\undefined
\end{document}
\fi
\ifx\allfiles\undefined

\begin{document}
\fi

% {\appendix
{
% Use $\backslash${\tt{appendix}} if you have a single appendix:
% Do not use $\backslash${\tt{section}} anymore after $\backslash${\tt{appendix}}, only $\backslash${\tt{section*}}.
% If you have multiple appendixes use $\backslash${\tt{appendices}} then use $\backslash${\tt{section}} to start each appendix.
% You must declare a $\backslash${\tt{section}} before using any $\backslash${\tt{subsection}} or using $\backslash${\tt{label}} ($\backslash${\tt{appendices}} by itself
%  starts a section numbered zero.)}
% For formal statement, we define Eq. \ref{prt} as \textit{RING Equation} and Eq. \ref{pr} as \textit{Simplified RING Equation} used for place recognition in practice. 

{\appendices
\section{Proof of Eq. \ref{pr}}
\label{appendix proof 1}
After normalizing TING $M$ to $\tilde{M}$ with a mean of 0 and a standard deviation of 1, the resultant $\tilde{M}$ can be written as:
\begin{equation}
    \tilde{M} = \frac{M - \mu(M)}{\sigma(M)}
    \label{normalized TING}
\end{equation}
where $\mu(M)$ is the mean of $M$, and $\sigma(M)$ is the standard deviation of $M$. 

Then the resultant RING element $\tilde{N}_{Q, i}$ based on normalized $\tilde{M}_{Q}$ and $\tilde{M}_{i}$ is 
 \begin{equation}
 \begin{aligned}
    \tilde{N}_{Q,i} &= \max_{k_{\theta}} \tilde{\mathfrak{C}}_i (k_{\theta}) \\
    &= \max_{k_{\theta}} \sum_{\theta_j} \tilde{M}_Q(\theta_j + k_{\theta}) \tilde{M}_i(\theta_j)^T
 \end{aligned}
 \label{normalized RING}   
\end{equation}

Substituting Eq. \ref{normalized RING} into Eq. \ref{pr}, we have:
 \begin{equation}
 \begin{aligned}
    n &= \arg\max_i \tilde{N}_{Q,i} \\
    &= \arg\max_i (\max_{k_{\theta}} \sum_{\theta_j} \tilde{M}_Q(\theta_j + k_{\theta}) \tilde{M}_i(\theta_j)^T)
 \end{aligned}
\end{equation}

For efficient place retrieval leveraging Eq. \ref{pr}, we need to prove that Eq. \ref{pr} is mathematically equivalent to Eq. \ref{prt}. Since the length $l$ of RING descriptor varies with the number of map scans, we begin with the proof at the simplest case $l = 2$. With $l = 2$, let $\tilde{N}_{Q} = (s_1, s_2)$, so $\tilde{N}_{Q,1} = s_1$ and $\tilde{N}_{Q,2} = s_2$. Suppose $\tilde{N}_{i_1}$ is the RING descriptor of the first map scan and $\tilde{N}_{i_2}$ is that of the second map scan. Due to normalization, we can easily conclude that $s_1 \leq 1$ and $s_2 \leq 1$ according to Eq. \ref{normalized RING}. In terms of $\tilde{N}_{i_1}$ and $\tilde{N}_{i_2}$, we have $\tilde{N}_{i_1, 1} = \tilde{N}_{i_2, 2} = 1$ because of auto-correlation:
 \begin{equation}
 \begin{aligned}
    \tilde{N}_{i,i} &= \max_{k_{\theta}} \sum_{\theta_j} \tilde{M}_i(\theta_j + k_{\theta}) \tilde{M}_i(\theta_j)^T \\
    &= \sum_{\theta_j} \tilde{M}_i(\theta_j) \tilde{M}_i(\theta_j)^T \\
    &= |\tilde{M}_i|^2 = 1
 \end{aligned}
 \label{autocorr}   
\end{equation}

Supposing $\tilde{N}_{i_1} = (1, k)$ where $k < 1$ since the two map scans are different, then $\tilde{N}_{i_2} = (k, 1)$. In this case, the problem turns to prove that $\|\tilde{N}_Q-\tilde{N}_{i_1}\| < \|\tilde{N}_Q-\tilde{N}_{i_2}\|$ if and only if $\tilde{N}_{Q,1} > \tilde{N}_{Q,2}$,  which involves two sub-proofs: sufficiency proof and necessity proof. 
% it turns to prove that $(s_1-1)^2 + (s_2-k)^2 < (s_1-k)^2 + (s_2-1)^2$ if and only if $s_1 > s_2$.

\textbf{Proof of Sufficiency:} We begin with proof of sufficiency. The Euclidean distance between $\tilde{N}_Q$ and $\tilde{N}_{i_1}$, $\tilde{N}_{i_2}$ is formulated as:
\begin{equation}
\begin{aligned}
\|\tilde{N}_Q-\tilde{N}_{i_1}\| &= (s_1-1)^2 + (s_2-k)^2 \\ 
\|\tilde{N}_Q-\tilde{N}_{i_2}\| &= (s_1-k)^2 + (s_2-1)^2 \\ \nonumber
\end{aligned}
\end{equation}

If $\tilde{N}_{Q,1} > \tilde{N}_{Q,2}$, that is $s_1 > s_2$, then we have:
\begin{equation}
\|\tilde{N}_Q-\tilde{N}_{i_1}\| - \|\tilde{N}_Q-\tilde{N}_{i_2}\| = 2(k-1)(s_1-s_2) < 0 \nonumber   
\end{equation}

Thus, if $\tilde{N}_{Q,1} > \tilde{N}_{Q,2}$, $\|\tilde{N}_Q-\tilde{N}_{i_1}\| < \|\tilde{N}_Q-\tilde{N}_{i_2}\|$. 

\textbf{Proof of Necessity:} After that, we embark on proof of the necessity. If $\|\tilde{N}_Q-\tilde{N}_{i_1}\| < \|\tilde{N}_Q-\tilde{N}_{i_2}\|$, we have:
\begin{equation}
\|\tilde{N}_Q-\tilde{N}_{i_1}\| - \|\tilde{N}_Q-\tilde{N}_{i_2}\| = 2(k-1)(s_1-s_2) < 0 \nonumber
\end{equation}

Since $k < 1$, then we arrive at:
\begin{equation}
s_1 > s_2 \nonumber
\end{equation}

Thus, if $\|\tilde{N}_Q-\tilde{N}_{i_1}\| < \|\tilde{N}_Q-\tilde{N}_{i_2}\|$, $\tilde{N}_{Q,1} > \tilde{N}_{Q,2}$. 

Therefore, we conclude that Eq. \ref{prt} holds if and only if Eq. \ref{pr} holds.

\section{Proof of Fast Implementation by FFT}
\label{appendix_proof_2}
In order to fasten the computation of circular cross-correlation employed in RING formation, we take advantage of parallelized fast Fourier transform (FFT) to efficiently carry out cross-correlation. To be more specific, we apply DFT along $\theta$ axis of TING $M_{Q}$ and $M_{i}$ for all $\omega \triangleq \{\omega_m\}$, resulting in the frequency spectrum $F_{Q}(u, \omega_m)$ and $F_{n}(u, \omega_m)$:
\begin{equation}
\begin{aligned}
    {F}_{Q}(u, \omega_m) &= \mathcal{F}({M}_{Q}(\theta, \omega_m)) = \sum_{\theta_j} M_Q(\theta_j, \omega_m) e^{-i 2\pi u \frac{j}{J}} \\
    {F}_{n}(u, \omega_m) &= \mathcal{F}({M}_{n}(\theta, \omega_m)) = \sum_{\theta_j} M_n(\theta_j, \omega_m) e^{-i 2\pi u \frac{j}{J}} 
%     &= \sum_{\theta_j} M_Q(\theta_j + k_{\theta}, \omega) M_i(\theta_j, \omega)^T
\end{aligned}
\label{fft}
\end{equation}
where $u$ is the discrete frequency of the frequency spectrum. After that, we multiply $F_{Q}(u, \omega_m)$ by the complex conjugate of  $F_{n}(u, \omega_m)$ and then employ inverse Fourier transform (IFT) to obtain a 2D correlation map $\mathfrak{C}_i(k_{\theta}, k_{\omega})$:
\begin{equation}
\begin{aligned}
    \mathfrak{C}_i(k_{\theta}, k_{\omega}) &= \mathcal{F}^{-1} ({F}_{Q}(u, \omega_m) {F}^{*}_{n}(u, \omega_m)) \\
    &= \sum_{\theta_j} \sum_{\omega_m} M_Q(\theta_j + k_{\theta}, \omega_m + k_{\omega}) M_i(\theta_j, \omega_m)
\end{aligned}
\label{corrcirc}
\end{equation}
By summing along $k_{\omega}$ axis of the correlation map, we can obtain the resultant 1D correlation map $\mathfrak{C}_i(k_{\theta})$ of 1D circular cross-correlation on 2D image defined in this paper:
\begin{equation}
\begin{aligned}
    \mathfrak{C}_i(k_{\theta}) &= \sum_{\theta_j} \sum_{\omega_m} M_Q(\theta_j + k_{\theta}, \omega_m) M_i(\theta_j, \omega_m)\\
    &= \sum_{\theta_j} M_Q(\theta_j + k_{\theta}) M_i(\theta_j)^T
\end{aligned}
\label{corrcirc}
\end{equation}

}}

\ifx\allfiles\undefined
\end{document}
\fi

\bibliographystyle{IEEEtran}
% \bibliography{IEEEabrv, IEEEref}

% \newpage
 
\vspace{-0.3cm}

\begin{IEEEbiography}[{\includegraphics[width=1in,height=1.25in,clip,keepaspectratio]{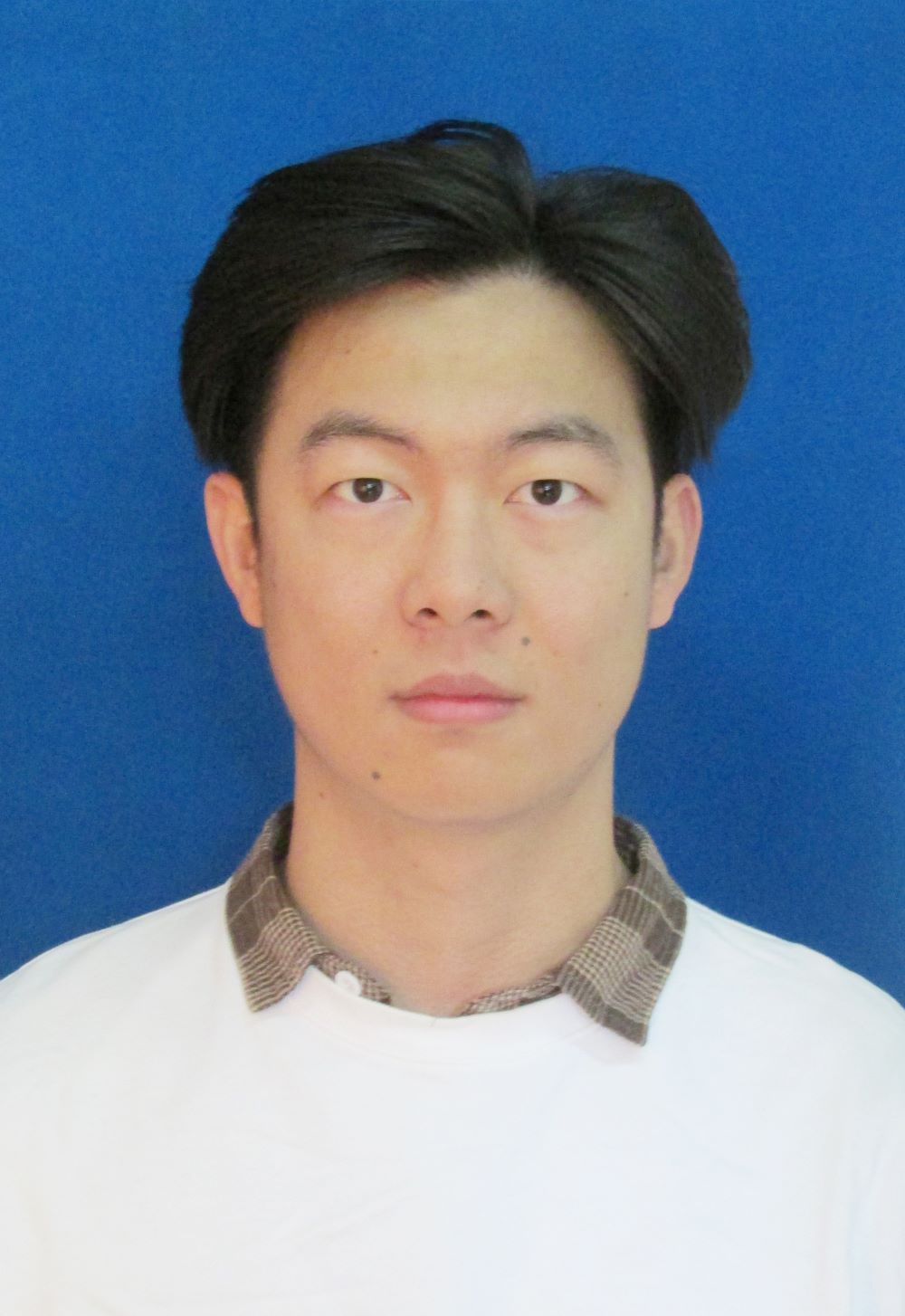}}]{Xuecheng Xu}
received the B.S. degree from the Department of Control Science and Engineering, Zhejiang University, Hangzhou, P.R. China, in 2019. He is currently a Ph.D student in the College of Control Science and Engineering, Zhejiang University, Hangzhou, P.R. China. His latest research interests include LiDAR simultaneous localization and mapping and multi-robot systems.  
\end{IEEEbiography}
\vspace{-5 mm}

\begin{IEEEbiography}[{\includegraphics[width=1in,height=1.25in,clip,keepaspectratio]{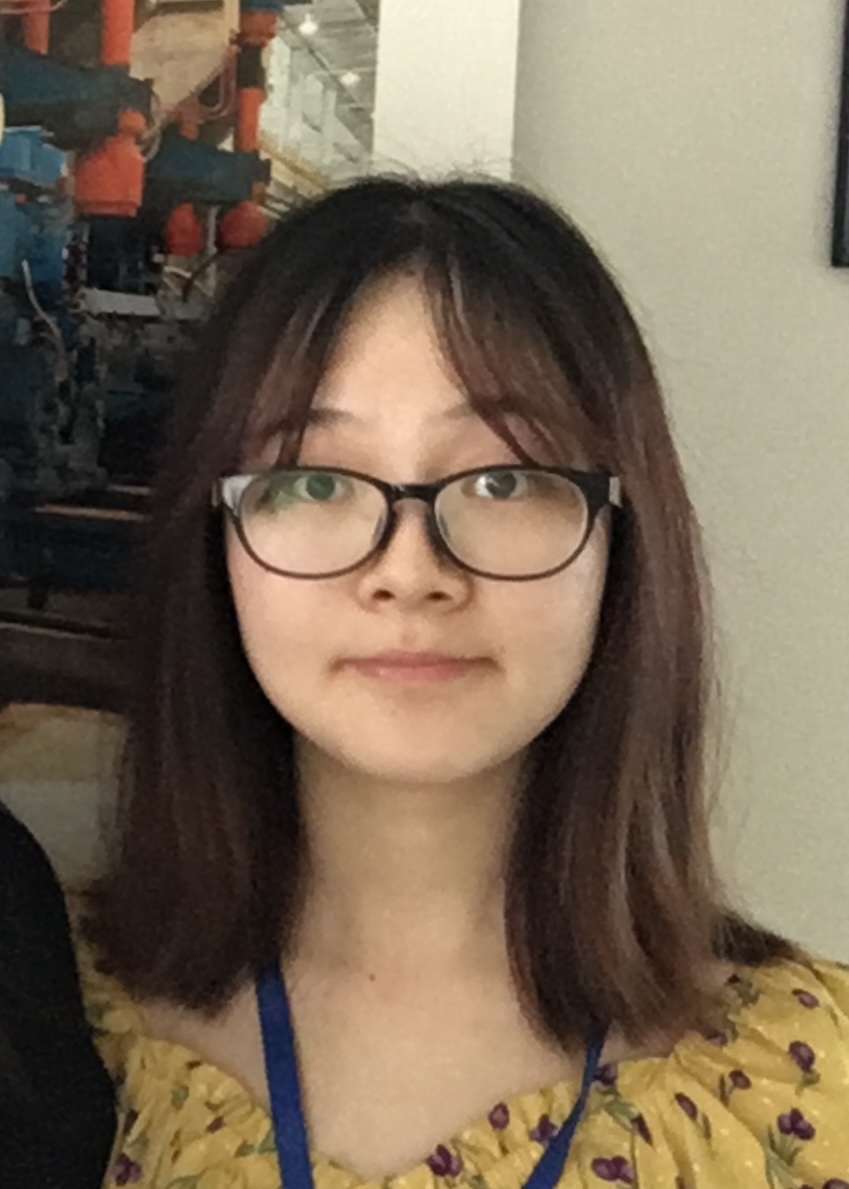}}]{Sha Lu}
received her B.S. from CQU-UC Joint Co-op Institute, Chongqing University, Chongqing, China, in 2021. She is currently working toward the M.S. degree at the State Key Laboratory of Industrial Control Technology and Institute of Cyber-Systems and Control, Zhejiang University, Hangzhou, China. Her research interests include robotics and deep learning. 
\end{IEEEbiography}
\vspace{-5 mm}

\begin{IEEEbiography}[{\includegraphics[width=1in,height=1.25in,clip,keepaspectratio]{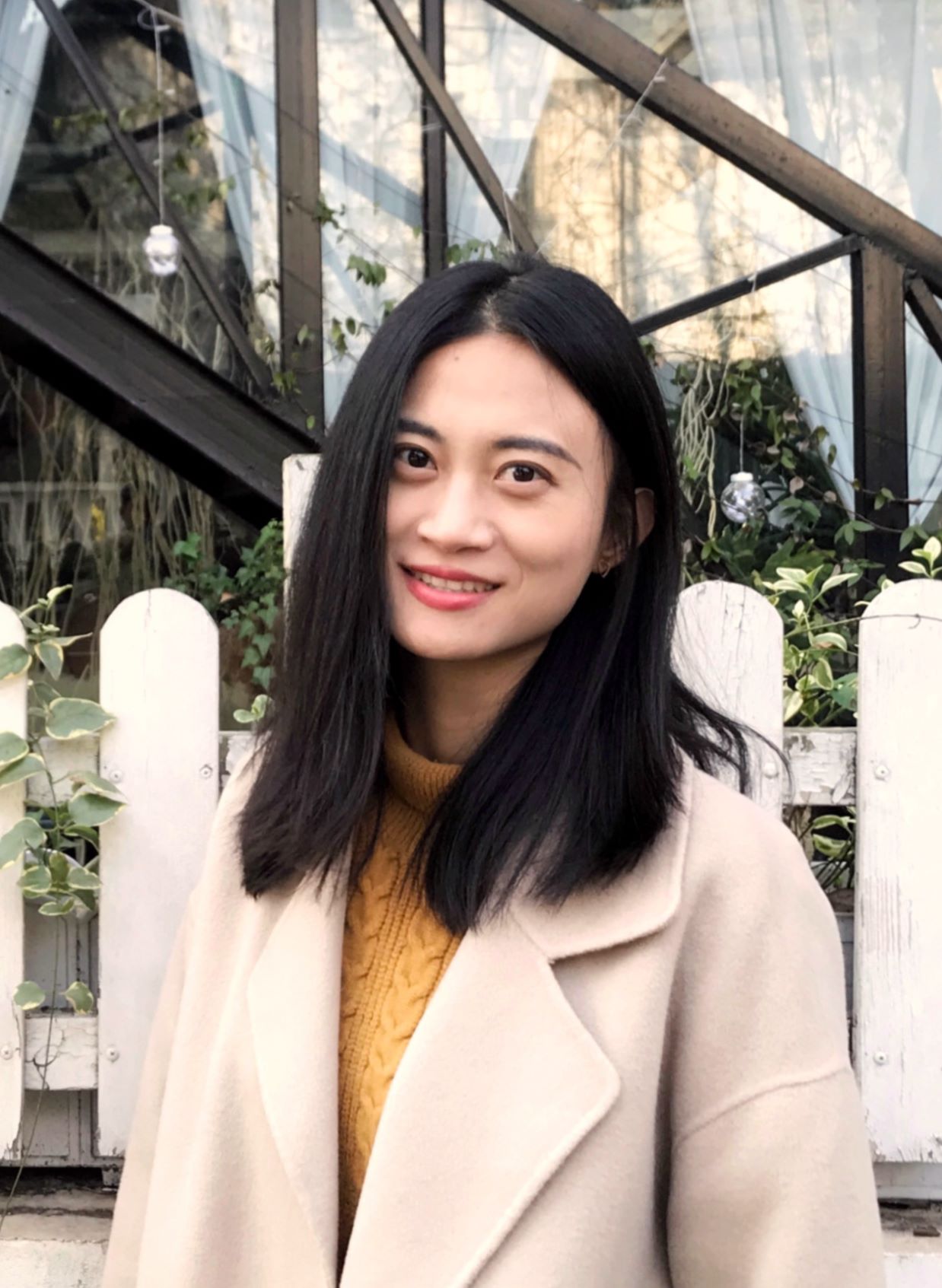}}]{Jun Wu}
received the M.E. in control engineering from Zhejiang University, Hangzhou, China, in 2018. She is currently working toward the Ph.D. degree at the State Key Laboratory of Industrial Control Technology and Institute of Cyber-Systems and Control, Zhejiang University, Hangzhou, China. Her research interests include robotics perception.
\end{IEEEbiography}
\vspace{-5 mm}

\begin{IEEEbiography}[{\includegraphics[width=1in,height=1.25in,clip,keepaspectratio]{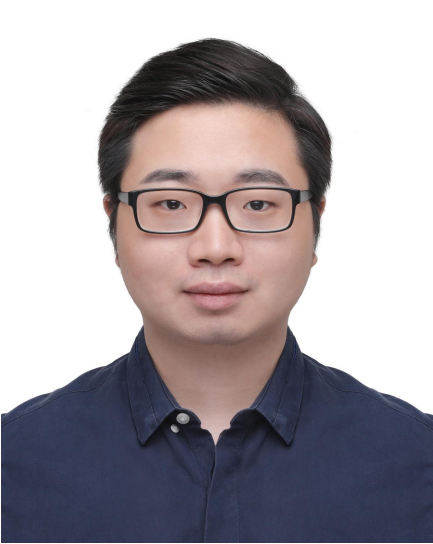}}]{Haojian Lu}
(Member, IEEE) received B.Eng. degree in Mechatronical Engineering from Beijing Institute of Technology in 2015, and he received
Ph.D degree in Robotics from City University of Hong Kong in 2019. He was a Research Assistant at City University of Hong Kong, from 2019 to
2020. He is currently a professor in the State Key Laboratory of Industrial Control and Technology, and Institute of Cyber-Systems and Control, Zhejiang University. His research interests include micro/nanorobotics, bioinspired robotics, medical robotics, micro aerial vehicle and soft robotics.
\end{IEEEbiography}
\vspace{-5 mm}

\begin{IEEEbiography}[{\includegraphics[width=1in,height=1.25in,clip,keepaspectratio]{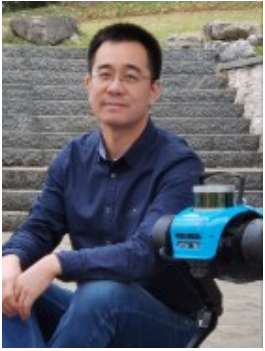}}]{Qiuguo Zhu}
received his B.S. degrees in Mechanical College in 2008, and Master degrees and Doctor degree in Control Science in 2011 and 2020, respectively. He is currently an Associate Professor at Department of Control Science in Zhejiang University, Zhejiang University. His research interests include the control of humanoid robots, bipedal and quadrupedal walking and running, manipulators, rehabilitation exoskeletons and machine intelligence.
\end{IEEEbiography}
\vspace{-5 mm}

\begin{IEEEbiography}[{\includegraphics[width=1in,height=1.25in,clip,keepaspectratio]{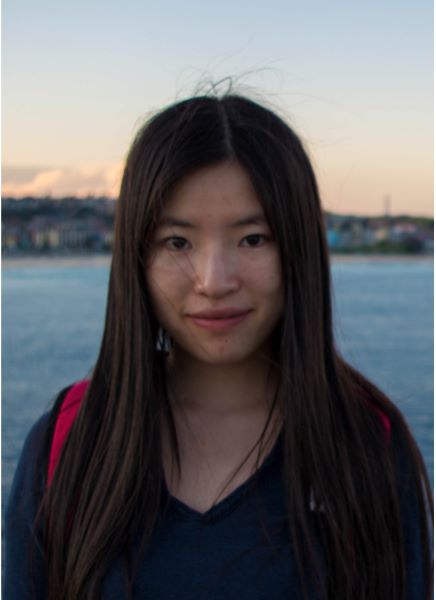}}]{Yiyi Liao}
received her Ph.D. degree from the Department of Control Science and Engineering, Zhejiang University, China in 2018. From 2018 to 2021, she was a postdoctoral researcher at the Autonomous Vision Group, University of Tubingen and Max Planck Institute for Intelligent Systems, Germany. She is currently an Assistant Professor at Zhejiang University. Her research
interests include 3D vision and scene understanding.
\end{IEEEbiography}
\vspace{-5 mm}

\begin{IEEEbiography}[{\includegraphics[width=1in,height=1.25in,clip,keepaspectratio]{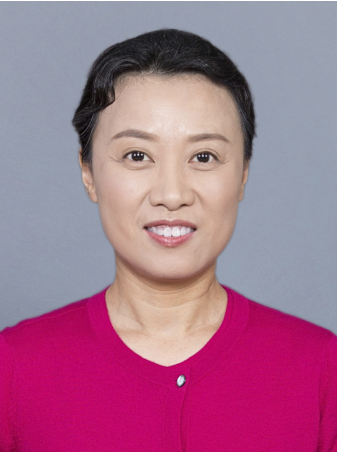}}]{Rong Xiong}
received her Ph.D. degree in Control Science and Engineering from the Department of Control Science and Engineering, Zhejiang University, Hangzhou, P.R. China in 2009. She is currently a Professor in the Department of Control Science and Engineering, Zhejiang University, Hangzhou, P.R. China. Her latest research interests include motion planning and SLAM.
\end{IEEEbiography}
\vspace{-5 mm}

\begin{IEEEbiography}[{\includegraphics[width=1in,height=1.25in,clip,keepaspectratio]{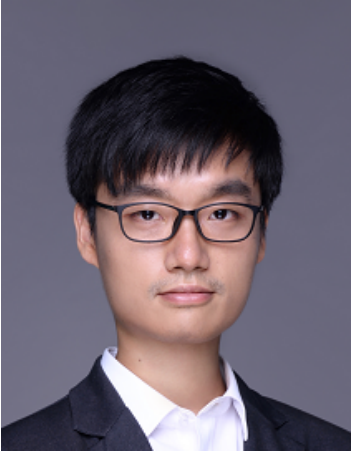}}]{Yue Wang}
(Member, IEEE) received the Ph.D. degree from the Department of Control Science and Engineering, Zhejiang University, Hangzhou, China, in 2016. He is currently working as an Associate Professor with the Department of Control Science and Engineering, Zhejiang University. His current research interests include mobile robotics and robot perception.
\end{IEEEbiography}

\vfill

\end{document}